\documentclass[10pt]{article} 
\usepackage[usenames,dvipsnames]{xcolor}
\usepackage[preprint]{tmlr}

\usepackage{hyperref}
\usepackage{url}

\usepackage{microtype}
\usepackage{graphicx}
\usepackage{booktabs} 
\usepackage{multirow}
\usepackage{tabularx}
\usepackage{amsmath,amsfonts,amssymb}
\usepackage{wrapfig}

\usepackage{caption}
\usepackage{subcaption}

\usepackage{enumitem}
\usepackage{CJKutf8}

\usepackage{mathtools}
\usepackage{amsthm}

\DeclareMathOperator*{\argmin}{argmin}
\DeclareMathOperator*{\argmax}{argmax}

\usepackage[utf8]{inputenc} 
\usepackage[T1]{fontenc}    
\usepackage{nicefrac}       

\usepackage{algorithm}
\let\classAND\AND
\let\AND\relax
\usepackage{algorithmic}

\let\AND\classAND
\AtBeginEnvironment{algorithmic}{\let\AND\algoAND}

\usepackage[capitalize,noabbrev]{cleveref}

\title{TransFool: An Adversarial Attack against Neural Machine Translation Models}

\author{\name Sahar Sadrizadeh \email sahar.sadrizadeh@epfl.ch \\
      \addr EPFL, Lausanne, Switzerland 
      \AND
      \name Ljiljana Dolamic \email ljiljana.dolamic@ar.admin.ch \\
      \addr Armasuisse S+T, Thun, Switzerlan
      \AND
      \name Pascal Frossard \email pascal.frossard@epfl.ch\\
      \addr EPFL, Lausanne, Switzerland }

\def\month{06}  
\def\year{2023} 
\def\openreview{\url{https://openreview.net/forum?id=sFk3aBNb81}} 

\begin{document}

\maketitle

\vspace{20pt}
\begin{abstract}
Deep neural networks have been shown to be vulnerable to small perturbations of their inputs, known as adversarial attacks. In this paper, we investigate the vulnerability of Neural Machine Translation (NMT) models to adversarial attacks and propose a new attack algorithm called \textit{TransFool}. To fool NMT models, TransFool builds on a multi-term optimization problem and a gradient projection step. By integrating the embedding representation of a language model, we generate fluent adversarial examples in the source language that maintain a high level of semantic similarity with the clean samples. Experimental results demonstrate that, for different translation tasks and NMT architectures, our white-box attack can severely degrade the translation quality while the semantic similarity between the original and the adversarial sentences stays high. Moreover, we show that TransFool is transferable to unknown target models. 
Finally, based on \textit{automatic} and \textit{human} evaluations,
TransFool leads to improvement in terms of success rate, semantic similarity, and fluency compared to the existing attacks both in white-box and black-box settings. Thus, TransFool permits us to better characterize the vulnerability of NMT models and outlines the necessity to design strong defense mechanisms and more robust NMT systems for real-life applications. 
\end{abstract}

\section{Introduction}

The impressive performance of Deep Neural Networks (DNNs) in different areas such as computer vision \citep{he2016deep} and Natural Language Processing (NLP) \citep{vaswani2017attention} has led to their widespread usage in various applications. With such an extensive usage of these models, it is important to analyze their robustness and potential vulnerabilities. In particular, it has been shown that the outputs of these models are susceptible to imperceptible changes in the input, known as adversarial attacks \citep{szegedy2014intriguing}.  Adversarial examples, which differ from the original inputs in an imperceptible manner, cause the  target  model to generate incorrect outputs. If these models are not robust enough to these attacks, they cannot be reliably used in  applications with security requirements. To address this issue, many studies have been recently devoted to the effective generation of adversarial examples, the defense against attacks, and the analysis of the vulnerabilities of DNN models \citep{moosavi2016deepfool,madry2018towards,ortiz2021optimism}.

The dominant methods to craft imperceptible attacks for continuous data, e.g., audio and image data, are based on gradient computing and various optimization strategies. However, these methods cannot be directly extended to NLP models due to the discrete nature of the tokens in the corresponding representations (i.e., words, subwords, and characters
). Another challenge in dealing with textual data is  the characterization of the imperceptibility of the adversarial perturbation.  The $\ell_p$-norm is highly utilized  in image data to measure imperceptibility but it does not apply to textual data where manipulating only one token in a sentence may significantly change the semantics. Moreover, in gradient-based methods, it is challenging to incorporate linguistic constraints in a differentiable manner.  Hence, optimization-based methods are more difficult and less investigated for 
adversarial attacks against NLP models. Currently, most attacks in textual data are gradient-free and simply based on heuristic word replacement, which may result in \textit{sub-optimal} performance {\citep{alzantot2018generating,ren2019generating,jin2020bert,li2020bert,morris2020reevaluating,zang2020word,guo2021gradient,sadrizadeh2022block,wang2022semattack}}. 

An important task in NLP is Neural Machine Translation (NMT), which involves automatically converting a sequence of words in a source language to a sequence of words in a target language \citep{bahdanau2015neural}. By using DNN models, NMT systems have achieved exceptional performance and are increasingly used in safety and security sensitive applications like medical and legal applications \citep{vieira2021understanding}. In this application also, adversarial attacks provide insights into the vulnerabilities of these systems, particularly when exposed to samples outside the training distribution. 
In particular, we study untargeted attacks, in which the adversary aims to generate adversarial examples that are semantically similar to the input sentences while their corresponding translations differ significantly \citep{michel2019evaluation,cheng2019robust,cheng2020seq2sick,zhang2021crafting}. It is expected that a good NMT model generates similar translations for similar sentences. However, even small perturbations such as changing a name, tense, gender, or numbers can degrade the translation quality. 
This unexpected behavior of NMT models has serious implications for real-world applications. For instance, if an NMT model performs well on a medical text, we expect that changing the name of the patient, diagnosis, or document number would still result in a good translation by the NMT model, which is not always the case. Therefore, these adversarial attacks, which are similar to the inputs but lead to a significant change in the model's output, outline weaknesses in the NMT model. Our goal is to evaluate the robustness of NMT models to such adversarial sentences to eventually  improve the \textit{security} of applications and the \textit{robustness} of such models.

In this paper, we propose \textit{TransFool} to build  \textit{semantically similar} and \textit{fluent} adversarial attacks against NMT models. We build a new solution to the challenges associated with gradient-based adversarial attacks against textual data. To find an adversarial sentence that is fluent and semantically similar to the input sentence but highly degrades the translation quality of the target model, we propose a  multi-term optimization problem over the tokens of the adversarial example. We consider the white-box attack setting, where the adversary has access to the target model and its parameters. White-box attacks are  widely studied since they reveal the vulnerabilities of the systems and are used in benchmarks.  
To ensure that the generated adversarial examples  are  similar to the original sentences,  we incorporate a Language Model (LM) in our method in two ways. First, we consider the 
loss of a Causal Language Model (CLM)  in our optimization problem in order to impose the syntactic correctness of the adversarial example. Second, by working with the embedding representation of  an LM,  instead of the NMT model, 
we ensure that  similar tokens are close to each other in the embedding space \citep{tenney2019you}. This enables the definition of a similarity term between the respective tokens of the clean and adversarial sequences. Hence, we include a similarity constraint in the proposed optimization problem, which uses the LM embeddings. Finally, our optimization contains an adversarial term to maximize the loss of the target NMT model.  

The generated adversarial example, i.e., the minimizer of the proposed optimization problem, should consist of meaningful tokens, and hence, the proposed optimization problem should be solved in a discrete space. By using a gradient projection technique, we first consider the continuous space of the embedding space and perform a gradient descent step and then, we project the resultant embedding vectors to the most similar valid token. 
In the projection step, we again use the LM embeddings and  project the output of the gradient descent step into the nearest meaningful token in the  embedding space (with maximum cosine similarity).  We test our method against different NMT models with transformer structures, which are now widely used for their exceptional performance. For different NMT architectures and translation tasks, experiments show that our white-box attack can reduce the translation quality 
while  it maintains a high level of semantic similarity with the clean samples. Also, we extend TransFool to black-box settings and show that it can fool unknown target models. 
Overall, automatic and human evaluations show that 
in both  white-box and black-box settings, TransFool outperforms the existing attacks in terms of success rate, semantic similarity, and fluency. 
In summary, our contributions are as follows:
\begin{itemize}[leftmargin=*]
    \item   We define a new optimization problem to compute semantic-preserving and fluent attacks against NMT models. The objective function contains several terms: adversarial loss to maximize the loss of the target NMT model; a similarity  term to ensure that the adversarial example is \textit{similar} to the original sentence; and  
    loss of a CLM to generate  \textit{fluent} and \textit{natural} adversarial examples. 
    \vspace{-2pt}
    
    \item We propose a new strategy to incorporate linguistic constraints in our attack in a differentiable manner. Since LM embeddings provide a meaningful representation of the tokens, we use them instead of the NMT embeddings to compute the similarity between two tokens.
    \vspace{-2pt}
    
    \item We design a white-box attack algorithm, 
    \textit{TransFool},  against NMT models by solving the proposed optimization problem with gradient projection. Our attack,  which operates at the token level, is effective against state-of-the-art  NMT models and \textit{outperforms} prior works.  
    \vspace{-2pt}
    
    \item By using the transferability of adversarial attacks to other models, we extend the proposed white-box attack to the black-box setting. Our attack is 
    highly effective 
    even when the \textit{target languages} of the target NMT model and the reference model are \textit{different}. To our knowledge, this type of  attack, \textit{cross-lingual}, has not been investigated.  
      
\end{itemize}

The rest of the paper is organized as follows. We  review the related works  in Section \ref{literature}. In Section \ref{method}, we formulate the problem of 
adversarial attacks against NMT models, and propose an optimization problem to generate adversarial examples. We  describe our attack algorithm in Section \ref{algorithm}.  In Section \ref{Experiments}, we discuss the experimental results and evaluate TransFool against different transformer models and translation tasks. Moreover, we evaluate our attack in black-box settings and show that TransFool has very good transfer properties. Finally, the paper is concluded in Section \ref{Conclusion}.

\section{Related Work} \label{literature}


Adversarial attacks against NMT systems have been studied in recent years.  First, \citet{belinkov2018synthetic} show that character-level  NMT models are highly vulnerable to character manipulations such as typos in a block-box setting. Similarly,  \citet{ebrahimi2018adversarial} investigate the robustness of character-level NMT models. They propose a white-box adversarial attack based on HotFlip \citep{ebrahimi2018hotflip} and greedily change the important characters to decrease the translation quality (untargeted attack) or mute/push a word in the translation (targeted attack). 
On the other hand, many of the adversarial attacks against NMT models are rather based on word replacement. \citet{cheng2019robust} propose a white-box attack where they first select random words of the input sentence and replace them with a similar word. In particular, in order to limit the search space, they find some candidates with the help of a language model and choose the token that aligns best with the gradient of the adversarial loss to cause more damage to the translation.  \citet{michel2019evaluation} and \citet{zhang2021crafting} similarly find important words in the sentence but replace them with a neighbor word in the embedding space to create adversarial examples.  
However, these methods use \textit{heuristic} strategies which are likely to have sub-optimal performance.  
\citet{cheng2020seq2sick} considers another strategy and propose Seq2Sick, a targeted white-box attack against NMT models. They introduce an optimization problem in the embedding space of the NMT model. 
and solve it by gradient projection. Although they have a projection step to the nearest embedding vector, they use the NMT embedding space, which is not effective in capturing the similarity between tokens. 

Other types of attacks against NMT models  with different threat models and purposes have also been investigated in the literature. \citep{wallace2020imitation} propose several  new types of attack: Universal adversarial attacks, which consist of a single snippet of text that can be added to any input sentence to mislead the NMT model, and malicious nonsense, which fools the NMT model to generate malicious translation from nonsense inputs.  
Some papers focus on making NMT models robust to perturbation to the inputs \citep{cheng2018towards,cheng2020advaug,tan2021doubly}. Some other papers use adversarial attacks to enhance the NMT models in some aspects, such as  word sense disambiguation \citep{emelin2020detecting}, robustness to  subword segmentation \citep{park2020adversarial}, and robustness of unsupervised NMT \citep{yu2021a2r2}. In \citep{xu2021targeted,wang2021putting}, the data poisoning attacks against NMT models are studied. Another type of attack whose purpose is to change multiple words while ensuring that the output of the NMT model remains unchanged is explored in \citep{chaturvedi2019exploring,chaturvedi2021ignorance}. Another attack  is presented in \citep{cai2021seeds}, where the adversary uses the hardware faults of systems to fool NMT models. All these attacks show the vulnerability of NMT systems to adversarial attacks in various scenarios different than ours.

In this paper, we focus on the untargeted attack against NMT models, which aims to deceive the NMT model into generating substantially different translations for similar sentences. Such attacks are particularly important since they expose an unexpected behaviour of the NMT systems.   In summary, existing adversarial attacks often use heuristic strategies based on \textit{word-replacement} and are likely to have sub-optimal performance. Or they use the \textit{NMT embedding} space to find similar tokens, which is not effective in capturing the similarity between tokens. Finally, none of these attacks study the \textit{transferability} to black-box settings. To address these issues, we introduce \textit{TransFool}, a method for crafting effective and fluent adversarial sentences that are similar to the original sentences.

\section{Optimization Problem} \label{method}

In this section, we first present our new formulation for generating adversarial examples  against NMT models, along with different terms that form our optimization problem.

\vspace{-3pt}
\paragraph*{Adversarial Attack.} Consider $\mathcal{X}$ to be the source language space and $\mathcal{Y}$ to be the target language space.
The NMT model $f: \mathcal{X}\rightarrow \mathcal{Y}$ generally has an encoder-decoder structure  \citep{bahdanau2015neural,vaswani2017attention} and aims to maximize the translation probability $p(\mathbf{y}_\text{ref}|\mathbf{x})$, where $\mathbf{x} \in \mathcal{X}$ is the input sentence in the source language and $\mathbf{y}_\text{ref} \in \mathcal{Y}$ is the ground-truth translation in the target language. To process textual data, each sentence is decomposed into a sequence of tokens. Therefore, the input sentence  $\mathbf{x}=x_1x_2...x_k$ is split into a sequence of $k$ tokens, where $x_i$ is a token from the vocabulary set $\mathcal{V}_\mathcal{X}$ of the NMT model, which contains all the tokens from the source language.
For each token in the translated sentence $\mathbf{y}_\text{ref} = \mathbf{y}_{\text{ref},1},...,\mathbf{y}_{\text{ref},l}$, the NMT model generates a probability vector over the target language vocabulary set $\mathcal{V}_\mathcal{Y}$ by applying a softmax function to the decoder output. 

The adversary is looking for an adversarial sentence $\mathbf{x'}$, which is tokenized into a sequence of $k$ tokens $\mathbf{x'}=x'_1x'_2...x'_k$, in the source language that fools the target NMT model, i.e., the translation of the adversarial example $f(\mathbf{x'})$ is far from the true translation. However, the adversarial example $\mathbf{x'}$ and the original sentence $\mathbf{x}$ should be similar so that the true translation of the adversarial example stays similar to $\mathbf{y}_\text{ref}$. 

As is common in the NMT models \citep{vaswani2017attention,tang2020multilingual}, 
to feed the discrete sequence of tokens into the NMT model, each token is converted to a continuous vector, known as an  embedding vector, using a lookup table. In particular, let $\text{emb}(.)$ be the embedding function that maps the input token $x_i$ to the continuous embedding vector $\text{emb}(x_i) = \mathbf{e}_i \in \mathbb{R}^m$, where $m$ is the embedding dimension of the target NMT model. Therefore, the input of the NMT model is a sequence of embedding vectors representing the tokens of the input sentence, i.e., $\mathbf{e_x} = [\mathbf{e}_1,\mathbf{e}_2, ...,\mathbf{e}_k] \in \mathbb{R}^{(k\times m)}$. In the same manner, for the adversarial example, we can define $\mathbf{e_{x'}} = [\mathbf{e}'_1,\mathbf{e}'_2, ...,\mathbf{e}'_k] \in \mathbb{R}^{(k\times m)}$. 

To generate an adversarial example for a given input sentence, we introduce an optimization problem with respect to the embedding vectors of the adversarial sentence $\mathbf{e_{x'}}$. Our optimization problem is composed of multiple terms: an adversarial loss, a similarity constraint, and the 
loss of a language model. An adversarial loss causes the target NMT model to generate faulty translation. Moreover, with a language model loss and a similarity constraint, we impose the generated adversarial example to be a fluent sentence and also semantically similar to the original sentence, respectively. The proposed optimization problem, which finds the adversarial example $\mathbf{x'}$ from its embedding representation $\mathbf{e_{x'}}$ by using a lookup table, is defined as follows:
\begin{equation} \label{optimization}
    \mathbf{x'} \leftarrow \argmin_{\substack{\mathbf{e}'_i\in \mathcal{E_{V_{\mathcal{X}}}} }} \;\; [\mathcal{L}_{Adv} + \alpha \mathcal{L}_{Sim} + \beta \mathcal{L}_{LM}], 
\end{equation}
where $\alpha$ and $\beta$ are the hyperparameters that control the relative importance of each term. Moreover, we call the continuous space of the embedding representations the embedding space and denote it by $\mathcal{E}$, and we show the discrete subspace of the embedding space $\mathcal{E}$ containing the embedding representation of every token in the source language vocabulary set by $\mathcal{E_{V_{\mathcal{X}}}}$. 
We now discuss the different terms of the optimization function in detail.

\vspace{-3pt}
\paragraph*{Adversarial Loss.} In order to create an adversarial example whose translation is far away from the reference translation $\mathbf{y}_\text{ref}$, we try to maximize the training loss of the target NMT model. Since the NMT models are trained to generate the next token of the translation given the translation up until that token, we are looking for the adversarial example 
that maximizes the probability of wrong translation (by minimizing the probability of correct translation)  for the $i$-th token, given that the NMT model has produced the correct translation up to step $(i-1)$:
\begin{equation}\label{eq.ladv}
    \mathcal{L}_{Adv} =  \frac{1}{l}\sum_{i=1}^l \log(p_{f}(y_{\text{ref},i}|\mathbf{e_{x'}},\{y_{\text{ref},1},...,y_{\text{ref},(i-1)}\})), 
\end{equation}
where $p_{f}(y_{\text{ref},i}|\mathbf{e_{x'}},\{y_{\text{ref},1},...,y_{\text{ref},(i-1)}\})$ is the cross entropy between the predicted token distribution by the NMT model and the delta distribution on the token $y_{\text{ref},i}$, which is one for the correct translated token, $y_{\text{ref},i}$, and zero otherwise. By minimizing $\log(p_{f}(.))$, normalized by the  sentence length $l$, we force the output probability vector of the NMT  model to differ from the delta distribution on the token $y_{\text{ref},i}$, which may cause the predicted translation to be wrong.  

\vspace{-3pt}
\paragraph*{Similarity Constraint.}
To ensure that the generated adversarial example is similar to the original sentence, we need to add a similarity constraint to our optimization problem. 
It has been shown that the embedding representation of a  language model captures the semantics of the tokens \citep{tenney2019you,shavarani2021better}. 
Suppose that  the embedding representation of the original sentence by a language model  (which may differ from the NMT embedding representation $\mathbf{e_x}$)  is  $\mathbf{v_x} = [\mathbf{v}_1,\mathbf{v}_2, ...,\mathbf{v}_k] \in \mathbb{R}^{(k\times n)}$, where $n$ is the embedding dimension of the LM. Likewise, let $\mathbf{v_{x'}}$ denote the sequence of LM embedding vectors 
regarding the tokens of the adversarial example. 
We can define the distance between the $i$-th tokens of the original and the adversarial sentences by computing the cosine distance  between their corresponding  LM embedding vectors: 
\begin{equation}
    \forall i \in \{1,...,k\}: \quad
    r_i = 1-\frac{\mathbf{v}_i^\intercal\mathbf{v}'_i}{\|\mathbf{v}_i\|_2.\|{\mathbf{v}'_i}\|_2}.
\end{equation}
The cosine distance is zero if the two tokens are the same and it has larger values for two unrelated tokens. We want the adversarial sentence to differ from the original sentence in only a few tokens. Therefore, the cosine distance between most of the tokens in the original and adversarial sentence should be zero, which causes the cosine distance vector $[r_1,r_2,...,r_k]$ to be  sparse. To ensure the sparsity of the cosine distance vector, instead of the $\ell_0$ norm, which is not differentiable, we can define the similarity constraint as the $\ell_1$ norm relaxation of the cosine distance vector normalized to the length of the sentence:
\begin{equation}
    \mathcal{L}_{Sim} = \frac{1}{k}\sum_{i=1}^{k} 1-\frac{\mathbf{v}_i^\intercal\mathbf{v}'_i}{\|\mathbf{v}_i\|_2.\|{\mathbf{v}'_i}\|_2}.
\end{equation}

\vspace{-3pt}
\paragraph*{Language Model Loss.} Causal  language models  are trained to maximize the probability of a token given the previous tokens. Hence, we can use the loss of a CLM, i.e., the negative log-probability, as a rough and differentiable measure for the fluency of the generated adversarial sentence. The loss of a CLM, which is normalized to the sentence length, is as follows:
\begin{equation}\label{eq.lperp}
    \mathcal{L}_{LM} = - \frac{1}{k}\sum_{i=1}^k \log(p_g(\mathbf{v}'_i|\mathbf{v}'_1,...,\mathbf{v}'_{(i-1)})), 
\end{equation}
where g is a  CLM, and  $p_g(\mathbf{v}'_i|\mathbf{v}'_1,...,\mathbf{v}'_{(i-1)})$ is the cross entropy between the predicted token distribution by the language model and the delta distribution on the token $\mathbf{v}'_i$, which is one for the 
corresponding  token in the adversarial example, $\mathbf{v}'_i$, and zero otherwise. 
To generate  adversarial examples against a target NMT model, we propose to solve the optimization problem  (\ref{optimization}), which contains an adversarial loss term, a similarity constraint, and a CLM loss. 


\section{TransFool Attack Algorithm}\label{algorithm}

\begin{wrapfigure}{r}{0.44\textwidth}
  \vspace{-15pt}
  \scalebox{1}{
\begin{minipage}{0.44\textwidth}
\centerline{\includegraphics[page=1,width=1\linewidth, trim={7.8cm 4.1cm 8.3cm 3.8cm},clip]{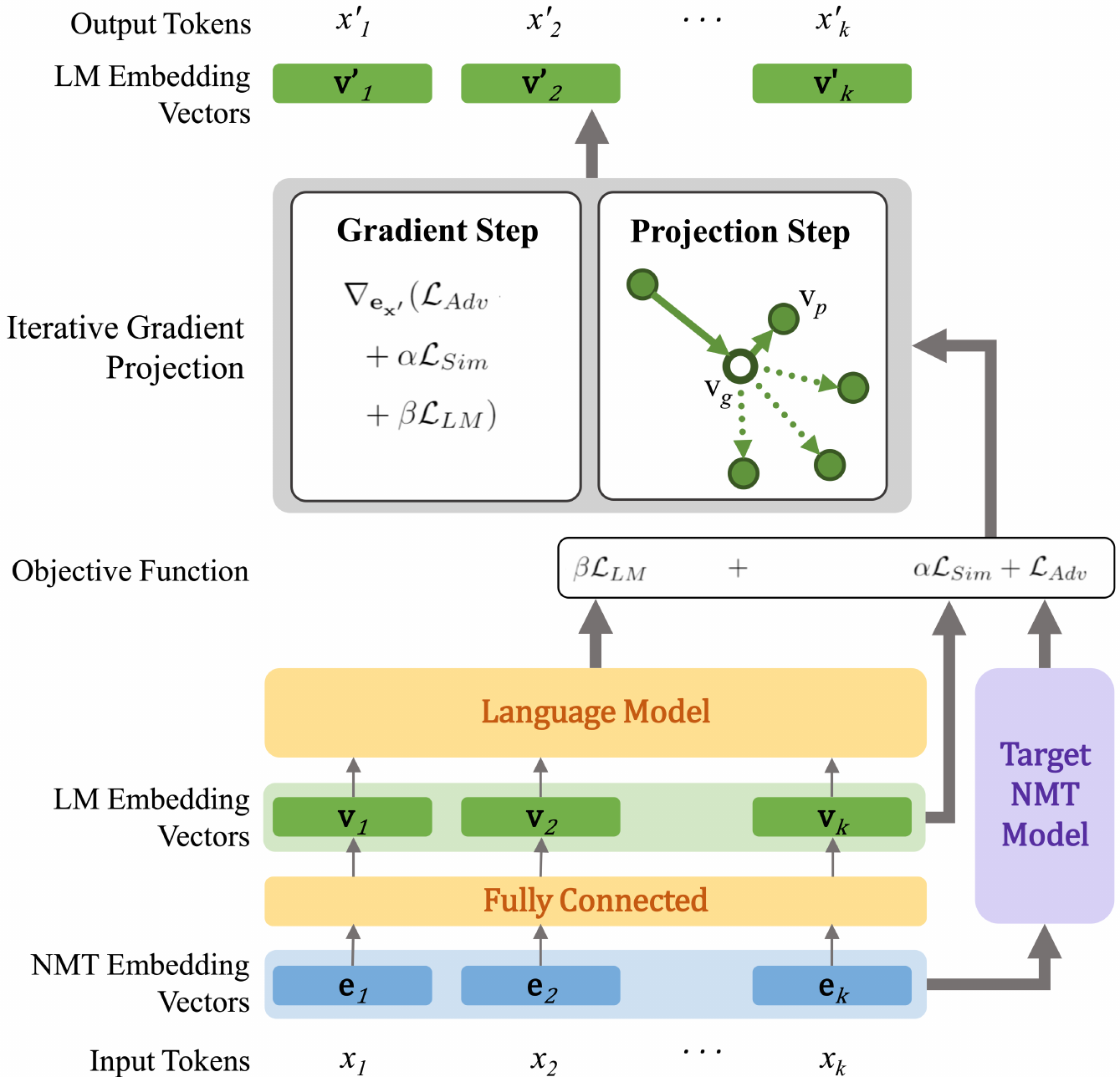}}
\caption{Block diagram of \textit{TransFool}.}
\label{fig:blockdiagram}
\end{minipage}
}


\vspace{7pt}
\scalebox{0.82}{
\begin{minipage}{0.5\textwidth}
\begin{algorithm}[H]

\caption{TransFool Adversarial Attack}
\label{alg}
\small
\begin{algorithmic}
\STATE {\bfseries{Input}:}
\STATE \quad $f(.)$: Target NMT model,  $\mathcal{V_X}$: Vocabulary set
\STATE \quad $FC$: Fully connected layer, $\mathbf{x}$:  Input sentence
\STATE \quad $\mathbf{y_{ref}}$: Ground-truth translation of  $\mathbf{x}$
\STATE \quad  $\lambda$: BLEU score ratio, $\alpha,\beta$: Hyperparameters 
\STATE \quad $K$: Maximum No. of iterations, $\gamma$: step size
\STATE {\bfseries{Output}:}
\STATE \quad $\mathbf{x'}$: Generated adversarial example
\STATE {\bfseries{initialization:}}
\STATE \quad $\forall i \in \{1,...,k\} \quad \mathbf{e}_{\mathbf{g},i},\mathbf{e}_{\mathbf{p},i}\leftarrow \mathbf{e}_i$, $\mathbf{s}\leftarrow\text{empty set}$ 
\STATE \quad  $itr \leftarrow 0$, $thr \leftarrow \text{BLEU}(f(\mathbf{e}_{x}),\mathbf{y_{ref}})) \times \lambda$
\WHILE {$itr < K$}
\STATE $itr \leftarrow itr + 1$
\STATE {\bfseries{Step 1:}} Gradient descent in the continuous 
\STATE embedding space:
\STATE $\mathbf{e_g} \leftarrow \mathbf{e}_{\mathbf{g}} - \gamma. \nabla_{\mathbf{e_{x'}}} (\mathcal{L}_{adv} + \alpha \mathcal{L}_{Sim} + \beta \mathcal{L}_{LM})$
\STATE $\mathbf{v_g} \leftarrow FC(\mathbf{e_g})$
\STATE {\bfseries{Step 2:}} Projection  to the discrete subspace 
\STATE $\mathcal{E_{V_X}}$  and update if the sentence is new: 
\FOR {$i \in \{1,...,k\}$} 
\STATE $\mathbf{e}_{\mathbf{p},i} \leftarrow \argmax\limits_{\mathbf{e} \in \mathcal{E_{V_X}}} \frac{FC(\mathbf{e})^\top \mathbf{v}_{\mathbf{g},i}}{\|FC(\mathbf{e})\|_2.\|\mathbf{v}_{\mathbf{g},i}\|_2}$
\ENDFOR
\IF {$\mathbf{e_p}$ not in set $\mathbf{s}$} 
\STATE add $\mathbf{e_p}$ to set $\mathbf{s}$
\STATE $\mathbf{e_g}\leftarrow \mathbf{e}_{\mathbf{p}}$
\IF {$\text{BLEU}(f(\mathbf{e}_{\mathbf{p}}),\mathbf{y_{ref}}))\leq thr$}
\STATE break (adversarial example is found)
\ENDIF
\ENDIF

\ENDWHILE
\STATE {\bfseries return} ${\mathbf{e_{x'}}} \leftarrow \mathbf{e_p}$

\end{algorithmic}
\end{algorithm}
\end{minipage}
}
\vspace{-20pt}
\end{wrapfigure}

We now introduce our algorithm for generating adversarial examples against NMT models. The block diagram of our proposed attack is presented in Figure \ref{fig:blockdiagram}. We are looking for an adversarial example with tokens in the vocabulary set $\mathcal{V_X}$ and the corresponding embedding vectors in the subspace $\mathcal{E_{V_X}}$. Hence, the optimization problem  (\ref{optimization}) is discrete. The high-level idea of our algorithm is to use gradient projection to solve \eqref{optimization} in the discrete  subspace $\mathcal{E_{V_X}}$. 

The objective function of \eqref{optimization} is a function of NMT and LM embedding representations of the adversarial example, $\mathbf{e_{x'}}$ and $\mathbf{v_{x'}}$, respectively.  Since we aim to minimize the optimization problem with respect to $\mathbf{e_{x'}}$, we need to find a transformation between the embedding space of the LM and the target NMT model.  To this aim, as depicted in Figure \ref{fig:blockdiagram}, we propose to replace the embedding layer of a pre-trained language model with a Fully Connected (FC) layer, which gets the embedding vectors of the NMT model as its input. Then, we train the language model and  the FC layer simultaneously with the causal language modeling objective. Therefore, we can compute the LM embedding vectors as a function of the NMT embedding vectors: 
$\mathbf{v}_i = FC(\mathbf{e}_i)$, where $FC \in \mathbb{R}^{m \times n}$ is the trained FC layer.

The pseudo-code of our attack can be found in Algorithm \ref{alg}. In more detail, we first convert the discrete tokens of the sentence to continuous embedding vectors of the target NMT model, then we use the FC layer to compute the embedding representations of the tokens by the language model. Afterwards,  we consider the continuous relaxation of the optimization problem, which means that we assume that the embedding vectors are in the continuous embedding space $\mathcal{E}$ instead of $\mathcal{E_{V_X}}$. In each iteration of the algorithm, we first update the sequence of embedding vectors $\mathbf{e_{x'}}$ in the opposite direction of the gradient (gradient descent). Let us denote the output of the gradient descent step for the $i$-th token by $\mathbf{e}_{\mathbf{g},i}$. Then we project the resultant embedding vectors, which are not necessarily in $\mathcal{E_{V_X}}$, to the nearest token in the vocabulary set $\mathcal{V_X}$. Since the distance in the embedding space of the language model approximates the similarity between the tokens, we use the LM embedding representations with cosine similarity metric in the projection step to find the most similar token in the vocabulary. We can apply the trained fully connected layer $FC$ to find the LM embedding representations: $\mathbf{v_g}= FC(\mathbf{e_g})$. Hence, the projected NMT embedding vector, $ \mathbf{e}_{\mathbf{p},i}$, for the $i$-th token is:
\begin{equation}
\setlength\belowdisplayskip{5pt}
\setlength\abovedisplayskip{7pt}
    \mathbf{e}_{\mathbf{p},i} =  \argmax\limits_{\mathbf{e} \in \mathcal{E_{V_X}}} \frac{FC(\mathbf{e})^\top \mathbf{v}_{\mathbf{g},i}}{\|FC(\mathbf{e})\|_2.\|\mathbf{v}_{\mathbf{g},i}\|_2}.
\end{equation}
However, due to the discrete nature of data,  by applying the projection step in every iteration of the algorithm, we may face an undesirable situation where the algorithm gets stuck  in a loop of previously computed steps. In order to circumvent this issue, we will only update the embedding vectors by the output of the projection step if the projected sentence has not been generated before. 

We perform the gradient descent and projection steps iteratively until a maximum number of iterations is reached, or the translation quality of the adversarial example relative to the original translation quality is less than a threshold. To evaluate the translation quality, we use the BLEU score, which is a widely used metric in the literature:
\begin{equation} \label{sar}
\setlength\belowdisplayskip{0pt}
\setlength\abovedisplayskip{7pt}
    \frac{\text{BLEU}(f(\mathbf{e}_{\mathbf{x'}}),\mathbf{y_{ref}}))}{\text{BLEU}(f(\mathbf{e}_{\mathbf{x}}),\mathbf{y_{ref}}))}\le\lambda.
\end{equation}

\section{Experiments} \label{Experiments}
\vspace{-1pt}
In this section, we first discuss our experimental setup, and then we evaluate  TransFool against different models and translation tasks, both in white-box and black-box settings.\footnote{Our source code is available at \url{https://github.com/sssadrizadeh/TransFool}. Appendix \ref{license} also contains the license information and details of the assets (datasets, codes, and models).} 

\vspace{-1pt}
\subsection{Experimental Setup}
\vspace{-1pt}
We conduct experiments on the English-French (En-Fr), English-German (En-De), and English-Chinese (En-Zh) translation tasks. We use the test set of  WMT14  \citep{bojar2014findings} for  En-Fr and En-De tasks, and the test set of  OPUS-100   \citep{zhang2020improving}  for  En-Zh task. Some statistics of these datasets are presented in  Appendix \ref{dataset}. 

We evaluate TransFool against transformer-based NMT models. To verify that our attack is effective against various architectures, we attack the HuggingFace implementation of Marian NMT models \citep{junczys-dowmunt-etal-2018-marian} and  mBART50 multilingual NMT model \citep{tang2020multilingual}.

As explained in Section \ref{algorithm}, the similarity constraint and the LM loss of the proposed optimization problem require an FC layer and a CLM. To this aim, for each NMT model,  we train an FC layer and a CLM (with GPT-2 structure \citep{radford2019language}) on WikiText-103 dataset. We note that the input of the FC layer is the target NMT embedding representation of the input sentence.

To find the minimizer of our optimization problem (\ref{optimization}), we use the Adam optimizer \citep{kingma2014adam} with step size $\gamma=0.016$. Moreover, we set the maximum number of iterations to 500. 
Our algorithm has three parameters: coefficients $\alpha$ and $\beta$ in the  optimization function (\ref{optimization}), and the relative BLEU score ratio $\lambda$ in the stopping criteria (\ref{sar}). We set $\lambda=0.4$, $\beta=1.8$, and $\alpha=20$. We chose these parameters experimentally according to the  ablation study available in Appendix \ref{abl:hyper}, to optimize the performance in terms of success rate, semantic similarity, and fluency.

We compare our attack with \citep{michel2019evaluation}, which is a white-box untargeted attack against NMT models.\footnote{Source codes of {\citep{cheng2019robust,cheng2020advaug}}, other  
untargeted white-box  attacks against NMTs
,  are not publicly available.} We only consider one of their attacks, called \textit{kNN}, which substitutes some words with their neighbors in the  embedding space; the other attack considers swapping the characters, which is too easy to detect. 
We also adapted \textit{Seq2Sick} \citep{cheng2020seq2sick},  a targeted attack against NMT models, which is based on an optimization problem in the NMT embedding space, to our untargeted setting. 

For evaluation, we report different performance metrics: \textbf{(1) Attack Success  Rate (ASR)}, which measures the rate of successful adversarial examples. Similar to \citep{ebrahimi2018adversarial}, we define the adversarial example as successful if the  BLEU score of its translation is \textit{less than half} of the BLEU score of the original translation. \textbf{(2) Relative decrease of translation quality}, by measuring the translation quality in terms of  \textit{BLEU score}\footnote{We use case-sensitive SacreBLEU 
on detokenized sentences.} and \textit{chrF} \citep{popovic2015chrf}. We denote these two metrics by \textbf{RDBLEU} and \textbf{RDchrF}, respectively. We choose to compute the \textit{relative decrease} in translation quality so that scores are comparable across different models and datasets  \citep{michel2019evaluation}. \textbf{(3) Semantic Similarity (Sim.)}, which is computed between the original and  adversarial sentences and commonly approximated by the \textit{Universal Sentence Encoder (USE)} \citep{yang2020multilingual}. USE model is trained on  various NLP tasks to embed sentences into high-dimensional vectors. The cosine similarity between USE embeddings of two sentence approximates their semantic similarity.    
\textbf{(4) Perplexity score (Perp.)}, which is a measure of the fluency of the adversarial example computed by the perplexity score of \textit{GPT-2 (large)}. 
 \textbf{(5) Token Error Rate (TER)}, which measures 
 the percentage of tokens that are modified by an adversarial attack.


\subsection{Performance in White-box Settings}
Now we evaluate TransFool in comparison to kNN and Seq2Sick against different NMT models. Table \ref{tab:white} shows the results in terms of different evaluation metrics.\footnote{
Since  our attack is  token-level, there is a small chance that, when the adversarial example is converted to text, re-tokenization does not produce the same set of tokens. Thus, all  results are computed after re-tokenization of the adversarial examples.} Overall, our attack is able to decrease the BLEU score of the target model to less than half of the BLEU score of the original translation for more than 60\% of the sentences for all tasks and  models (except for the En-Zh mBART50 model, where  ASR is 57.50\%). Also, in all cases, semantic similarity  is more than 0.83, which shows that our attack can maintain a high level of semantic similarity with the clean sentences.

\begin{table*}[t]
	\centering
		\renewcommand{\arraystretch}{0.95}
	\setlength{\tabcolsep}{4pt}
	\caption{Performance of white-box attack against different NMT models
	.}
	\vspace{-5pt}
	\label{tab:white}
	\scalebox{0.75}{
		\begin{tabular}[t]{@{} lcccccccccccccc @{}}
			\toprule[1pt]
		    \multirow{2}{*}{\textbf{Task}}  &
		    \multirow{2}{*}{\textbf{Method}}  & \multicolumn{6}{c}{\textbf{Marian NMT}} &&   \multicolumn{6}{c}{\textbf{mBART50}} \\
			\cline{3-8}
			\cline{10-15}
			\rule{0pt}{2.5ex}    
			& & \scalebox{0.95}{ASR$\uparrow$} & \scalebox{0.95}{RDBLEU$\uparrow$} & \scalebox{0.95}{RDchrF$\uparrow$} & \scalebox{0.95}{Sim.$\uparrow$} & \scalebox{0.95}{Perp.$\downarrow$} & \scalebox{0.95}{TER$\downarrow$} && \scalebox{0.95}{ASR$\uparrow$} & \scalebox{0.95}{RDBLEU$\uparrow$} & \scalebox{0.95}{RDchrF$\uparrow$} & \scalebox{0.95}{Sim.$\uparrow$} & \scalebox{0.95}{Perp.$\downarrow$} & \scalebox{0.95}{TER$\downarrow$}\\
			\midrule[1pt]
			\multirow{3}{*}{En-Fr} & TransFool &  \textbf{69.38} & \textbf{0.57} & \textbf{0.23} & \textbf{0.85} &  \underline{182.45} & \textbf{13.91} && \textbf{60.68} & \textbf{0.53} & \textbf{0.22} & \underline{0.84} & \textbf{121.12} & \textbf{10.58} \\ 
			& kNN &   \underline{36.53} & \underline{0.36} & \underline{0.16} & \underline{0.82} & 389.78 & 19.15 && \underline{30.84} & \underline{0.29} & 0.11 & \textbf{0.85} &  336.47 & 21.03 \\
			& Seq2Sick & 27.01 & 0.21 & \underline{0.16} & 0.75 & \textbf{175.31} & \underline{13.97} && 25.53 & 0.19 & \underline{0.13} & 0.75 &  \underline{151.92} & \underline{13.55} \\
			\midrule[1pt]
			\multirow{3}{*}{En-De} & TransFool &  \textbf{69.49} & \textbf{0.65} & \textbf{0.23} & \textbf{0.84} & \textbf{165.53} & \textbf{13.57} && \textbf{62.87} & \textbf{0.61} & \textbf{0.22} & \underline{0.83} & \textbf{134.90} & \textbf{11.07}\\ 
			& kNN &   \underline{39.22} & \underline{0.40} & 0.17 & \underline{0.82} & 441.62 & 19.42 && \underline{35.99} & \underline{0.39} & 0.12 & \textbf{0.86} &  375.32 & 21.22 \\
			& Seq2Sick &  35.60 & 0.31 & \underline{0.21} & 0.67 & \underline{290.32} & \underline{18.13} && 35.59 & 0.31 & \underline{0.20} & 0.66 &  \underline{265.62} & \underline{18.18}\\
			\midrule[1pt]
			\multirow{3}{*}{En-Zh} & TransFool &  \textbf{73.82} & \textbf{0.74} & \textbf{0.31} & \textbf{0.88} & \textbf{102.49} & \textbf{11.82} && \textbf{57.50} & \textbf{0.67} & \textbf{0.26} & \textbf{0.90} & \textbf{74.75} & \textbf{7.77}\\ 
			& kNN &   \underline{31.12} & \underline{0.33} & 0.18 & \underline{0.86} & 180.27 & 15.95 && \underline{27.25} & \underline{0.32} & 0.14 & \textbf{0.90} & 160.27 & 16.58 \\
			& Seq2Sick & 28.76 & 0.26 & \underline{0.25} & 0.73 & \underline{161.84} & \underline{17.48} && 24.25 & 0.31 & \underline{0.18} & {0.78} & \underline{105.42} & \underline{13.58} \\
			
			\bottomrule[1pt]
		\end{tabular}
	}
	\vspace{-10pt}
\end{table*}

In comparison to the baselines, TransFool obtains a higher  success  rate against different model structures and translation tasks, and it is able to reduce the translation quality more severely. Since the algorithm uses the gradients of the proposed optimization problem and is not based on token replacement, TransFool can highly degrade the translation quality. 
Furthermore, the perplexity score of the adversarial example generated by TransFool is much less than the ones of both baselines (except for the En-Fr Marian model, where it is a little higher than Seq2Sick), which is due to the integration of the LM embeddings and the LM loss term in the optimization problem.
Moreover, the token error rate of our attack is lower than both baselines, and the semantic similarity is preserved better by TransFool in almost all cases since we use the LM embeddings instead of the NMT ones 
for the similarity constraint. 
While kNN can also maintain similarity, Seq2Sick does not perform well in this criterion. We also computed similarity by BERTScore \citep{zhang2019bertscore} and BLEURT-20 \citep{sellam2020learning} that highly correlate with human judgments  
in Appendix \ref{sim_white}, which shows that TransFool is better than both baselines in maintaining the semantics.  We also show the generalizability of  TransFool against other translation tasks in Appendix \ref{other-lang}. Finally, as presented in Appendix \ref{success_white}, the \textit{successful} attacks by the baselines, as opposed to TransFool, are not semantic-preserving or fluent.  

We also compare the runtimes of TransFool and both baselines. In each iteration of our proposed attack,  we need to perform a back-propagation through the target  model and the language model to compute the gradients. Also, in some iterations (27 iterations per sentence on average), a forward pass is required to compute the output of the target model to check the stopping criteria. For the Marian NMT (En-Fr) model, on a system equipped with an NVIDIA A100 GPU, it takes 26.45 seconds to generate adversarial examples by TransFool. On the same system, kNN needs 1.45 seconds, and Seq2Sick needs 38.85 seconds to generate adversarial examples for less effective adversarial attacks, however.

Table \ref{tab:sample}  shows
an adversarial example against mBART50 (En-De).  
In comparison to the baselines, TransFool makes smaller changes to the sentence, and the adversarial example is a correct English sentence similar to the original one. However, kNN and Seq2Sick
generate adversarial sentences that are not necessarily natural or similar to the original ones. 
More examples  by TransFool, kNN, and Seq2Sick can be found in Appendix \ref{samples_white}. We also provide some adversarial sentences when we do not use the LM embeddings in our algorithm  to show the importance of this component.

Indeed, TransFool outperforms both baselines in terms of success rate. It is able to generate more natural adversarial examples with a lower number of perturbations (TER) and higher semantic similarity with the clean samples in almost all cases. A complete ablation study of the effect of hyperparameters, the language model used in TransFool, and the beam-size parameter of the target NMT model 
is presented in Appendix \ref{abl:hyper}, \ref{abl:lm}, and \ref{abl:beam}, respectively. 

\begin{table}[t]
	\centering
		\renewcommand{\arraystretch}{.85}

	\setlength{\tabcolsep}{2pt}
	\caption{Adversarial examples against Marian NMT (En-Fr) by various methods (white-box).}\vspace{-2pt}
	\scalebox{0.73}{
		\begin{tabular}[t]{@{} l| >{\parfillskip=0pt}p{16cm} @{}}
			\toprule[1pt]
		    \textbf{Sentence}  & 
      \textbf{Text}\\
		
			\midrule[1pt]
			
		    \multirow{1}{*}{Org.} &  
      The most eager is \textcolor{blue}{\textbf{Oregon}}, which is enlisting 5,000 drivers in the country's biggest experiment. \hfill\mbox{}  \\
			 
			\cline{2-2}
			\rule{0pt}{2.5ex} 
			
			\multirow{2}{*}{Ref. Trans.} & 
   Le plus déterminé est l'Oregon, qui a mobilisé 5 000 conducteurs pour mener l'expérience la plus importante du pays. \hfill\mbox{}  \\
			
			\cline{2-2}
			\rule{0pt}{2.5ex} 
			
			\multirow{1}{*}{Org. Trans.} &  
   Le plus avide est l'Oregon, qui recrute 5 000 pilotes dans la plus grande expérience du pays. \hfill\mbox{}  \\
			 
			\cline{1-2}
			\rule{0pt}{2.5ex}

			\multirow{1}{*}{Adv. TransFool} &  
   The most eager is\textcolor{red}{\textbf{Quebec}}, which is enlisting 5,000 drivers in the country's biggest experiment. \hfill\mbox{} \\
			
			\cline{2-2}
			\rule{0pt}{2.5ex} 
			
			\multirow{2}{*}{Trans.} &  
   Le \textcolor{Brown}{Québec}, qui \textcolor{orange}{fait partie de} la plus grande expérience du pays, \textcolor{orange}{compte} 5 000 pilotes. \textcolor{orange}{(\textit{some parts are not translated.})} \hfill\mbox{} \\

			\cline{1-2}
			\rule{0pt}{2.5ex} 
			
			\multirow{1}{*}{Adv. kNN} &  
   \textcolor{red}{\textbf{Theve}} eager is Oregon, \textcolor{red}{\textbf{C aren}} enlisting 5,000 drivers in \textcolor{red}{\textbf{theau}}'s biggest experiment. \hfill\mbox{} \\
			
			\cline{2-2}
			\rule{0pt}{2.5ex} 
			
			\multirow{1}{*}{Trans.} &  
   \textcolor{Brown}{Theve} avide est Oregon, \textcolor{Brown}{C sont} \textcolor{orange}{enrôlés} 5 000 pilotes dans la plus grande expérience de \textcolor{Brown}{Theau}. \hfill\mbox{} \\
			
			\cline{1-2}
			\rule{0pt}{2.5ex} 
			
			\multirow{1}{*}{Adv. Seq2Sick} &  
   The most \textcolor{red}{\textbf{buzz}} is \textcolor{red}{\textbf{FREE}}, which is \textcolor{red}{\textbf{chooseing Games comments}} in the country's great \textcolor{red}{\textbf{developer}}. \hfill\mbox{} \\
			
			\cline{2-2}
			\rule{0pt}{2.5ex} 
			
			\multirow{1}{*}{Trans.} &  
   Le plus \textcolor{Brown}{buzz est GRATUIT}, qui \textcolor{Brown}{est de choisir Jeux commentaires} dans le grand \textcolor{Brown}{développeur} du pays. \hfill\mbox{} \\

			\bottomrule[1pt]
			\multicolumn{2}{l}{{\rule{0pt}{2ex}\footnotesize  
			$^*$Perturbed tokens are in \textcolor{red}{\textbf{red}}, \kern0.2em and in the original sentence, the perturbations by TransFool are in \textcolor{blue}{\textbf{blue}}. \kern-0.2em The changes in the translation that}}\\
			
                \multicolumn{2}{l}{{\footnotesize   are the direct result of the  perturbations are in \textcolor{Brown}{brown}, while the changes that are due to the failure of the target model are in \textcolor{orange}{orange}.}} \\

		\end{tabular}
	}
	\vspace{-10pt}
	\label{tab:sample}
	
\end{table}

\vspace{-5pt}
\subsection{Performance in Black-box Settings}\label{exp:blackbox}

In practice, the adversary's access to the learning system may be limited. Hence, we propose to analyze the performance of TransFool in a black-box scenario.  It has been shown that  adversarial attacks often transfer to  another model that has a different architecture and is even trained with different datasets \citep{szegedy2014intriguing}. By utilizing this property of adversarial attacks, we extend TransFool to the black-box scenario. We consider that we have complete access to one NMT model (the reference model), including its gradients. We implement the proposed gradient-based attack in algorithm \ref{alg} with this model. However, for the stopping criteria of the algorithm, we query the black-box target NMT model to compute the BLEU score. 
We can also implement the black-box transfer attack in the case where the source languages of the reference model and the target model are the same, but their target languages are different. Since  Marian NMT is faster and lighter than  mBART50, we use it as the reference model and evaluate the performance of the  black-box 
attack against mBART50. 

\begin{wrapfigure}{R}{0.56\textwidth}
\vspace{-25pt}
\begin{small}
\begin{minipage}{0.56\textwidth}
\begin{table}[H]
	\centering
		\renewcommand{\arraystretch}{1}
	\setlength{\tabcolsep}{3pt}
	\caption{Performance of  black-box attack against \scalebox{0.93}{mBART50}
	.}
	\vspace{-5pt}
	\label{tab:black}
	\scalebox{0.71}{
		\begin{tabular}[t]{@{} lcccccccc @{}}
			\toprule[1pt]
		    \multirow{1}{*}{\textbf{Task}}  &
		    \multirow{1}{*}{\textbf{Method}}  & 
		    \scalebox{0.95}{ASR$\uparrow$} & \scalebox{0.95}{RDBLEU$\uparrow$} & \scalebox{0.95}{RDchrF$\uparrow$} & \scalebox{0.95}{Sim.$\uparrow$} & \scalebox{0.95}{Perp.$\downarrow$} & \scalebox{0.95}{TER$\downarrow$} &
		    \scalebox{0.95}{\#Queries$\downarrow$}   \\
			\midrule[1pt]
			\multirow{4}{*}{En-Fr} & TransFool &  \textbf{70.19} & \textbf{0.58} & \underline{0.22} & \textbf{0.85} & \underline{175.39} & \textbf{17.08} & 27 \\ 
			& kNN & 33.74 & 0.33 & 0.15 & 0.82 & 383.71 & 22.57 & - \\
			& Seq2Sick & 25.97 & 0.21 & 0.14 & 0.75 &  \textbf{173.63} & \underline{21.13} &- \\
			& WSLS &   \underline{56.21} & \textbf{0.58} & \textbf{0.27} & \underline{0.84} & 214.23 & 31.30 & 1423 \\
			\midrule[1pt]
			\multirow{4}{*}{En-De} & TransFool &   \textbf{66.76} & \textbf{0.65} & \textbf{0.22} & \underline{0.84} & \textbf{167.54} & \textbf{16.73} & 23\\ 
			& kNN &   36.70 & 0.39 & 0.16 & 0.82 & 435.02 & \underline{22.34} & -\\
			& Seq2Sick & 32.17 & 0.29 & \underline{0.20} & 0.67 & 286.67 & 26.59 & -\\
			& WSLS &   \underline{44.33} & \underline{0.50} & 0.19 & \textbf{0.86} & \underline{219.32} & 29.12 & 1262\\
			\midrule[1pt]
			\multirow{4}{*}{En-Zh} & TransFool &  \textbf{63.27} & \underline{0.71} & \underline{0.27} & \textbf{0.88} & \textbf{100.14} & \textbf{14.76} & 36 \\ 
			& kNN & 26.89 & 0.31 & 0.17 & \underline{0.86} & 176.34 & \underline{17.07} & -\\
			& Seq2Sick & 23.65 & 0.30 & 0.23 & 0.73 & \underline{162.67} & 25.17 & -\\
			& WSLS &   \underline{40.00} & \textbf{0.72} & \textbf{0.52} & 0.83 & 186.44 & 32.35 & 1782\\
			
			\bottomrule[1pt]
		\end{tabular}
	}
	\vspace{-2pt} 
\end{table}
\end{minipage}
\end{small}
\vspace{-10pt}
\end{wrapfigure}
We compare the performance of TransFool with WSLS \citep{zhang2021crafting}, a black-box untargeted attack against NMT models based on word-replacement 
(the choice of back-translation model used in WSLS is investigated in Appendix \ref{back_trans}). We also evaluate the performance of kNN and Seq2Sick in the black-box settings by attacking mBART50 with the adversarial example generated against Marian NMT (in the white-box settings). The results are reported in Table \ref{tab:black}. We also report the performance when attacking Google Translate, some generated adversarial samples, and similarity performance computed by BERTScore and BLEURT-20 in Appendix \ref{black}.

Table \ref{tab:black} shows that in all tasks, with a few queries to the target model, our black-box attack achieves better performance than the white-box attack against the target model (mBART50) but a little worse performance than the white-box attack against the reference model (Marian NMT). 
In all cases, the success rate, token error rate, and perplexity of TransFool are  better than all baselines (except for the En-Fr task, where perplexity is a little higher than Seq2Sick). 
The ability of TransFool and WSLS to maintain semantic similarity is comparable and better than both other baselines. However, WSLS has the highest token error rate, which makes the attack detectable.   The effect of TransFool on BLEU score is larger than that of the other methods, and its effect on chrF  comes after WSLS (except for the En-DE task, where  TransFool is the best). 

Regarding the complexity, TransFool requires only a few queries to the target model for translation, while WSLS queries the model more than a thousand times, which is costly and may not be feasible in practice. For the En-Fr  task, on a system equipped with an NVIDIA A100 GPU, it takes 43.36 and 1904.98 seconds to generate adversarial examples by TransFool and WSLS, respectively, which shows that WSLS is very time-consuming. 

We also analyze the \textit{cross-lingual} transferability of the generated adversarial examples to a black-box NMT model with the same source language but a different target language. Since we need a dataset with the same set of sentences for different language pairs, we use the validation set of  WMT14 for En-Fr and En-De tasks. Table \ref{tab:2lang} shows the results for two cases: Marian NMT or mMBART50 as the target  model. 
We use  Marian NMT as the reference model with a different target language than that of the target model. In all settings, the generated adversarial examples are highly transferable to another NMT model with a different target language (i.e., they have high attack success rate and large semantic similarity). 

\begin{table*}[t]
	\centering
		\renewcommand{\arraystretch}{1}
	\setlength{\tabcolsep}{3.5pt}
	\caption{Performance of  black-box attack, when the target language is different
	.}
	\vspace{-5pt}
	\label{tab:2lang}
	\scalebox{0.74}{
		\begin{tabular}[t]{@{} lccccccccccccc @{}}
			\toprule[1pt]
		    \multirow{2}{*}{\textbf{Task}}  &
		    \multicolumn{6}{c}{\textbf{Marian NMT}} &&   \multicolumn{6}{c}{\textbf{mBART50}} \\
			\cline{2-7}
			\cline{9-14}
			\rule{0pt}{2.5ex}    
		    & \scalebox{0.95}{ASR$\uparrow$} & \scalebox{0.95}{RDBLEU$\uparrow$} & \scalebox{0.95}{RDchrF$\uparrow$} & \scalebox{0.95}{Sim.$\uparrow$} & \scalebox{0.95}{Perp.$\downarrow$} & 
		    \scalebox{0.95}{\#Queries$\downarrow$}  && \scalebox{0.95}{ASR$\uparrow$} & \scalebox{0.95}{RDBLEU$\uparrow$} & \scalebox{0.95}{RDchrF$\uparrow$} & \scalebox{0.95}{Sim.$\uparrow$} & \scalebox{0.95}{Perp.$\downarrow$} & 
		    \scalebox{0.95}{\#Queries$\downarrow$}  \\
			\midrule[1pt]
	
			\multirow{1}{*}{En-De $\rightarrow$ En-Fr} & 60.53 & 0.55 & 0.22 & 0.84 & 169.49 & 24 && 61.68 & 0.56 & 0.22 & 0.84 &  169.51 & 23\\ 
			\midrule[1pt]
			\multirow{1}{*}{En-Fr $\rightarrow$ En-De} &  66.22 & 0.63 & 0.22 & 0.84 & 198.04 &  23 && 63.86 & 0.63 & 0.21 & 0.84 & 195.50 & 24 \\ 
			
			\bottomrule[1pt]
		\end{tabular}
	}
\end{table*}
To the best of our knowledge this type of transferability have not been studied before. Moreover, the high transferability of TransFool, even to other languages,  shows that it is able to capture the common failure modes in different  NMT models, which can be dangerous in real-world applications.

\vspace{-5pt}
\subsection{Discussion}

\subsubsection{Comparison to Related Works} \label{LM}

Unlike several word-replacement-based methods \citep{michel2019evaluation,li2020bert,jin2020bert}, TransFool does not explicitly select a few important tokens in the beginning of the attack algorithm. Instead, all tokens are modified during the gradient step of the TransFool algorithm. However, in the projection step, most tokens are projected back to the original ones, while a few are replaced with the closest tokens in the embedding space. Not limiting the search space from the beginning and using a gradient-based approach lead TransFool to achieve a high success rate.

\begin{wrapfigure}{R}{0.55\textwidth}
\vspace{-25pt}
\begin{small}
\begin{minipage}{0.55\textwidth}
\begin{table}[H]
	\centering
		\renewcommand{\arraystretch}{1}
	\setlength{\tabcolsep}{4pt}
	\caption{Performance of white-box attack against Marian NMT (En-Fr) with/without language model embeddings. }
	\label{tab:LMemb}
	\scalebox{0.8}{
		\begin{tabular}[t]{@{} lccccc @{}}
			\toprule[1pt]
		    \multirow{1}{*}{\textbf{Method}}  &
		    \scalebox{0.95}{ASR$\uparrow$} & \scalebox{0.95}{RDBLEU$\uparrow$} & \scalebox{0.95}{RDchrF$\uparrow$} & \scalebox{0.95}{Sim.$\uparrow$} & \scalebox{0.95}{Perp.$\downarrow$}\\
			\midrule[1pt]
			\multirow{1}{*}{TransFool w/ LM Emb.} &   69.48 & 0.56 & 0.23 & 0.85 & 177.20   \\ 
			\multirow{1}{*}{TransFool w/ NMT Emb.} &   68.27 & 0.57 & 0.26 & 0.78 & 193.32 \\  
			\midrule[1pt]
			kNN w/ LM Emb.  &   32.13 & 0.32 & 0.15 & 0.85 & 246.52 \\
			kNN w/ NMT Emb. &   36.65 & 0.35 & 0.16 & 0.82 & 375.84 \\
			
			\bottomrule[1pt]
		\end{tabular}
	}
\end{table}

\end{minipage}
\end{small}
\vspace{-10pt}
\end{wrapfigure}
On another note, it is challenging to incorporate linguistic constraints in a differentiable manner with our optimization-based method. TransFool solves this challenge by finding a transformation between the embedding representations of the NMT model and that of the language model. This is particularly important, as we see in Table \ref{tab:LMemb}. This Table shows the results of TransFool and kNN when we use LM embeddings or NMT embeddings for measuring the similarity between two tokens.\footnote{In order to have a fair comparison, we fine-tuned hyperparameters of Transfool, in the case when we do not use LM embeddings, to have a similar  attack success rate.} The LM embeddings result in lower perplexity and higher semantic similarity for both methods, which demonstrates the importance of this component of the TransFool algorithm. 
As a matter of fact, the main difference between TransFool and Seq2Sick is employing the LM embeddingsinstead of the NMT ones, which results in lower perplexity and higher similarity for TransFool compared to Seq2Sick. We should note that more ablation studies on the LM part of our attack are available in Appendix \ref{abl:lm}. 

\subsubsection{Translation Quality Metric}
\begin{wrapfigure}{R}{0.6\textwidth}
\vspace{-25pt}
\begin{small}
\begin{minipage}{0.6\textwidth}
\begin{table}[H]
	\centering
		\renewcommand{\arraystretch}{1}
	\setlength{\tabcolsep}{4pt}
	\caption{Performance of white-box attack against Marian NMT (En-Fr) with BLEURT-20 as translation metric. }
	\label{tab:bleurt}
	\scalebox{0.8}{
		\begin{tabular}[t]{@{} lcccccc @{}}
			\toprule[1pt]
		    \multirow{1}{*}{\textbf{Metric}}  &
		    \scalebox{0.95}{ASR$\uparrow$} & \scalebox{0.95}{RDBLEURT$\uparrow$} & \scalebox{0.95}{Sim. USE$\uparrow$} & \scalebox{0.95}{Sim. BLUERT$\uparrow$} & \scalebox{0.95}{Perp.$\downarrow$} & \scalebox{0.95}{TER$\downarrow$}\\
			\midrule[1pt]
			\multirow{1}{*}{BLEURT-20} &   72.60 & 0.56 & 0.85 & 0.64 & 186.55 &  13.17  \\ 

			\bottomrule[1pt]
		\end{tabular}
	}
\end{table}

\end{minipage}
\end{small}
\vspace{-10pt}
\end{wrapfigure}
In TransFool, we have used the BLEU score to evaluate the translation quality for the success criterion. We chose the BLEU score since it has been used in previous works \citep{cheng2019robust,cheng2020advaug,wallace2020imitation,zhang2021crafting}, is still common in benchmarks, and is fast. However,  any other metric can be considered for evaluating the translation quality in the attack algorithm. We report the performance of TransFool against Marain NMT using BLEURT-20 \citep{sellam2020learning}, a recent metric for translation quality, in Table \ref{tab:bleurt}. We should note that the success rate is measured based on BLEURT-20, and the relative decrease in translation quality in terms of this metric is denoted by RDBLEURT. The results indicate that other metrics, such as BLEURT-20, can yield similar results to those of the BLEU score. We can also use the same metric to evaluate the semantic similarity in the source language and the translation quality in the target language to make interpreting the evaluation results easier. Therefore, we report semantic similarity in terms of both USE and BLUERT-20 in this Table. The translation quality, in terms of BLEURT-20, has dropped from 0.73 to 0.32, while the similarity in the source language remains 0.64.

\subsubsection{Effect of TransFool on NMT Performance through Back-Translation}
\begin{wrapfigure}{r}{0.6\textwidth}
  \vspace{-20pt}
  \scalebox{1}{
\begin{minipage}{0.6\textwidth}
\centerline{\includegraphics[page=1,width=1\linewidth, trim={3.6cm 1.5cm 3.1cm 0cm},clip]{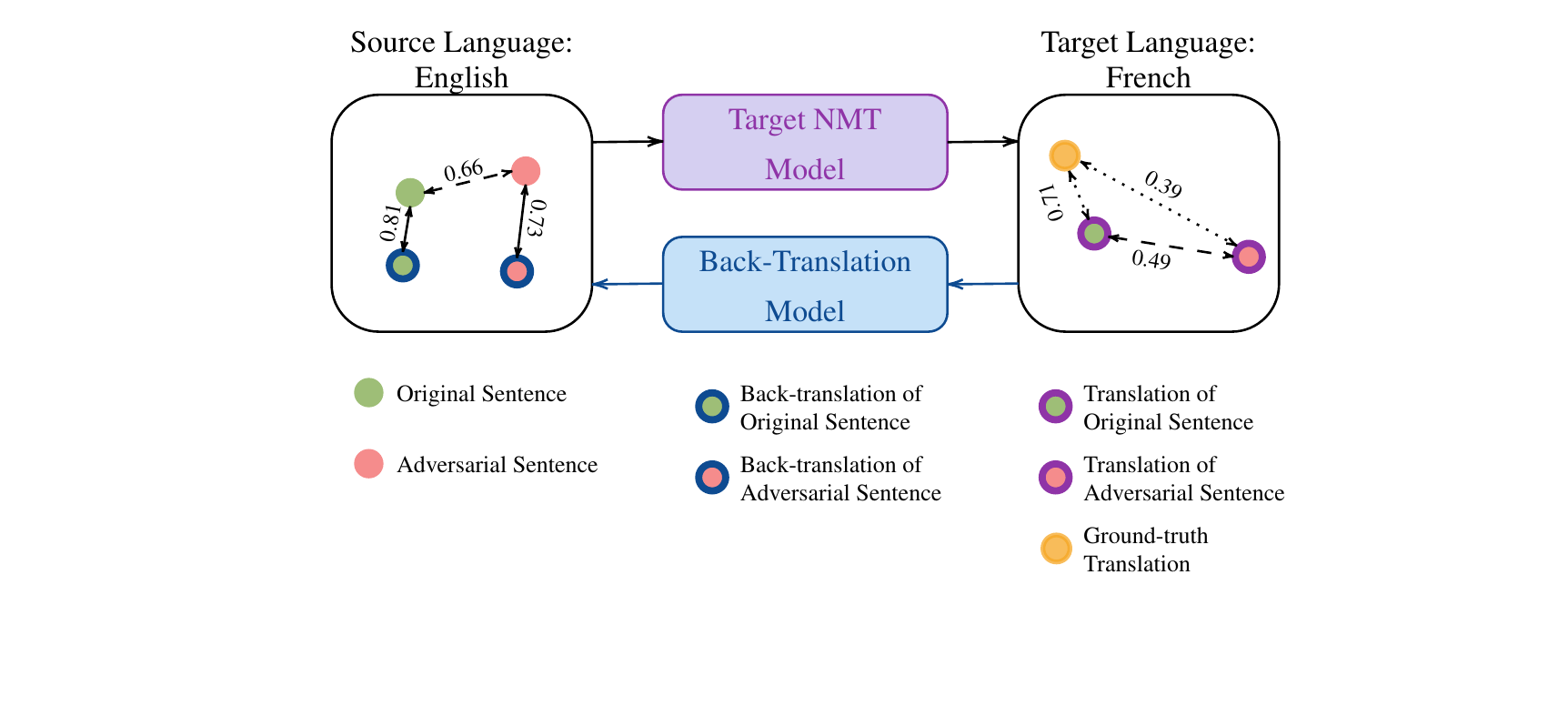}}
\caption{Similarity by BLEURT-20 between different pairs in case of attacking Marian NMT (En-Fr).}
\label{fig:sim_diag}
\end{minipage}
}
\vspace{-10pt}
\end{wrapfigure}
Given the nature of the translation task, we can observe that some adversarial perturbations appear directly in the translation of the adversarial examples. However, other changes to the translation are caused by the degradation in the performance of the NMT model resulting from the attack. For example in Table \ref{tab:sample}, changing Oregon to Quebec results in two types of changes in the translation: type 1) \textit{"l'Oregon"} is replaced with \textit{"le Québec"}, and type 2) some part of the input sentence, \textit{"The most eager is"}, is not translated. In order to evaluate if the model is truly fooled by TransFool (i.e., type 2 of the changes), we consider the Marian NMT (En-Fr) as the target model and  use a back translation model to do a round-trip translation (En-Fr and Fr-En). If the target NMT model (En-Fr) is performing well, we expect that the English sentence and its back-translated counterpart to have a high similarity. We measure this similarity, in terms of BLEURT-20, for the original and adversarial sentences. A lower similarity score for the adversarial sentences indicates that the model's performance is degraded by the attack. Moreover, we can measure the similarity between the original and adversarial sentences in the source language (En) and the similarity between their translations in the target language (Fr). The results averaged across the dataset are depicted in Figure \ref{fig:sim_diag}. There are two key observations:

\begin{enumerate}
    \item The round-trip translation quality for the original sentences is higher than that of the adversarial sentences (0.81 vs. 0.73). This finding suggests that the attack successfully degraded the model's performance, and not all of the changes to the translation are of type one, i.e., the direct results of the perturbations.
    \item The similarity between the original and adversarial sentences in the source domain (En) is 0.66, while the similarity between their translations in the target domain (Fr) is 0.49. This unexpected discrepancy again suggests that the attack successfully degraded the model's performance.
\end{enumerate}

Therefore, these two observations confirm that TransFool is successful in fooling the NMT model.

\subsubsection{Human Evaluation}
We conduct a human evaluation campaign to further evaluate the adversarial attacks against Marian NMT (En-Fr) in the white-box setting. Specifically, we assess TransFool, kNN, and Seq2Sick attacks based on three criteria: fluency, semantic preservation, and translation quality. Appendix \ref{hu} provides more details about our setup, while the results of each survey along with their 95\% confidence intervals are reported. 

Our findings demonstrate that the adversarial examples generated by TransFool are more semantic-preserving and fluent than both baselines. We also conduct a separate survey only for the adversarial examples generated by TransFool, where the annotators are asked to rate the accuracy of the reference translations and the  NMT translations. The results show that the NMT translation has a lower score than the reference translations. We also report the  Inter-Annotator Agreement (IAA) between the human judgments in Appendix \ref{hu}. Overall, our human evaluation results demonstrate the superiority of TransFool over the baselines.


\section{Conclusion} \label{Conclusion}
In this paper, we proposed \textit{TransFool}, a white-box adversarial attack against NMT models, by introducing a new optimization problem solved by an iterative method based on gradient projection. 
We utilized the embedding representation of a language model to impose a similarity constraint on the adversarial examples. Moreover, by considering the loss of an LM in our optimization problem, the generated adversarial examples are more fluent. 
Extensive 
automatic and human evaluations 
show that  TransFool is highly effective in different translation tasks and against different NMT models.  Our attack is also transferable to black-box settings with different structures and even different target languages. In both white-box and black-box scenarios, TransFool obtains  improvement over the baselines in terms of  success rate, semantic similarity, and fluency. TransFool demonstrate that the translations of two similar sentences (original and adversarial ones) by the NMT model can differ significantly, and even small perturbations such as a name change can degrade the translation quality. These adversarial attacks against NMT models are critical to analyze the vulnerabilities of NMT models, to measure their robustness, and to eventually build more robust NMT models.

\section*{Acknowledgments}
This work has been partially supported by armasuisse Science and Technology project MULAN. We would also like to thank the anonymous reviewers whose valuable comments improved the paper.

\bibliography{main}

\begin{thebibliography}{59}
\providecommand{\natexlab}[1]{#1}
\providecommand{\url}[1]{\texttt{#1}}
\expandafter\ifx\csname urlstyle\endcsname\relax
  \providecommand{\doi}[1]{doi: #1}\else
  \providecommand{\doi}{doi: \begingroup \urlstyle{rm}\Url}\fi

\bibitem[Alzantot et~al.(2018)Alzantot, Sharma, Elgohary, Ho, Srivastava, and
  Chang]{alzantot2018generating}
Moustafa Alzantot, Yash Sharma, Ahmed Elgohary, Bo-Jhang Ho, Mani Srivastava,
  and Kai-Wei Chang.
\newblock Generating natural language adversarial examples.
\newblock In \emph{Proceedings of the 2018 Conference on Empirical Methods in
  Natural Language Processing}, pp.\  2890--2896, 2018.

\bibitem[Bahdanau et~al.(2015)Bahdanau, Cho, and Bengio]{bahdanau2015neural}
Dzmitry Bahdanau, Kyung~Hyun Cho, and Yoshua Bengio.
\newblock Neural machine translation by jointly learning to align and
  translate.
\newblock In \emph{3rd International Conference on Learning Representations,
  ICLR 2015}, 2015.

\bibitem[Belinkov \& Bisk(2018)Belinkov and Bisk]{belinkov2018synthetic}
Yonatan Belinkov and Yonatan Bisk.
\newblock Synthetic and natural noise both break neural machine translation.
\newblock In \emph{International Conference on Learning Representations}, 2018.

\bibitem[Bojar et~al.(2014)Bojar, Buck, Federmann, Haddow, Koehn, Leveling,
  Monz, Pecina, Post, Saint-Amand, et~al.]{bojar2014findings}
Ond{\v{r}}ej Bojar, Christian Buck, Christian Federmann, Barry Haddow, Philipp
  Koehn, Johannes Leveling, Christof Monz, Pavel Pecina, Matt Post, Herve
  Saint-Amand, et~al.
\newblock Findings of the 2014 workshop on statistical machine translation.
\newblock In \emph{Proceedings of the ninth workshop on statistical machine
  translation}, pp.\  12--58, 2014.

\bibitem[Cai et~al.(2021)Cai, Chowdhuryy, Zhang, and Yao]{cai2021seeds}
Kunbei Cai, Md~Hafizul~Islam Chowdhuryy, Zhenkai Zhang, and Fan Yao.
\newblock Seeds of seed: Nmt-stroke: Diverting neural machine translation
  through hardware-based faults.
\newblock In \emph{2021 International Symposium on Secure and Private Execution
  Environment Design (SEED)}, pp.\  76--82. IEEE, 2021.

\bibitem[Cer et~al.(2017)Cer, Diab, Agirre, Lopez-Gazpio, and
  Specia]{cer2017semeval}
Daniel Cer, Mona Diab, Eneko Agirre, I{\~n}igo Lopez-Gazpio, and Lucia Specia.
\newblock Semeval-2017 task 1: Semantic textual similarity multilingual and
  crosslingual focused evaluation.
\newblock In \emph{Proceedings of the 11th International Workshop on Semantic
  Evaluation (SemEval-2017)}, pp.\  1--14, 2017.

\bibitem[Chaturvedi et~al.(2019)Chaturvedi, KP, and
  Garain]{chaturvedi2019exploring}
Akshay Chaturvedi, Abijith KP, and Utpal Garain.
\newblock Exploring the robustness of nmt systems to nonsensical inputs.
\newblock \emph{arXiv preprint arXiv:1908.01165}, 2019.

\bibitem[Chaturvedi et~al.(2021)Chaturvedi, Chakrabarty, Utiyama, Sumita, and
  Garain]{chaturvedi2021ignorance}
Akshay Chaturvedi, Abhisek Chakrabarty, Masao Utiyama, Eiichiro Sumita, and
  Utpal Garain.
\newblock Ignorance is bliss: Exploring defenses against invariance-based
  attacks on neural machine translation systems.
\newblock \emph{IEEE Transactions on Artificial Intelligence}, 2021.

\bibitem[Cheng et~al.(2020{\natexlab{a}})Cheng, Yi, Chen, Zhang, and
  Hsieh]{cheng2020seq2sick}
Minhao Cheng, Jinfeng Yi, Pin-Yu Chen, Huan Zhang, and Cho-Jui Hsieh.
\newblock Seq2sick: Evaluating the robustness of sequence-to-sequence models
  with adversarial examples.
\newblock In \emph{Proceedings of the AAAI Conference on Artificial
  Intelligence}, volume~34, pp.\  3601--3608, 2020{\natexlab{a}}.

\bibitem[Cheng et~al.(2018)Cheng, Tu, Meng, Zhai, and Liu]{cheng2018towards}
Yong Cheng, Zhaopeng Tu, Fandong Meng, Junjie Zhai, and Yang Liu.
\newblock Towards robust neural machine translation.
\newblock In \emph{Proceedings of the 56th Annual Meeting of the Association
  for Computational Linguistics (Volume 1: Long Papers)}, pp.\  1756--1766,
  2018.

\bibitem[Cheng et~al.(2019)Cheng, Jiang, and Macherey]{cheng2019robust}
Yong Cheng, Lu~Jiang, and Wolfgang Macherey.
\newblock Robust neural machine translation with doubly adversarial inputs.
\newblock In \emph{Proceedings of the 57th Annual Meeting of the Association
  for Computational Linguistics}, pp.\  4324--4333, 2019.

\bibitem[Cheng et~al.(2020{\natexlab{b}})Cheng, Jiang, Macherey, and
  Eisenstein]{cheng2020advaug}
Yong Cheng, Lu~Jiang, Wolfgang Macherey, and Jacob Eisenstein.
\newblock Advaug: Robust adversarial augmentation for neural machine
  translation.
\newblock In \emph{Proceedings of the 58th Annual Meeting of the Association
  for Computational Linguistics}, pp.\  5961--5970, 2020{\natexlab{b}}.

\bibitem[Ebrahimi et~al.(2018{\natexlab{a}})Ebrahimi, Lowd, and
  Dou]{ebrahimi2018adversarial}
Javid Ebrahimi, Daniel Lowd, and Dejing Dou.
\newblock On adversarial examples for character-level neural machine
  translation.
\newblock In \emph{Proceedings of the 27th International Conference on
  Computational Linguistics}, pp.\  653--663, 2018{\natexlab{a}}.

\bibitem[Ebrahimi et~al.(2018{\natexlab{b}})Ebrahimi, Rao, Lowd, and
  Dou]{ebrahimi2018hotflip}
Javid Ebrahimi, Anyi Rao, Daniel Lowd, and Dejing Dou.
\newblock Hotflip: White-box adversarial examples for text classification.
\newblock In \emph{Proceedings of the 56th Annual Meeting of the Association
  for Computational Linguistics (Volume 2: Short Papers)}, pp.\  31--36,
  2018{\natexlab{b}}.

\bibitem[Emelin et~al.(2020)Emelin, Titov, and Sennrich]{emelin2020detecting}
Denis Emelin, Ivan Titov, and Rico Sennrich.
\newblock Detecting word sense disambiguation biases in machine translation for
  model-agnostic adversarial attacks.
\newblock In \emph{The 2020 Conference on Empirical Methods in Natural Language
  Processing}, pp.\  7635--7653. Association for Computational Linguistics,
  2020.

\bibitem[Freitag et~al.(2021)Freitag, Rei, Mathur, Lo, Stewart, Foster, Lavie,
  and Bojar]{freitag2021results}
Markus Freitag, Ricardo Rei, Nitika Mathur, Chi-kiu Lo, Craig Stewart, George
  Foster, Alon Lavie, and Ond{\v{r}}ej Bojar.
\newblock Results of the wmt21 metrics shared task: Evaluating metrics with
  expert-based human evaluations on ted and news domain.
\newblock In \emph{Proceedings of the Sixth Conference on Machine Translation},
  pp.\  733--774, 2021.

\bibitem[Graham et~al.(2013)Graham, Baldwin, Moffat, and
  Zobel]{graham2013continuous}
Yvette Graham, Timothy Baldwin, Alistair Moffat, and Justin Zobel.
\newblock Continuous measurement scales in human evaluation of machine
  translation.
\newblock In \emph{Proceedings of the 7th Linguistic Annotation Workshop and
  Interoperability with Discourse}, pp.\  33--41, 2013.

\bibitem[Graham et~al.(2017)Graham, Baldwin, Moffat, and Zobel]{graham2017can}
Yvette Graham, Timothy Baldwin, Alistair Moffat, and Justin Zobel.
\newblock Can machine translation systems be evaluated by the crowd alone.
\newblock \emph{Natural Language Engineering}, 23\penalty0 (1):\penalty0 3--30,
  2017.

\bibitem[Guo et~al.(2021)Guo, Sablayrolles, J{\'e}gou, and
  Kiela]{guo2021gradient}
Chuan Guo, Alexandre Sablayrolles, Herv{\'e} J{\'e}gou, and Douwe Kiela.
\newblock Gradient-based adversarial attacks against text transformers.
\newblock In \emph{Proceedings of the 2021 Conference on Empirical Methods in
  Natural Language Processing}, pp.\  5747--5757, 2021.

\bibitem[He et~al.(2016)He, Zhang, Ren, and Sun]{he2016deep}
Kaiming He, Xiangyu Zhang, Shaoqing Ren, and Jian Sun.
\newblock Deep residual learning for image recognition.
\newblock In \emph{Proceedings of the IEEE conference on computer vision and
  pattern recognition}, pp.\  770--778, 2016.

\bibitem[Jin et~al.(2020)Jin, Jin, Zhou, and Szolovits]{jin2020bert}
Di~Jin, Zhijing Jin, Joey~Tianyi Zhou, and Peter Szolovits.
\newblock Is bert really robust? a strong baseline for natural language attack
  on text classification and entailment.
\newblock In \emph{Proceedings of the AAAI conference on artificial
  intelligence}, volume~34, pp.\  8018--8025, 2020.

\bibitem[Junczys-Dowmunt et~al.(2018)Junczys-Dowmunt, Grundkiewicz, Dwojak,
  Hoang, Heafield, Neckermann, Seide, Germann, Aji, Bogoychev, Martins, and
  Birch]{junczys-dowmunt-etal-2018-marian}
Marcin Junczys-Dowmunt, Roman Grundkiewicz, Tomasz Dwojak, Hieu Hoang, Kenneth
  Heafield, Tom Neckermann, Frank Seide, Ulrich Germann, Alham~Fikri Aji,
  Nikolay Bogoychev, Andr{\'e} F.~T. Martins, and Alexandra Birch.
\newblock {M}arian: Fast neural machine translation in {C}++.
\newblock In \emph{Proceedings of {ACL} 2018, System Demonstrations}, pp.\
  116--121, Melbourne, Australia, July 2018.

\bibitem[Kingma \& Ba(2014)Kingma and Ba]{kingma2014adam}
Diederik~P Kingma and Jimmy Ba.
\newblock Adam: A method for stochastic optimization.
\newblock \emph{arXiv preprint arXiv:1412.6980}, 2014.

\bibitem[Lhoest et~al.(2021)Lhoest, Villanova~del Moral, Jernite, Thakur, von
  Platen, Patil, Chaumond, Drame, Plu, Tunstall, Davison, {\v{S}}a{\v{s}}ko,
  Chhablani, Malik, Brandeis, Le~Scao, Sanh, Xu, Patry, McMillan-Major, Schmid,
  Gugger, Delangue, Matussi{\`e}re, Debut, Bekman, Cistac, Goehringer, Mustar,
  Lagunas, Rush, and Wolf]{lhoest-etal-2021-datasets}
Quentin Lhoest, Albert Villanova~del Moral, Yacine Jernite, Abhishek Thakur,
  Patrick von Platen, Suraj Patil, Julien Chaumond, Mariama Drame, Julien Plu,
  Lewis Tunstall, Joe Davison, Mario {\v{S}}a{\v{s}}ko, Gunjan Chhablani,
  Bhavitvya Malik, Simon Brandeis, Teven Le~Scao, Victor Sanh, Canwen Xu,
  Nicolas Patry, Angelina McMillan-Major, Philipp Schmid, Sylvain Gugger,
  Cl{\'e}ment Delangue, Th{\'e}o Matussi{\`e}re, Lysandre Debut, Stas Bekman,
  Pierric Cistac, Thibault Goehringer, Victor Mustar, Fran{\c{c}}ois Lagunas,
  Alexander Rush, and Thomas Wolf.
\newblock Datasets: A community library for natural language processing.
\newblock In \emph{Proceedings of the 2021 Conference on Empirical Methods in
  Natural Language Processing: System Demonstrations}, pp.\  175--184, Online
  and Punta Cana, Dominican Republic, November 2021. Association for
  Computational Linguistics.
\newblock URL \url{https://aclanthology.org/2021.emnlp-demo.21}.

\bibitem[Li et~al.(2020)Li, Ma, Guo, Xue, and Qiu]{li2020bert}
Linyang Li, Ruotian Ma, Qipeng Guo, Xiangyang Xue, and Xipeng Qiu.
\newblock Bert-attack: Adversarial attack against bert using bert.
\newblock In \emph{Proceedings of the 2020 Conference on Empirical Methods in
  Natural Language Processing (EMNLP)}, pp.\  6193--6202, 2020.

\bibitem[Ma et~al.(2018)Ma, Bojar, and Graham]{ma2018results}
Qingsong Ma, Ond{\v{r}}ej Bojar, and Yvette Graham.
\newblock Results of the wmt18 metrics shared task: Both characters and
  embeddings achieve good performance.
\newblock In \emph{Proceedings of the third conference on machine translation:
  shared task papers}, pp.\  671--688, 2018.

\bibitem[Madry et~al.(2018)Madry, Makelov, Schmidt, Tsipras, and
  Vladu]{madry2018towards}
Aleksander Madry, Aleksandar Makelov, Ludwig Schmidt, Dimitris Tsipras, and
  Adrian Vladu.
\newblock Towards deep learning models resistant to adversarial attacks.
\newblock In \emph{International Conference on Learning Representations, ICLR
  2018}, 2018.

\bibitem[Michel et~al.(2019)Michel, Li, Neubig, and Pino]{michel2019evaluation}
Paul Michel, Xian Li, Graham Neubig, and Juan Pino.
\newblock On evaluation of adversarial perturbations for sequence-to-sequence
  models.
\newblock In \emph{Proceedings of the 2019 Conference of the North American
  Chapter of the Association for Computational Linguistics: Human Language
  Technologies, Volume 1 (Long and Short Papers)}, pp.\  3103--3114, 2019.

\bibitem[Moosavi-Dezfooli et~al.(2016)Moosavi-Dezfooli, Fawzi, and
  Frossard]{moosavi2016deepfool}
Seyed-Mohsen Moosavi-Dezfooli, Alhussein Fawzi, and Pascal Frossard.
\newblock Deepfool: a simple and accurate method to fool deep neural networks.
\newblock In \emph{Proceedings of the IEEE conference on computer vision and
  pattern recognition}, pp.\  2574--2582, 2016.

\bibitem[Morris et~al.(2020)Morris, Lifland, Lanchantin, Ji, and
  Qi]{morris2020reevaluating}
John Morris, Eli Lifland, Jack Lanchantin, Yangfeng Ji, and Yanjun Qi.
\newblock Reevaluating adversarial examples in natural language.
\newblock In \emph{Findings of the Association for Computational Linguistics:
  EMNLP 2020}, pp.\  3829--3839, 2020.

\bibitem[Naber et~al.(2003)]{naber2003rule}
Daniel Naber et~al.
\newblock A rule-based style and grammar checker.
\newblock 2003.

\bibitem[Ng et~al.(2019)Ng, Yee, Baevski, Ott, Auli, and
  Edunov]{ng2019facebook}
Nathan Ng, Kyra Yee, Alexei Baevski, Myle Ott, Michael Auli, and Sergey Edunov.
\newblock Facebook fair’s wmt19 news translation task submission.
\newblock In \emph{Proceedings of the Fourth Conference on Machine Translation
  (Volume 2: Shared Task Papers, Day 1)}, pp.\  314--319, 2019.

\bibitem[Ortiz-Jim{\'e}nez et~al.(2021)Ortiz-Jim{\'e}nez, Modas,
  Moosavi-Dezfooli, and Frossard]{ortiz2021optimism}
Guillermo Ortiz-Jim{\'e}nez, Apostolos Modas, Seyed-Mohsen Moosavi-Dezfooli,
  and Pascal Frossard.
\newblock Optimism in the face of adversity: Understanding and improving deep
  learning through adversarial robustness.
\newblock \emph{Proceedings of the IEEE}, 109\penalty0 (5):\penalty0 635--659,
  2021.

\bibitem[Park et~al.(2020)Park, Sung, Lee, and Kang]{park2020adversarial}
Jungsoo Park, Mujeen Sung, Jinhyuk Lee, and Jaewoo Kang.
\newblock Adversarial subword regularization for robust neural machine
  translation.
\newblock In \emph{Findings of the Association for Computational Linguistics,
  ACL 2020: EMNLP 2020}, pp.\  1945--1953. Association for Computational
  Linguistics (ACL), 2020.

\bibitem[Paszke et~al.(2019)Paszke, Gross, Massa, Lerer, Bradbury, Chanan,
  Killeen, Lin, Gimelshein, Antiga, et~al.]{paszke2019pytorch}
Adam Paszke, Sam Gross, Francisco Massa, Adam Lerer, James Bradbury, Gregory
  Chanan, Trevor Killeen, Zeming Lin, Natalia Gimelshein, Luca Antiga, et~al.
\newblock Pytorch: An imperative style, high-performance deep learning library.
\newblock \emph{Advances in neural information processing systems}, 32, 2019.

\bibitem[Popovi{\'c}(2015)]{popovic2015chrf}
Maja Popovi{\'c}.
\newblock chrf: character n-gram f-score for automatic mt evaluation.
\newblock In \emph{Proceedings of the Tenth Workshop on Statistical Machine
  Translation}, pp.\  392--395, 2015.

\bibitem[Radford et~al.(2019)Radford, Wu, Child, Luan, Amodei, Sutskever,
  et~al.]{radford2019language}
Alec Radford, Jeffrey Wu, Rewon Child, David Luan, Dario Amodei, Ilya
  Sutskever, et~al.
\newblock Language models are unsupervised multitask learners.
\newblock \emph{OpenAI blog}, 1\penalty0 (8):\penalty0 9, 2019.

\bibitem[Ren et~al.(2019)Ren, Deng, He, and Che]{ren2019generating}
Shuhuai Ren, Yihe Deng, Kun He, and Wanxiang Che.
\newblock Generating natural language adversarial examples through probability
  weighted word saliency.
\newblock In \emph{Proceedings of the 57th annual meeting of the association
  for computational linguistics}, pp.\  1085--1097, 2019.

\bibitem[Sadrizadeh et~al.(2022)Sadrizadeh, Dolamic, and
  Frossard]{sadrizadeh2022block}
Sahar Sadrizadeh, Ljiljana Dolamic, and Pascal Frossard.
\newblock Block-sparse adversarial attack to fool transformer-based text
  classifiers.
\newblock In \emph{ICASSP 2022-2022 IEEE International Conference on Acoustics,
  Speech and Signal Processing (ICASSP)}, pp.\  7837--7841. IEEE, 2022.

\bibitem[Sellam et~al.(2020)Sellam, Pu, Chung, Gehrmann, Tan, Freitag, Das, and
  Parikh]{sellam2020learning}
Thibault Sellam, Amy Pu, Hyung~Won Chung, Sebastian Gehrmann, Qijun Tan, Markus
  Freitag, Dipanjan Das, and Ankur Parikh.
\newblock Learning to evaluate translation beyond english: Bleurt submissions
  to the wmt metrics 2020 shared task.
\newblock In \emph{Proceedings of the Fifth Conference on Machine Translation},
  pp.\  921--927, 2020.

\bibitem[Shavarani \& Sarkar(2021)Shavarani and Sarkar]{shavarani2021better}
Hassan~S Shavarani and Anoop Sarkar.
\newblock Better neural machine translation by extracting linguistic
  information from bert.
\newblock In \emph{Proceedings of the 16th Conference of the European Chapter
  of the Association for Computational Linguistics: Main Volume}, pp.\
  2772--2783, 2021.

\bibitem[Szegedy et~al.(2014)Szegedy, Zaremba, Sutskever, Bruna, Erhan,
  Goodfellow, and Fergus]{szegedy2014intriguing}
Christian Szegedy, Wojciech Zaremba, Ilya Sutskever, Joan Bruna, Dumitru Erhan,
  Ian Goodfellow, and Rob Fergus.
\newblock Intriguing properties of neural networks.
\newblock In \emph{2nd International Conference on Learning Representations,
  ICLR 2014}, 2014.

\bibitem[Tan et~al.(2021)Tan, Ding, Khayrallah, and Koehn]{tan2021doubly}
Weiting Tan, Shuoyang Ding, Huda Khayrallah, and Philipp Koehn.
\newblock Doubly-trained adversarial data augmentation for neural machine
  translation.
\newblock \emph{arXiv e-prints}, pp.\  arXiv--2110, 2021.

\bibitem[Tang et~al.(2020)Tang, Tran, Li, Chen, Goyal, Chaudhary, Gu, and
  Fan]{tang2020multilingual}
Yuqing Tang, Chau Tran, Xian Li, Peng-Jen Chen, Naman Goyal, Vishrav Chaudhary,
  Jiatao Gu, and Angela Fan.
\newblock Multilingual translation with extensible multilingual pretraining and
  finetuning.
\newblock \emph{arXiv preprint arXiv:2008.00401}, 2020.

\bibitem[Tenney et~al.(2019)Tenney, Xia, Chen, Wang, Poliak, McCoy, Kim,
  Van~Durme, Bowman, Das, et~al.]{tenney2019you}
Ian Tenney, Patrick Xia, Berlin Chen, Alex Wang, Adam Poliak, R~Thomas McCoy,
  Najoung Kim, Benjamin Van~Durme, Samuel~R Bowman, Dipanjan Das, et~al.
\newblock What do you learn from context? probing for sentence structure in
  contextualized word representations.
\newblock In \emph{7th International Conference on Learning Representations,
  ICLR 2019}, 2019.

\bibitem[Tiedemann(2012)]{tiedemann2012parallel}
J{\"o}rg Tiedemann.
\newblock Parallel data, tools and interfaces in opus.
\newblock In \emph{Eight International Conference on Language Resources and
  Evaluation, MAY 21-27, 2012, Istanbul, Turkey}, pp.\  2214--2218, 2012.

\bibitem[Vaswani et~al.(2017)Vaswani, Shazeer, Parmar, Uszkoreit, Jones, Gomez,
  Kaiser, and Polosukhin]{vaswani2017attention}
Ashish Vaswani, Noam Shazeer, Niki Parmar, Jakob Uszkoreit, Llion Jones,
  Aidan~N Gomez, {\L}ukasz Kaiser, and Illia Polosukhin.
\newblock Attention is all you need.
\newblock In \emph{Advances in neural information processing systems}, pp.\
  5998--6008, 2017.

\bibitem[Vieira et~al.(2021)Vieira, O’Hagan, and
  O’Sullivan]{vieira2021understanding}
Lucas~Nunes Vieira, Minako O’Hagan, and Carol O’Sullivan.
\newblock Understanding the societal impacts of machine translation: a critical
  review of the literature on medical and legal use cases.
\newblock \emph{Information, Communication \& Society}, 24\penalty0
  (11):\penalty0 1515--1532, 2021.

\bibitem[Wallace et~al.(2020)Wallace, Stern, and Song]{wallace2020imitation}
Eric Wallace, Mitchell Stern, and Dawn Song.
\newblock Imitation attacks and defenses for black-box machine translation
  systems.
\newblock In \emph{Proceedings of the 2020 Conference on Empirical Methods in
  Natural Language Processing (EMNLP)}, pp.\  5531--5546, 2020.

\bibitem[Wang et~al.(2022)Wang, Xu, Liu, Cheng, and Li]{wang2022semattack}
Boxin Wang, Chejian Xu, Xiangyu Liu, Yu~Cheng, and Bo~Li.
\newblock Semattack: Natural textual attacks via different semantic spaces.
\newblock In \emph{Findings of the Association for Computational Linguistics:
  NAACL 2022}, pp.\  176--205, 2022.

\bibitem[Wang et~al.(2021)Wang, Xu, Guzm{\'a}n, El-Kishky, Tang, Rubinstein,
  and Cohn]{wang2021putting}
Jun Wang, Chang Xu, Francisco Guzm{\'a}n, Ahmed El-Kishky, Yuqing Tang,
  Benjamin Rubinstein, and Trevor Cohn.
\newblock Putting words into the system’s mouth: A targeted attack on neural
  machine translation using monolingual data poisoning.
\newblock In \emph{Findings of the Association for Computational Linguistics:
  ACL-IJCNLP 2021}, pp.\  1463--1473, 2021.

\bibitem[Wolf et~al.(2020)Wolf, Debut, Sanh, Chaumond, Delangue, Moi, Cistac,
  Rault, Louf, Funtowicz, Davison, Shleifer, von Platen, Ma, Jernite, Plu, Xu,
  Scao, Gugger, Drame, Lhoest, and Rush]{wolf-etal-2020-transformers}
Thomas Wolf, Lysandre Debut, Victor Sanh, Julien Chaumond, Clement Delangue,
  Anthony Moi, Pierric Cistac, Tim Rault, Rémi Louf, Morgan Funtowicz, Joe
  Davison, Sam Shleifer, Patrick von Platen, Clara Ma, Yacine Jernite, Julien
  Plu, Canwen Xu, Teven~Le Scao, Sylvain Gugger, Mariama Drame, Quentin Lhoest,
  and Alexander~M. Rush.
\newblock Transformers: State-of-the-art natural language processing.
\newblock In \emph{Proceedings of the 2020 Conference on Empirical Methods in
  Natural Language Processing: System Demonstrations}, pp.\  38--45, Online,
  October 2020. Association for Computational Linguistics.
\newblock URL \url{https://www.aclweb.org/anthology/2020.emnlp-demos.6}.

\bibitem[Xu et~al.(2021)Xu, Wang, Tang, Guzm{\'a}n, Rubinstein, and
  Cohn]{xu2021targeted}
Chang Xu, Jun Wang, Yuqing Tang, Francisco Guzm{\'a}n, Benjamin~IP Rubinstein,
  and Trevor Cohn.
\newblock A targeted attack on black-box neural machine translation with
  parallel data poisoning.
\newblock In \emph{Proceedings of the Web Conference 2021}, pp.\  3638--3650,
  2021.

\bibitem[Yang et~al.(2020)Yang, Cer, Ahmad, Guo, Law, Constant, Abrego, Yuan,
  Tar, Sung, et~al.]{yang2020multilingual}
Yinfei Yang, Daniel Cer, Amin Ahmad, Mandy Guo, Jax Law, Noah Constant,
  Gustavo~Hernandez Abrego, Steve Yuan, Chris Tar, Yun-Hsuan Sung, et~al.
\newblock Multilingual universal sentence encoder for semantic retrieval.
\newblock In \emph{Proceedings of the 58th Annual Meeting of the Association
  for Computational Linguistics: System Demonstrations}, pp.\  87--94, 2020.

\bibitem[Yu et~al.(2021)Yu, Luo, Yi, and Cheng]{yu2021a2r2}
Heng Yu, Haoran Luo, Yuqi Yi, and Fan Cheng.
\newblock A2r2: Robust unsupervised neural machine translation with adversarial
  attack and regularization on representations.
\newblock \emph{IEEE Access}, 9:\penalty0 19990--19998, 2021.

\bibitem[Zang et~al.(2020)Zang, Qi, Yang, Liu, Zhang, Liu, and
  Sun]{zang2020word}
Yuan Zang, Fanchao Qi, Chenghao Yang, Zhiyuan Liu, Meng Zhang, Qun Liu, and
  Maosong Sun.
\newblock Word-level textual adversarial attacking as combinatorial
  optimization.
\newblock In \emph{Proceedings of the 58th Annual Meeting of the Association
  for Computational Linguistics}, pp.\  6066--6080, 2020.

\bibitem[Zhang et~al.(2020)Zhang, Williams, Titov, and
  Sennrich]{zhang2020improving}
Biao Zhang, Philip Williams, Ivan Titov, and Rico Sennrich.
\newblock Improving massively multilingual neural machine translation and
  zero-shot translation.
\newblock In \emph{Proceedings of the 58th Annual Meeting of the Association
  for Computational Linguistics}, pp.\  1628--1639, 2020.

\bibitem[Zhang et~al.(2019)Zhang, Kishore, Wu, Weinberger, and
  Artzi]{zhang2019bertscore}
Tianyi Zhang, Varsha Kishore, Felix Wu, Kilian~Q Weinberger, and Yoav Artzi.
\newblock Bertscore: Evaluating text generation with bert.
\newblock In \emph{International Conference on Learning Representations}, 2019.

\bibitem[Zhang et~al.(2021)Zhang, Zhang, Chen, and He]{zhang2021crafting}
Xinze Zhang, Junzhe Zhang, Zhenhua Chen, and Kun He.
\newblock Crafting adversarial examples for neural machine translation.
\newblock In \emph{Proceedings of the 59th Annual Meeting of the Association
  for Computational Linguistics and the 11th International Joint Conference on
  Natural Language Processing (Volume 1: Long Papers)}, pp.\  1967--1977, 2021.

\end{thebibliography}
\bibliographystyle{tmlr}

\newpage
\clearpage

\appendix


\begin{figure*}
\noindent\rule[0.25in]{\textwidth}{4pt}
\begin{center}
\vskip -0.15in
\Large{\bf 
Supplementary Material\\
\large{TransFool: An Adversarial Attack against Neural Machine Translation Models} }  
\end{center}
\vskip 0.15in
\noindent\rule[0.09 in]{\textwidth}{1pt}
\end{figure*}

\begin{abstract}
In this supplementary material, we first provide some statistics of the evaluation datasets in Section \ref{dataset}. The ablation study of the  is presented in Section \ref{Analysis}: the effect of hyperparameters (Section \ref{abl:hyper}), different aspect of the language model (Section \ref{abl:lm}), and effect of NMT model beam size (Section \ref{abl:beam}).  We conducted human evaluation whose results and details are presented in Section \ref{hu}. 
More results of the white-box attack are reported in \ref{white}: the results of other similarity metrics (Section \ref{sim_white}), performance of TransFool against other translation tasks (Section \ref{other-lang}), performance over successful attacks (Section \ref{success_white}), trade-off between success rate and similarity/fluency (Section \ref{trade}), and some generated adversarial examples (Section \ref{samples_white}). Section \ref{black} provides more experiments on the black-box attack:  the performance of attacking \textit{Google Translate} (Section \ref{google}), results of other similarity metrics (Section \ref{sim_black}), and some generated adversarial examples (Section \ref{samples_black}). We discuss the effect of the back-translation model choice on WSLS in Section \ref{back_trans}. Finally, the license information and more details of the assets (datasets, codes, and models) are provided in Section \ref{license}.

\end{abstract}

\vspace{-5pt}

\section{Some statistics of the Datasets}\label{dataset}

\begin{wrapfigure}{R}{0.52\textwidth}
\vspace{-30pt}
\begin{small}
\begin{minipage}{0.52\textwidth}
\begin{table}[H]
	\centering
		\renewcommand{\arraystretch}{0.7}
	\setlength{\tabcolsep}{2.7pt}
	
	\caption{Some statistics of the evaluation datasets.}
	\vspace{-7pt}
	\label{tab:dataset}
	\scalebox{0.79}{
		\begin{tabular}[t]{@{} lccccccc @{}}
			\toprule[1pt]
		    \multirow{2}{*}{\textbf{Dataset}}  &    \multirow{1}{*}{{\textbf{Average}}} &
		    \multirow{1}{*}{{\textbf{\#Test}}} & 
		    \multicolumn{2}{c}{\textbf{Marian NMT}} &&
		    \multicolumn{2}{c}{\textbf{mBART50}} \\
		    
		    \cline{4-5}
			\cline{7-8}
			\rule{0pt}{2.5ex}    
			&  \textbf{Length} & \textbf{Samples}& BLEU & chrF && BLEU & chrF  \\
			
			\midrule[1pt]
			En-Fr  &  \multirow{2}{*}{27} & \multirow{2}{*}{3003} & \multirow{2}{*}{39.88} & \multirow{2}{*}{64.94} & & \multirow{2}{*}{36.17} & \multirow{2}{*}{62.66}\\
			WMT14 \\
			\midrule
			En-De  &  \multirow{2}{*}{26} & \multirow{2}{*}{3003} & \multirow{2}{*}{27.72} & \multirow{2}{*}{58.50} && \multirow{2}{*}{25.66} & \multirow{2}{*}{57.02} \\ 
			WMT14 \\
			\midrule
			En-Zh  &  \multirow{2}{*}{18} & \multirow{2}{*}{2000} & \multirow{2}{*}{33.11} & \multirow{2}{*}{50.98} && \multirow{2}{*}{29.27} & \multirow{2}{*}{41.92}\\
			OPUS-100 &&\\
			
			\bottomrule[1pt]
		\end{tabular}
	}
	\vspace{2pt}
\end{table}
\end{minipage}
\end{small}
\vspace{-30pt}
\end{wrapfigure}
Some statistics, including the number of samples, the Average length of the sentences, and the translation quality of Marian NMT and mBART50, of the evaluation datasets, i.e., OPUS100 (En-Zh) WMT14 (En-FR) and (En-De), are reported in table \ref{tab:dataset}.

\section{Ablation Study} \label{Analysis}

\subsection{Effect of Hyperparameters} \label{abl:hyper}
In this Section, we 
analyze the effect of different hyperparameters (including the coefficients $\alpha$ and $\beta$ in our optimization problem (\ref{optimization}), the step size of the gradient descent $\gamma$, and the relative BLEU score ratio $\lambda$ in the stopping criteria Eq. (\ref{sar})) on the white-box attack performance in terms of success  rate, semantic similarity, and perplexity score.  

In all the experiments, we consider English to French Marian NMT model and evaluate over the first 1000 sentences of the test set of WMT14. The default values for the hyperparameters are as follows, except for the hyperparameter that varies in the different experiments, respectively: $\alpha = 20$, $\beta = 1.8$, $\gamma = 0.016$, and $\lambda =  0.4$.

\begin{figure}[tb]
     \centering%
     \begin{subfigure}{0.26\textwidth}
         \centering
         \includegraphics[width=\textwidth]{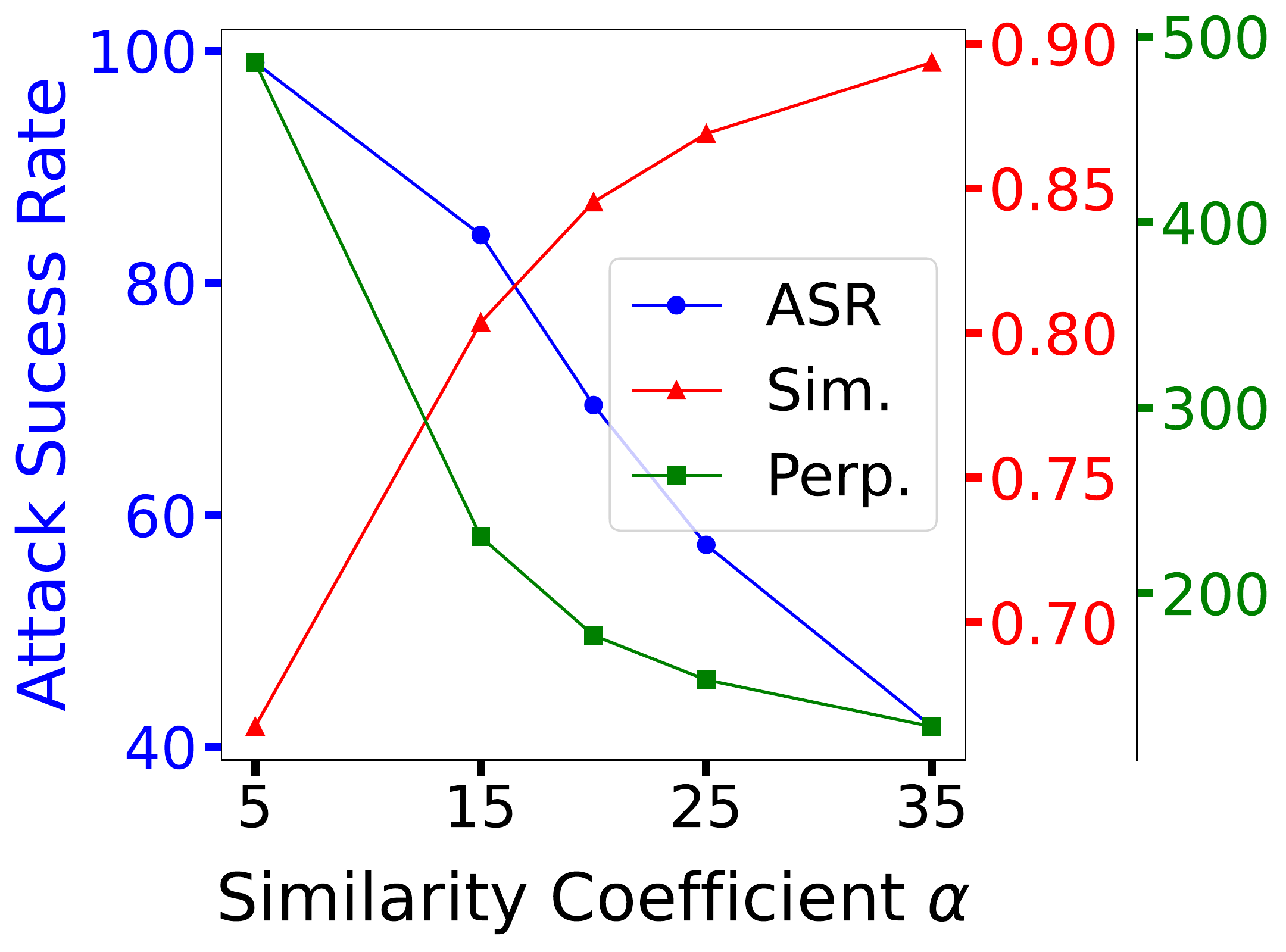}
         \vspace{-15pt}
         \caption{}
         \label{fig:sim}
     \end{subfigure}\,%
     \begin{subfigure}{0.24\textwidth}
         \centering
         \includegraphics[width=\textwidth]{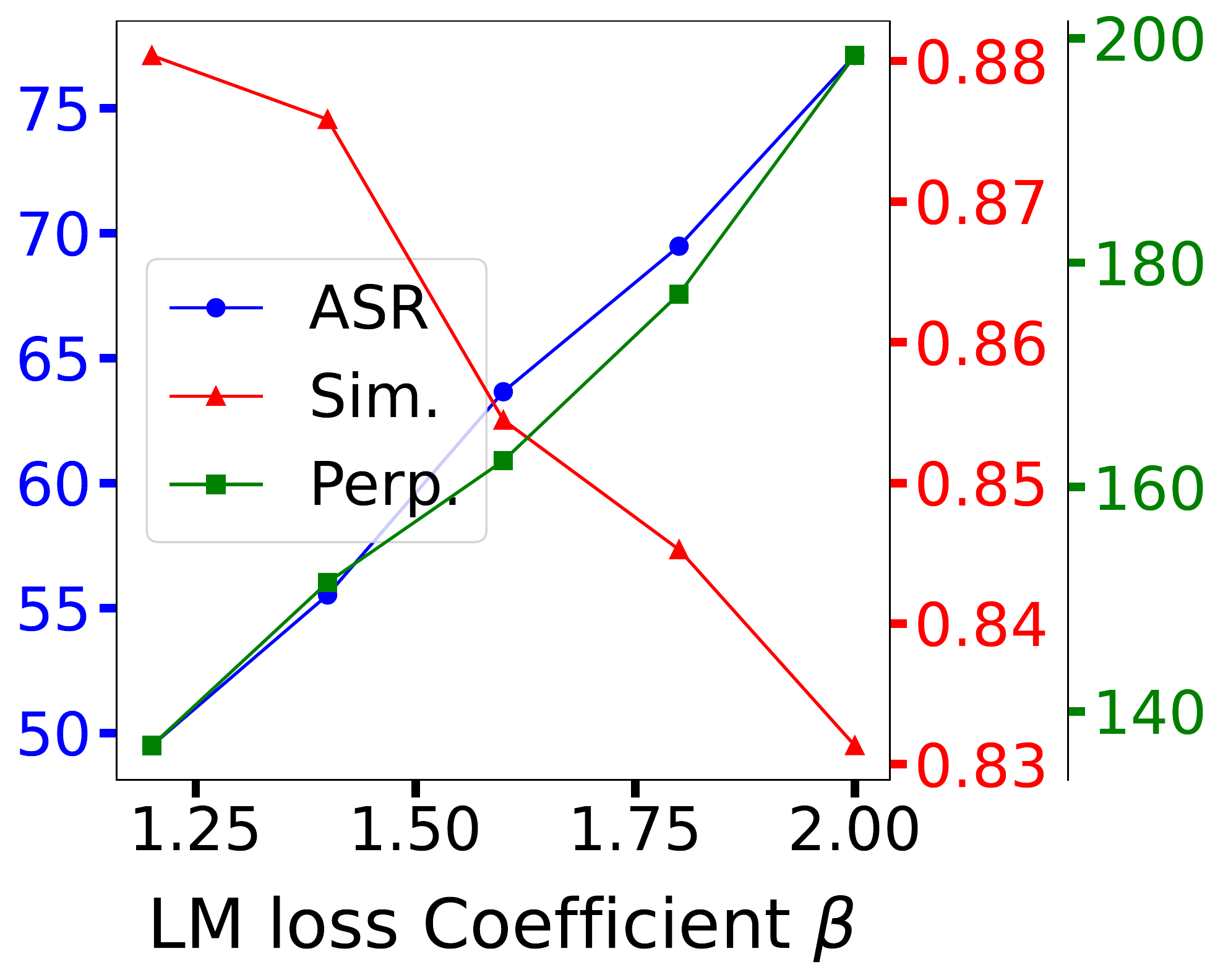}
         \vspace{-15pt}
         \caption{}
         \label{fig:perp}
     \end{subfigure}\,%
     \begin{subfigure}{0.24\textwidth}
         \centering
         \includegraphics[width=\textwidth]{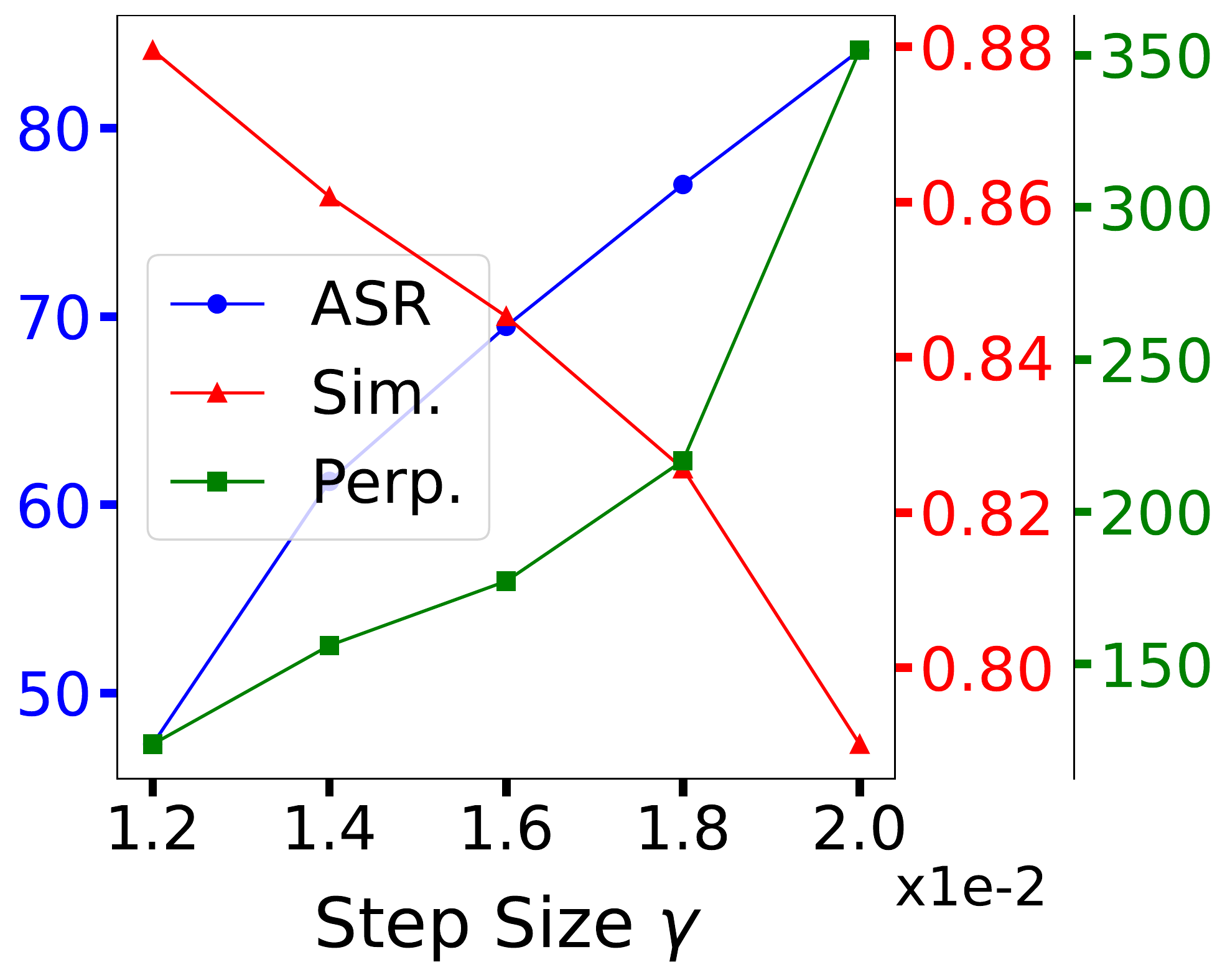}
         \vspace{-15pt}
         \caption{}
         \label{fig:lr}
     \end{subfigure}\,%
    \begin{subfigure}{0.26\textwidth}
         \centering
         \includegraphics[width=\textwidth]{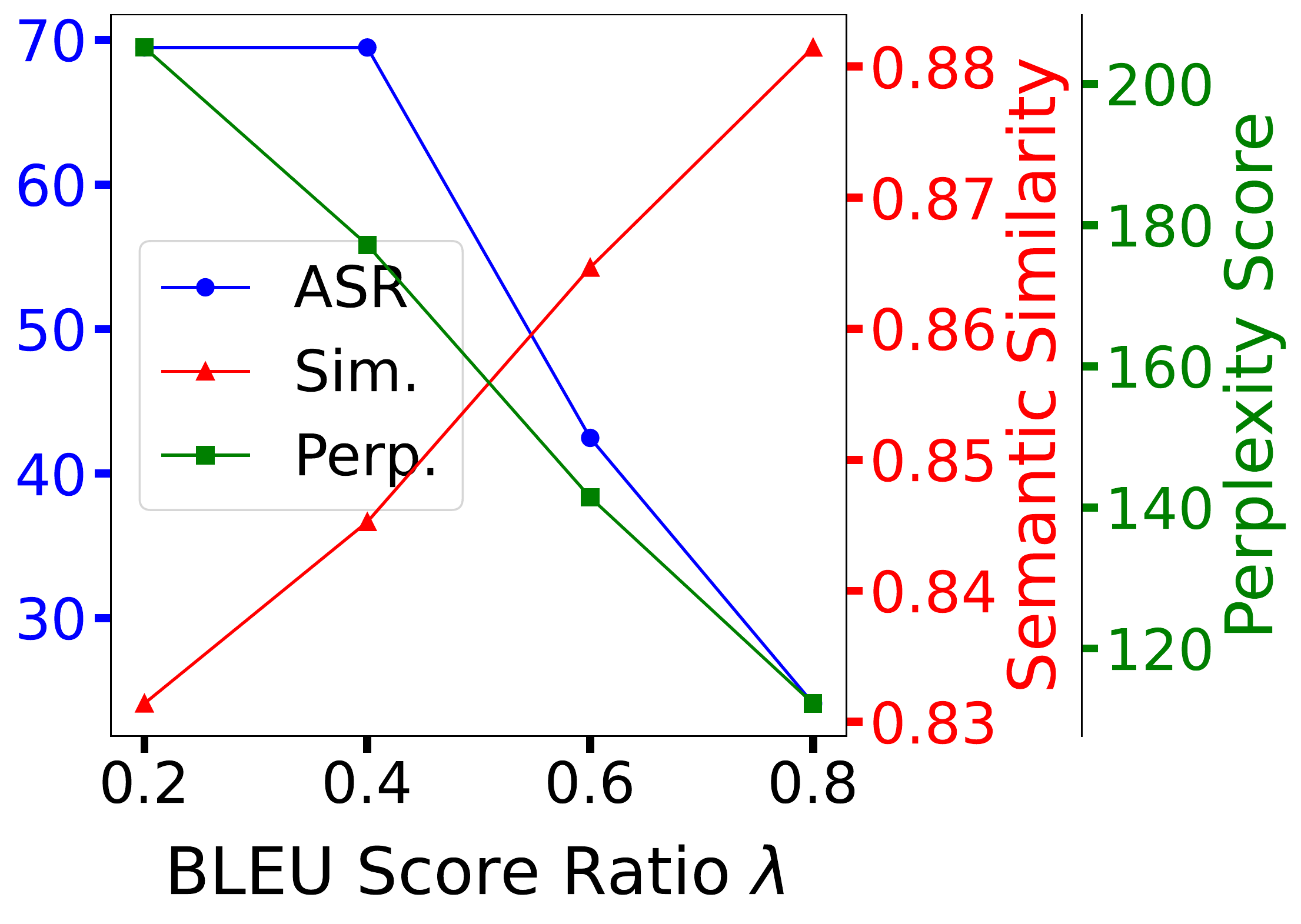}
         \vspace{-15pt}
         \caption{}
         \label{fig:bleu}
     \end{subfigure}\,%
     
     \smallskip
     
     \hspace{-12pt}
     \begin{subfigure}{0.24\textwidth}
         \centering
         
         \includegraphics[width=\textwidth]{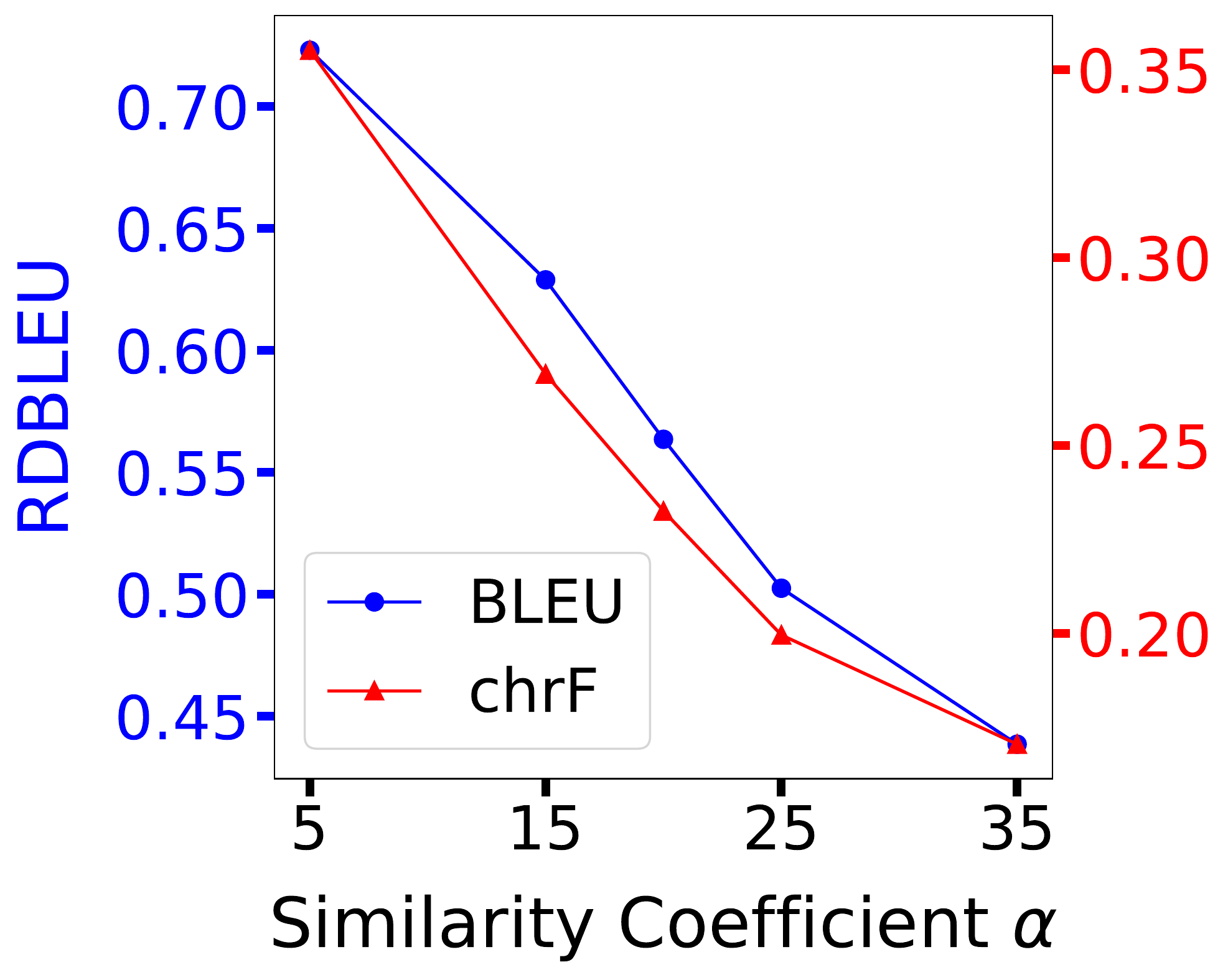}
         \vspace{-15pt}
         \caption{}
         \label{fig:sim_tr}
     \end{subfigure}\hspace{9pt}%
     \begin{subfigure}{0.22\textwidth}
         \centering
         \includegraphics[width=\textwidth]{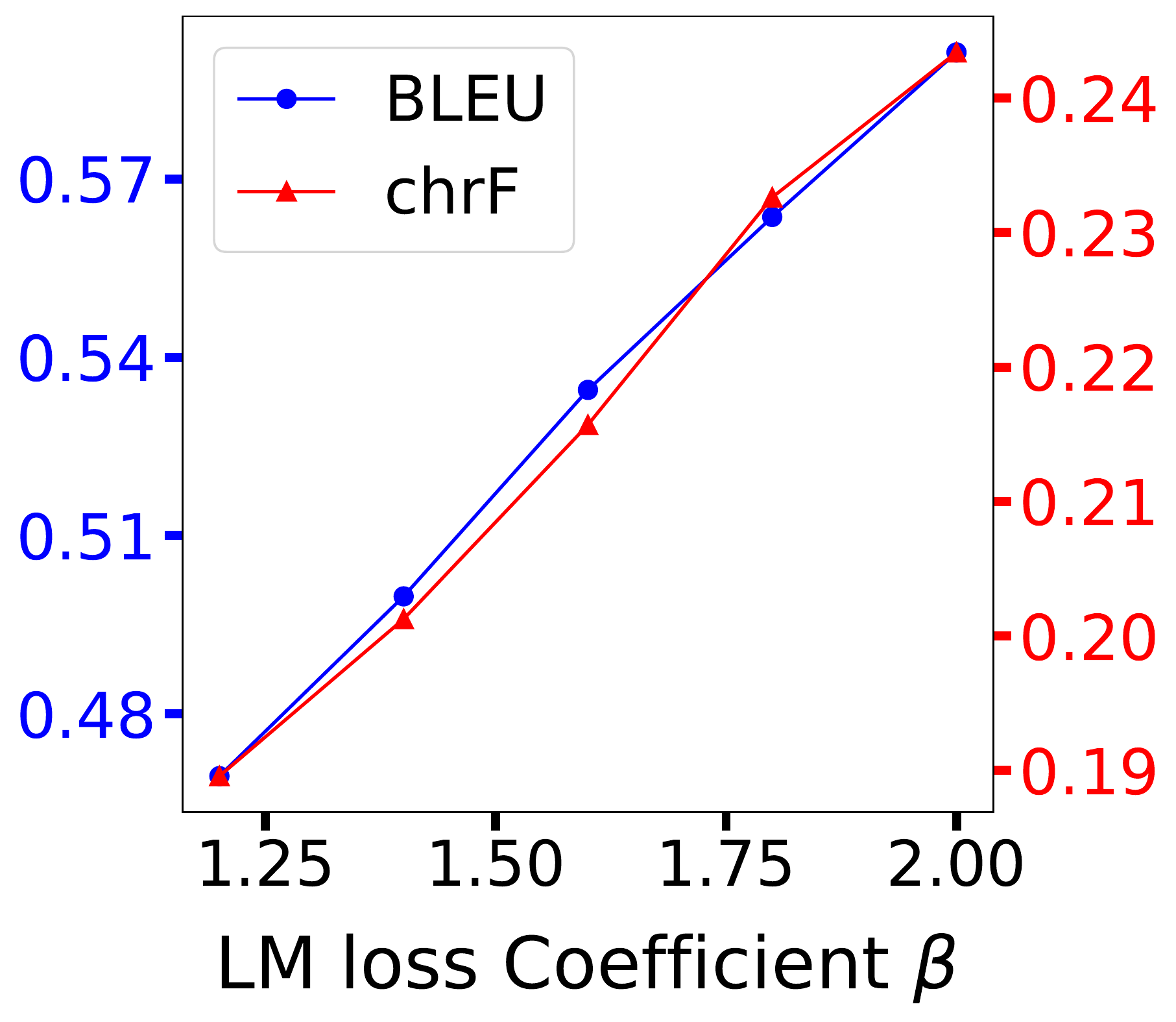}
         \vspace{-15pt}
         \caption{}
         \label{fig:perp_tr}
     \end{subfigure}\hspace{9pt}%
     \begin{subfigure}{0.22\textwidth}
         \centering
         \includegraphics[width=\textwidth]{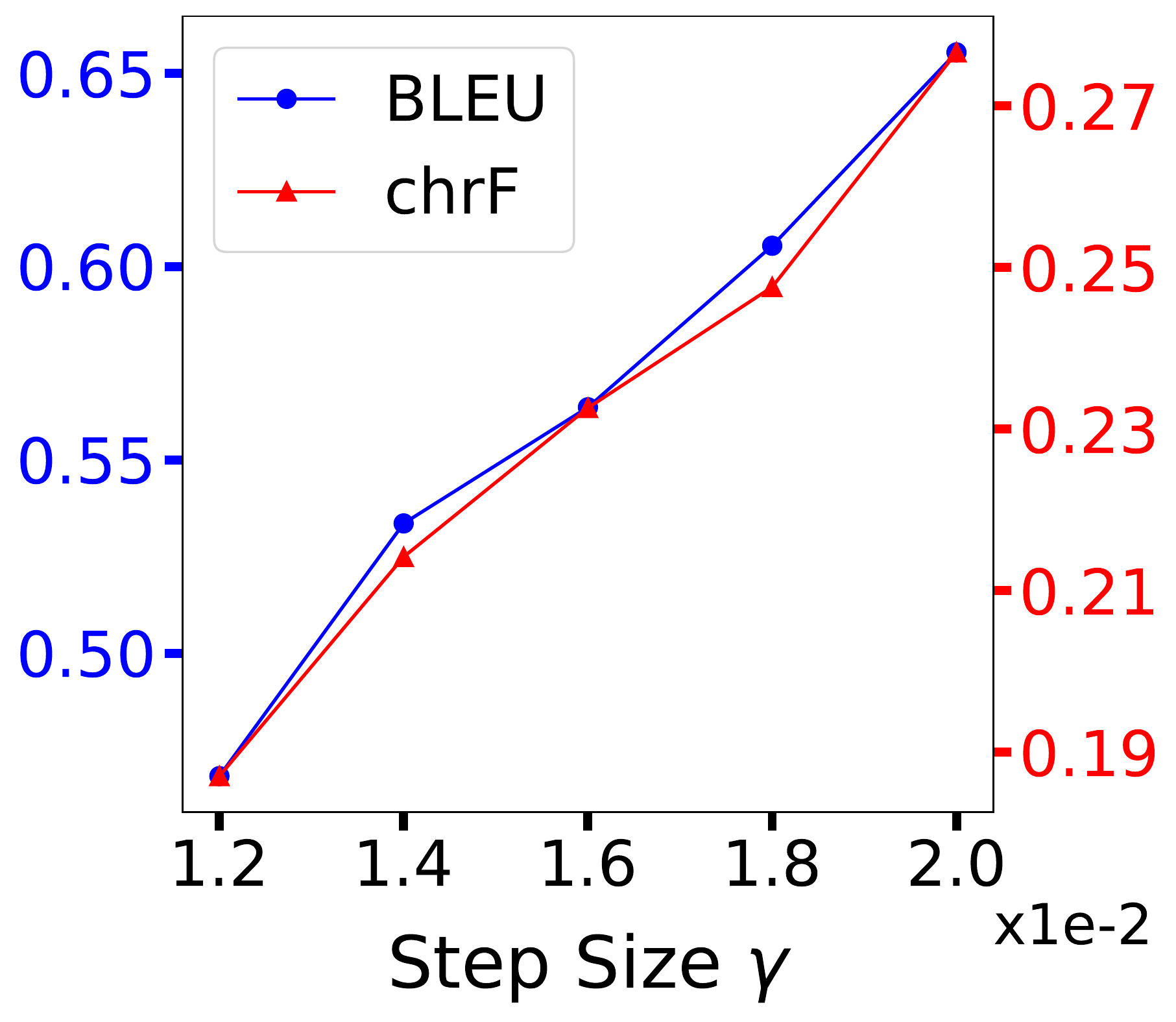}
         \vspace{-15pt}
         \caption{}
         \label{fig:lr_tr}
     \end{subfigure}\hspace{9pt}%
    \begin{subfigure}{0.24\textwidth}
         \centering
         \includegraphics[width=\textwidth]{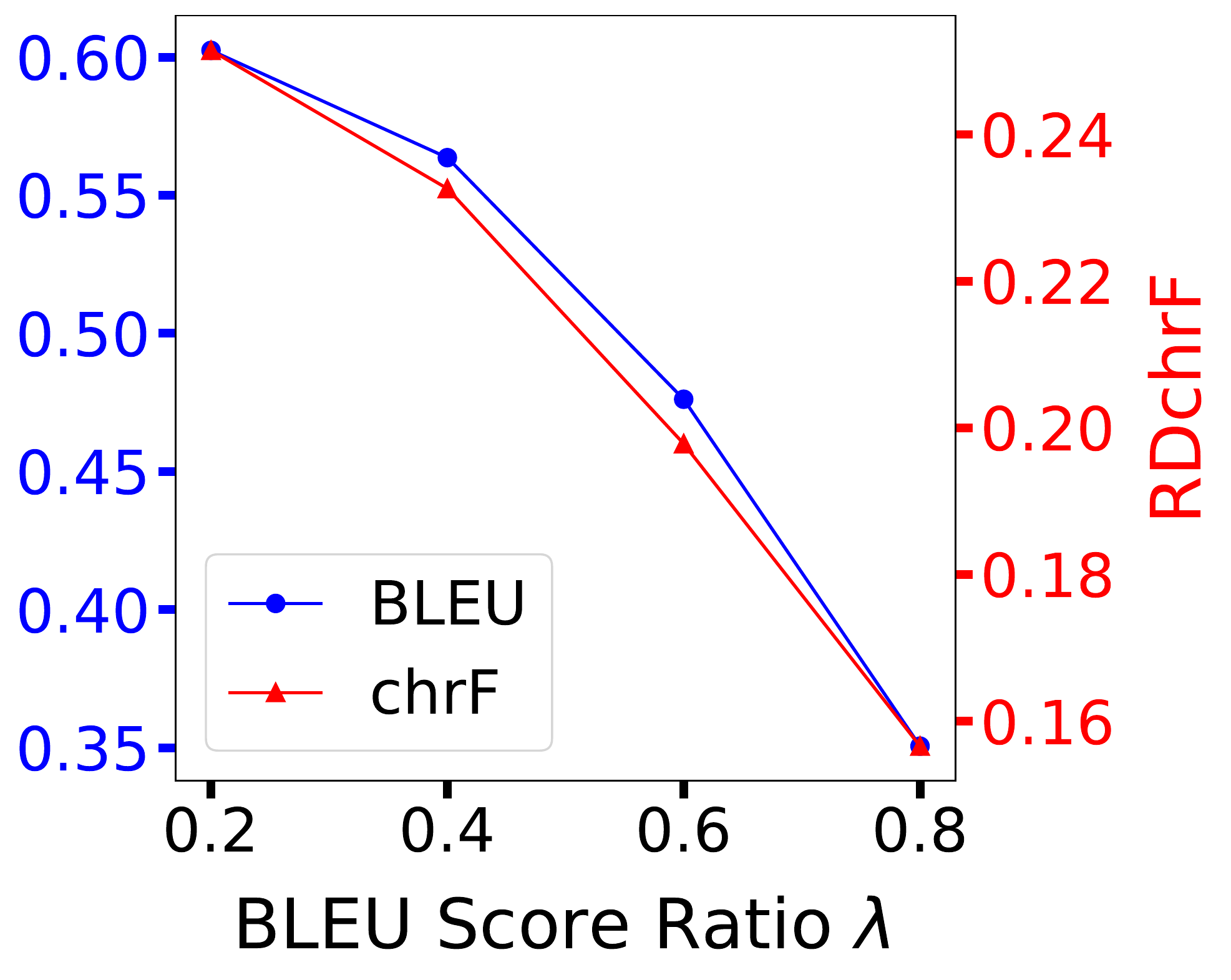}
         \vspace{-15pt}
         \caption{}
         \label{fig:bleu_tr}
     \end{subfigure}\,%
     
        \caption{Effect of different hyperparameters on the performance of TransFool.}
        \label{fig:hyperparams}
\end{figure}

\paragraph{Effect of the similarity coefficient $\alpha$.} This hyperparameter determines the strength of the similarity term in the optimization problem (\ref{optimization}). Figure \ref{fig:sim} shows the effect of $\alpha$ on the performance of our attack. By increasing the similarity coefficient of the proposed optimization problem, we are forcing our algorithm to find adversarial sentences that are more similar to the original sentence. Therefore, as shown in Figure \ref{fig:sim}, larger values of $\alpha$ result in higher semantic similarity. However, in this case, it is harder to fool the NMT model, i.e., lower attack success  rate, RDBLEU, and RDchrF. Moreover, it seems that, since the generated adversarial examples are more  similar to the original sentence, they are more natural, and their perplexity score is lower.

\paragraph{Effect of the language model loss coefficient $\beta$.} We analyze the impact of the hyperparameter $\beta$, which controls the importance of the language model loss term in the proposed optimization problem,  in Figure \ref{fig:perp}. By increasing this coefficient, 
we weaken the effect of the similarity term, i.e., the generated adversarial examples are less similar to the original sentence. As a result, the success rate and the effect on translation quality, i.e., RDBLEU and RDchrF, increase. 

\paragraph{Effect of the step size $\gamma$.} The step size of the gradient descent step of the algorithm can impact the performance of our attack, which is investigated in Figure \ref{fig:lr}.
Increasing the step size results in larger movement in the embedding space in each iteration of the algorithm. Hence, the generated  adversarial examples are  more aggressive, which results in lower semantic similarity and higher perplexity scores. However, we can find adversarial examples more easily and achieve a higher attack success rate, RDBLEU, and RDchRF.  

\paragraph{Effect of the BLEU score ratio $\lambda$.} This hyperparameter determines the stopping criteria of our iterative algorithm. Figure \ref{fig:bleu} studies the effects of this  hyperparameter on the performance of our attack. As this figure shows, a higher BLEU score ratio causes the algorithm to end in earlier iterations. Therefore, the changes applied to the sentence are less aggressive, and hence, we achieve higher semantic similarity and a lower perplexity score. However, the attack success rate, RDBLEU, and RDchrF decrease since we make fewer changes to the sentences.

\subsection{Ablation study on the language model} \label{abl:lm}
\begin{wrapfigure}{R}{0.42\textwidth}
\vspace{-30pt}
\begin{small}
\begin{minipage}{0.42\textwidth}
     \centering%

     \includegraphics[width=0.9\textwidth]{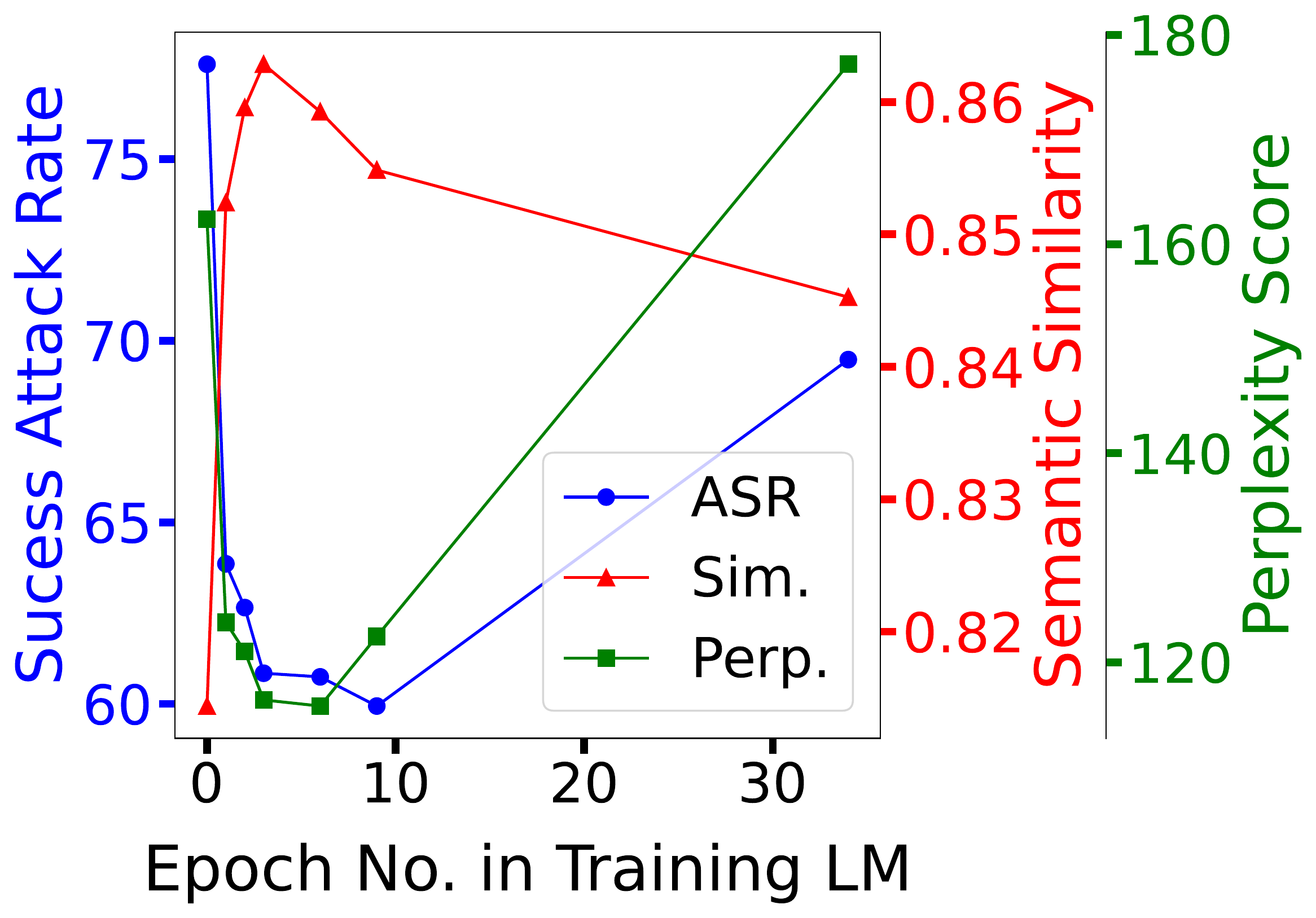}
     \vspace{-5pt}
     \caption{Effect of LM quality on performance.}
     \label{fig:lm-check}
         
\end{minipage}
\end{small}
\vspace{-20pt}
\end{wrapfigure}
One of the main component of our attack is the language model, which is used to measure the similarity between tokens.  We showed in Section \ref{LM} that  using the LM embeddings instead of the NMT ones highly affects our attack performance. In this Section, we study the impact of different properties of the LM used in TransFool on the attack performance.

\paragraph{Language Model Quality}

First, we investigate the effect of the language model quality on the performance. For Marian NMT (En-Fr), we save the checkpoints of the LM and FC layer after each epoch. Then, we use each of these models, and attack 1000 sentences of dataset. The results are presented in Figure \ref{fig:lm-check}. We can see that until a certain epoch, similarity and perplexity improves and then, they gradually deteriorate. This may be due to the fact that the LM becomes less generalizable to the dataset we attack after training for some epochs. We should note that we have considered the last LM from the last epoch in all our experiments.

\paragraph{Training Dataset of the Language Model}
\begin{wrapfigure}{R}{0.54\textwidth}
\vspace{-27pt}
\begin{small}
\begin{minipage}{0.54\textwidth}
\begin{table}[H]
	\centering
		\renewcommand{\arraystretch}{1}
	\setlength{\tabcolsep}{4pt}
	\caption{Performance of TransFool white-box attack against Marian NMT (En-Fr) with/without fine-tuned LM. }
	\label{tab:finetune}
	\scalebox{0.78}{
		\begin{tabular}[t]{@{} lcccccc @{}}
			\toprule[1pt]
		    \multirow{1}{*}{\textbf{Method}}  &
		    \scalebox{0.95}{ASR$\uparrow$} & \scalebox{0.95}{RDBLEU$\uparrow$} & \scalebox{0.95}{RDchrF$\uparrow$} & \scalebox{0.95}{Sim.$\uparrow$} & \scalebox{0.95}{Perp.$\downarrow$} & \scalebox{0.95}{TER$\downarrow$}\\
			\midrule[1pt]
			\multirow{1}{*}{with Original LM} &   69.48 & 0.56 & 0.23 & 0.85 & 177.20 & 13.91  \\ 
			\multirow{1}{*}{with fine-tuned LM} &   61.14 & 0.52 & 0.21 & 0.86 & 151.92 & 12.35 \\

			\bottomrule[1pt]
		\end{tabular}
	}
\end{table} 

\end{minipage}
\end{small}
\vspace{-10pt}
\end{wrapfigure}
To further study the effect of the language model, we can also fine-tune the language model and the fully-connected layer on the translation dataset (WMT14 En-Fr training set). Table \ref{tab:finetune} show the results when we use this language model in our attack. These results demonstrate that with a language model trained over sentences similar to those we are trying to attack, we can improve the perplexity score and similarity of the adversarial examples. 

\subsection{Effect of NMT model ‌Beam Size} \label{abl:beam}
\begin{wrapfigure}{R}{0.54\textwidth}
\vspace{-25pt}
\begin{small}
\begin{minipage}{0.54\textwidth}
\begin{table}[H]
	\centering
		\renewcommand{\arraystretch}{1}
	\setlength{\tabcolsep}{4pt}
	\caption{Performance of TransFool white-box attack against Marian NMT (En-Fr) with different beam sizes. }
	\label{tab:beam}
	\scalebox{0.8}{
		\begin{tabular}[t]{@{} lcccccc @{}}
			\toprule[1pt]
		    \multirow{1}{*}{\textbf{Beam Size}}  &
		    \scalebox{0.95}{ASR$\uparrow$} & \scalebox{0.95}{RDBLEU$\uparrow$} & \scalebox{0.95}{RDchrF$\uparrow$} & \scalebox{0.95}{Sim.$\uparrow$} & \scalebox{0.95}{Perp.$\downarrow$} & \scalebox{0.95}{TER$\downarrow$}\\
			\midrule[1pt]
			\multirow{1}{*}{1} &   69.28 & 0.58 & 0.24 & 0.85 & 172.18     & 13.70   \\ 
			\multirow{1}{*}{2} &   70.38 & 0.57 & 0.23 & 0.85 & 178.07     & 13.81 \\  
                \multirow{1}{*}{3} &   68.67 & 0.56 & 0.23 & 0.85 & 176.39 & 13.92 \\
                \multirow{1}{*}{4$^*$} &   69.48 & 0.56 & 0.23 & 0.85 & 177.20 & 13.91\\  
                \multirow{1}{*}{6} &   68.07 & 0.56 & 0.23 & 0.85 & 174.77 & 13.80 \\

			\bottomrule[1pt]
		\end{tabular}
	}
\end{table} 

\end{minipage}
\end{small}
\vspace{-10pt}
\end{wrapfigure}
Translation models use beam search to generate high-quality outputs. Marian NMT uses a beam size of 4, and mBART50 uses a beam size of 5. To see the effect of this parameter on TransFool performance, we attack Marian NMT  (En-Fr) with different beam sizes and the results are presented in Table \ref{tab:beam}. The attack performance is not significantly impacted by the beam size, possibly because  we employ the training loss for the adversarial loss, and beam size is only used for inference. This analysis shows the consistency in the performance of TransFool when the NMT model has different settings.

\section{Human Evaluation} \label{hu}

We conduct a human evaluation campaign of TransFool, kNN, and Seq2Sick attacks on Marian NMT (En-Fr) in the white-box setting. We randomly choose 90 sentences from the test set of the WMT14 (En-FR) dataset with the adversarial samples and their translations by the NMT model. We split 90 sentences into three different surveys to obtain a manageable size for each annotator. We recruited two volunteer annotators for each survey. For the English surveys, we ensure that the annotators are highly proficient   English speakers. Similarly, for the French survey, we ensure that the annotators are highly proficient in French. 

Before starting the rating task, we provided annotators with detailed guidelines similar to \citep{cer2017semeval, michel2019evaluation}. The task is to rate the sentences for each criterion on a continuous scale (0-100) inspired by WMT18 practice \citep{ma2018results} and Direct Assessment  \citep{graham2013continuous,graham2017can}. For each sentence, we evaluate three aspects in three different surveys:
\begin{itemize}
    \item \textit{Fluency}: We show the three adversarial sentences and the original sentence on the same page (in random order). We request the annotators to rate their level of agreement with the statement  \textit{"The sentence is fluent."}  for each sentence.
    \item \textit{Semantic preservation}: We show the original sentence on top, and below that, we display three adversarial sentences (in random order). We request the annotators to rate their level of agreement with the statement \textit{"The sentence is similar to the reference text."}  for each sentence.
    \item \textit{Translation quality}: Inspired by monolingual direct assessment  \citep{ma2018results,graham2013continuous,graham2017can}, we evaluate the translation quality by displaying the reference translation at the top and, below that, presenting translations of the three adversarial sentences (in a random order). We request the annotators to rate their level of agreement with the statement \textit{"The sentence is similar to the reference text."}  for each translation.
    
\end{itemize}

Only for the TransFool adversarial sentences we conduct a fourth survey. We show the English adversarial sentence on top, and below that, we display the reference translation or the translation generated by the NMT model. We request the annotators to rate their level of agreement with the statement \textit{"The French sentence is a good translation for the English sentence."}. We do this survey solely for TransFool sine the adversarial sentences by other attacks are not fluent and the task could be excessively challenging for the annotators.

\begin{figure}[tb]
     \centering%
     \begin{subfigure}{0.25\textwidth}
         \centering
         \includegraphics[width=\textwidth]{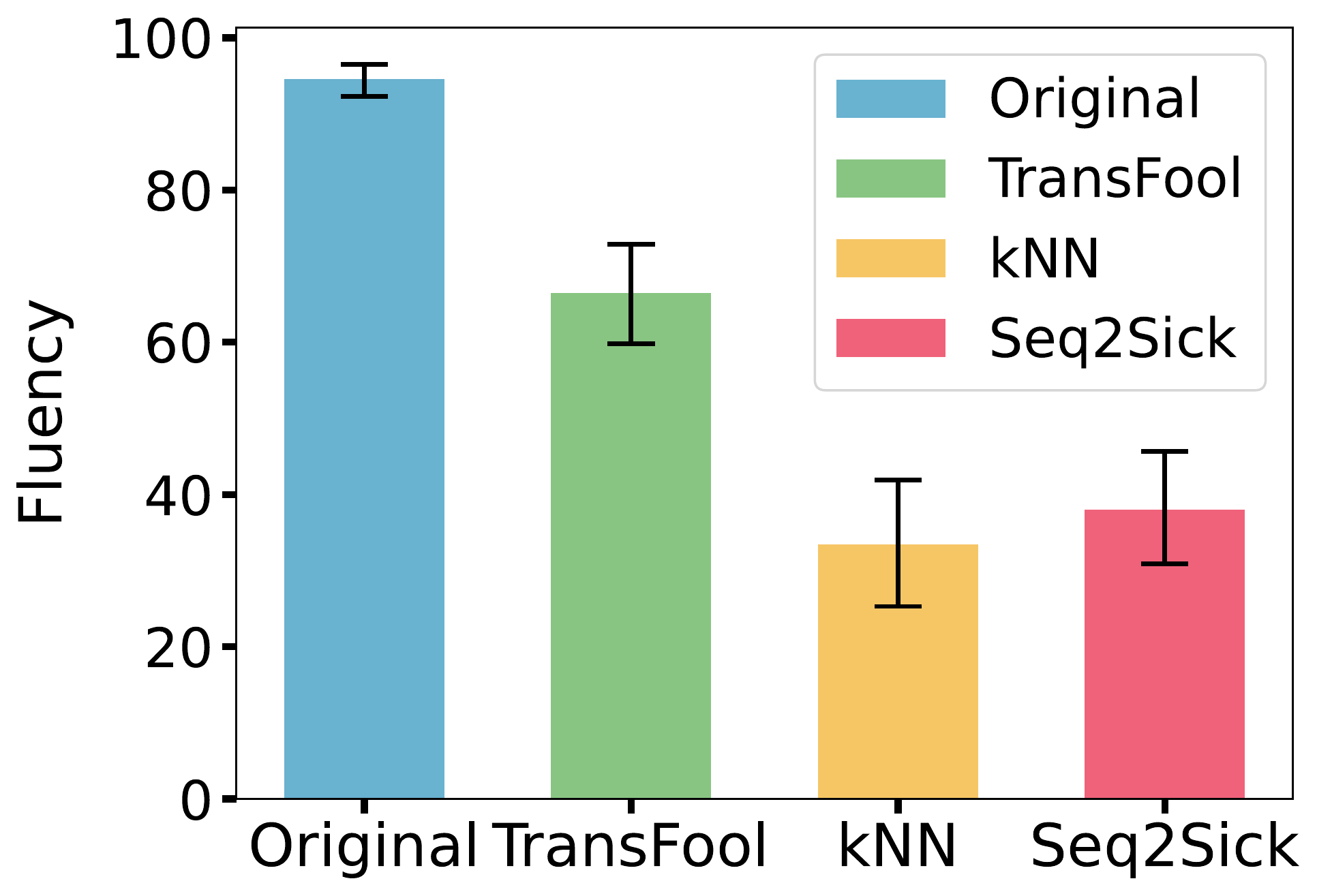}
         \vspace{-15pt}
         \caption{}
         \label{fig:flu}
     \end{subfigure}\,%
     \begin{subfigure}{0.25\textwidth}
         \centering
         \includegraphics[width=\textwidth]{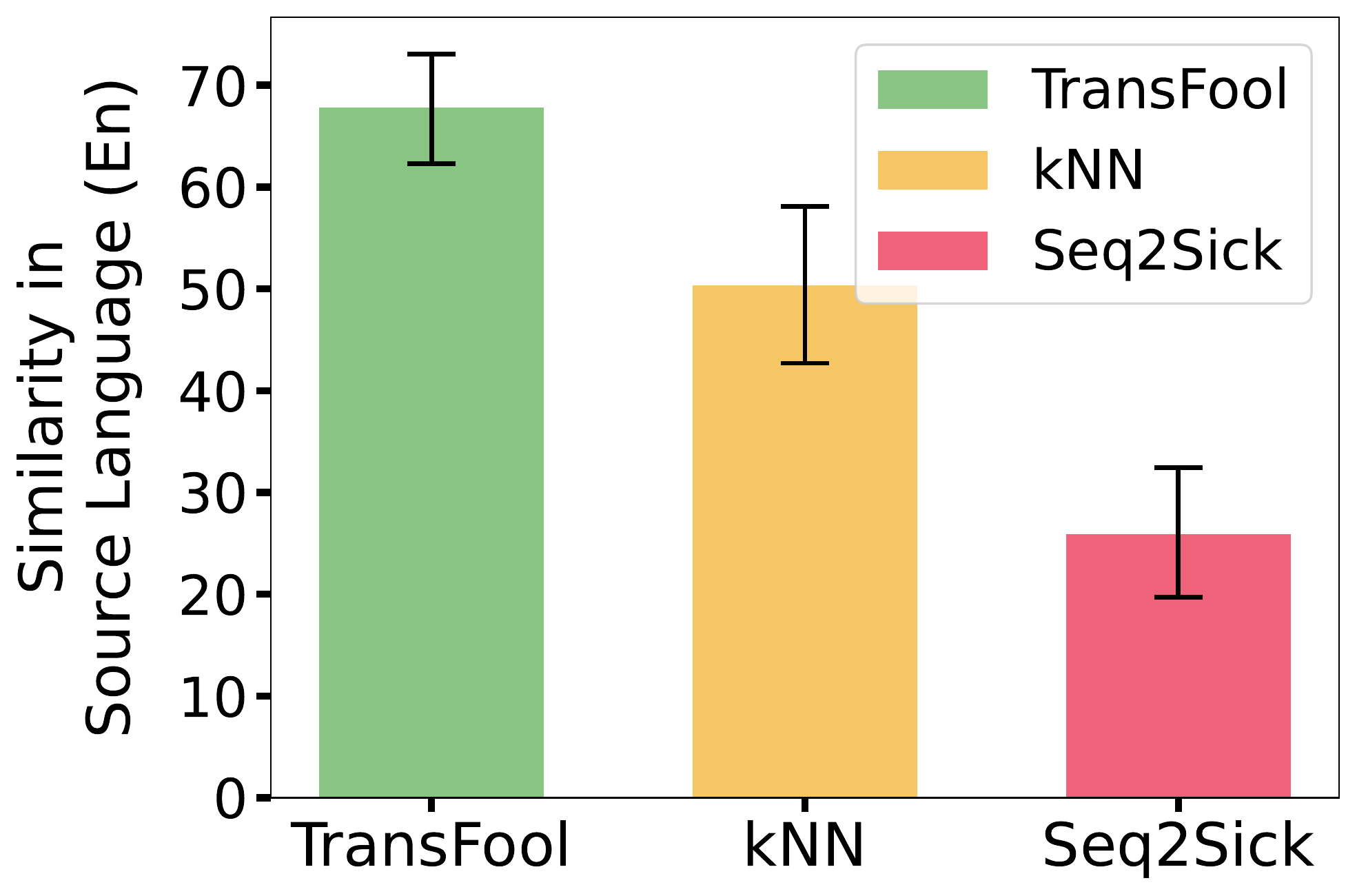}
         \vspace{-15pt}
         \caption{}
         \label{fig:sim_en}
     \end{subfigure}\,%
     \begin{subfigure}{0.25\textwidth}
         \centering
         \includegraphics[width=\textwidth]{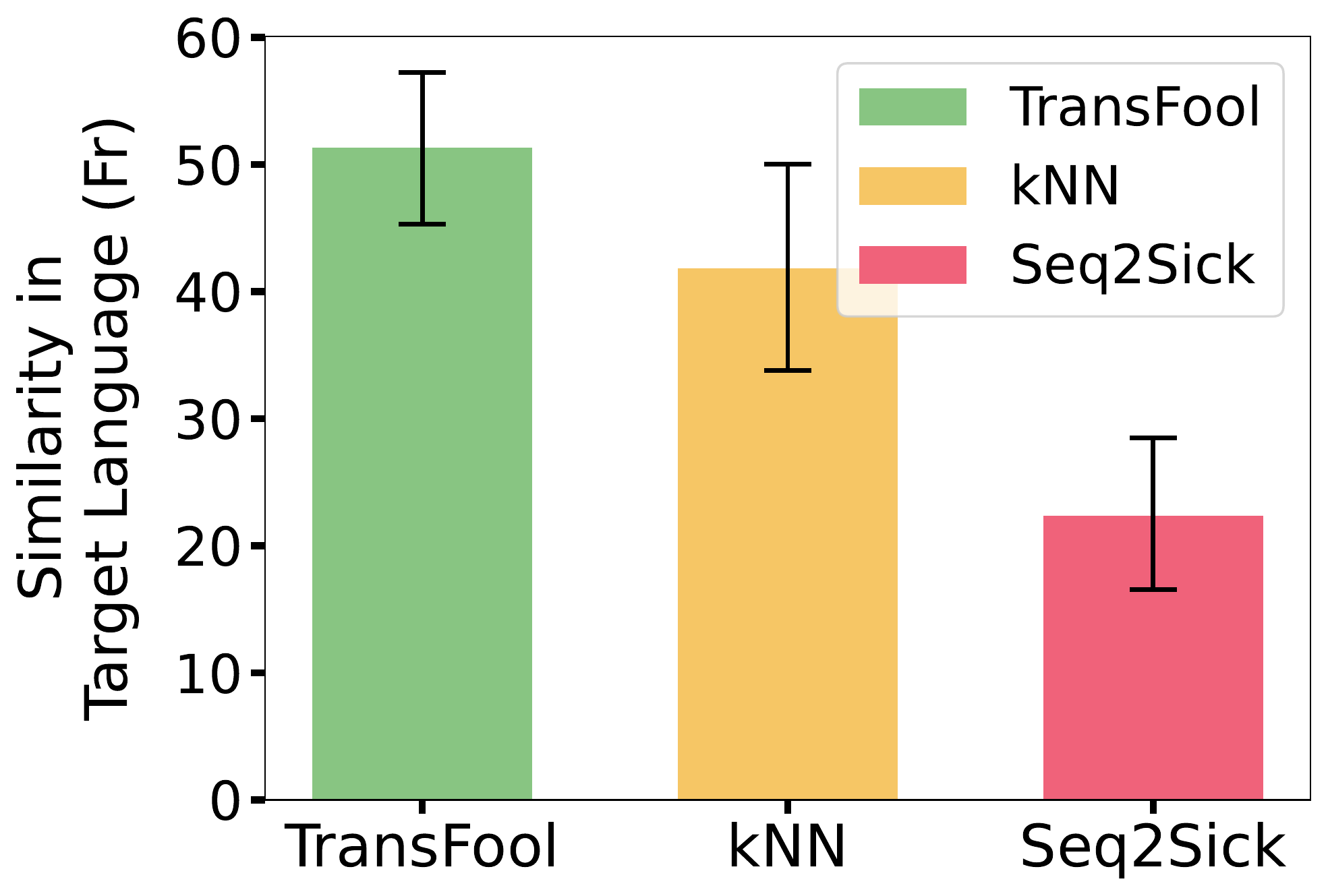}
         \vspace{-15pt}
         \caption{}
         \label{fig:sim_fr}
     \end{subfigure}\,%
    \begin{subfigure}{0.25\textwidth}
         \centering
         \includegraphics[width=\textwidth]{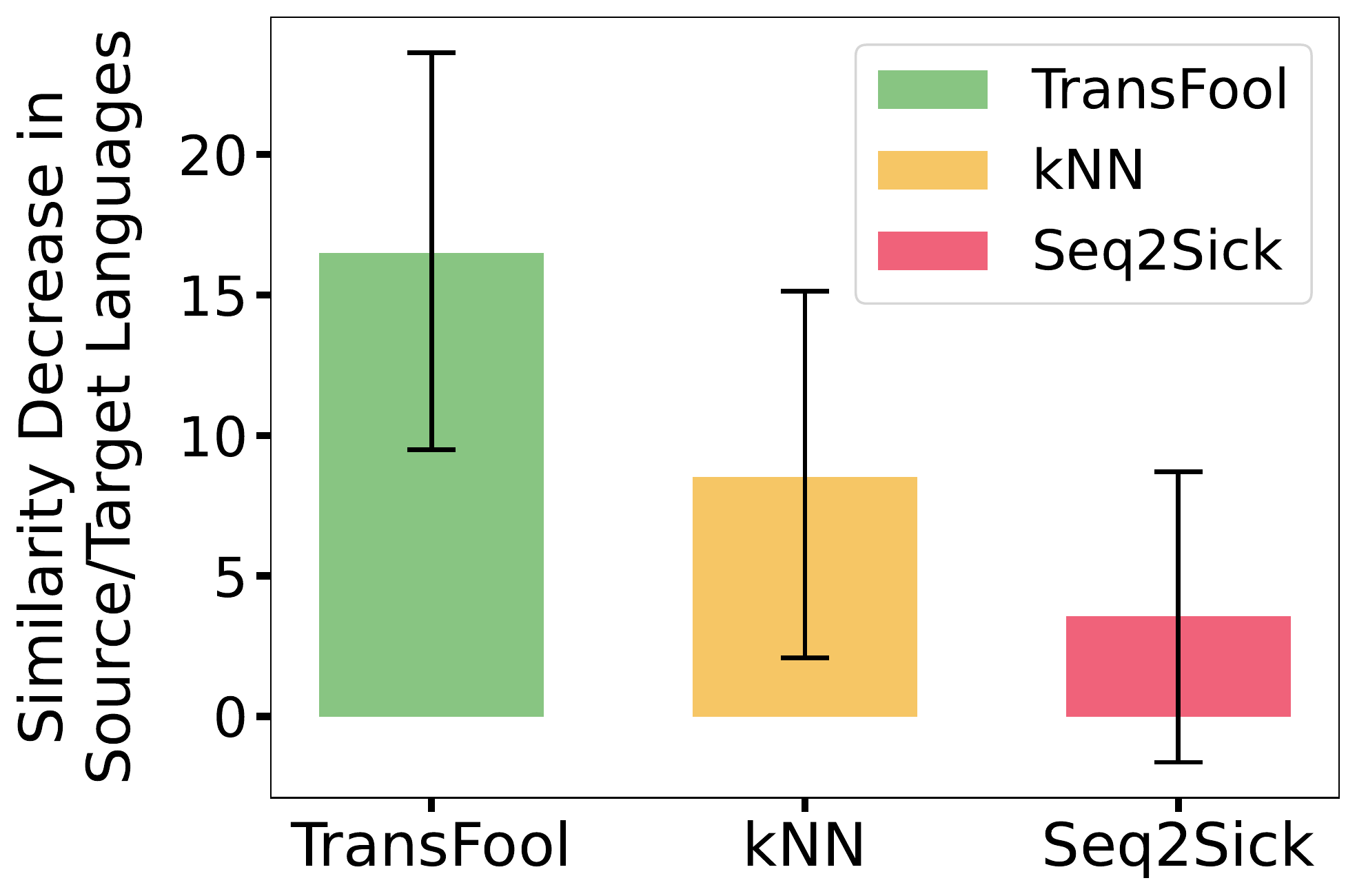}
         \vspace{-15pt}
         \caption{}
         \label{fig:sim_dec}
     \end{subfigure}\,%
    
     \vspace{-5pt}
        \caption{Human evaluation results for TransFool, kNN, and Seq2Sick attacks against Marian NMT (En-Fr).}
        \label{fig:HU}
        \vspace{-10pt}
\end{figure}

To ensure that the two annotators agree, we only consider sentences where their two corresponding scores are less than 30. We average both scores to compute the final score for each sentence. We calculate 95\% confidence intervals by using 15K bootstrap replications. The results are depicted in Figure \ref{fig:HU}. These results show that the adversarial examples generated by TransFool are more semantic-preserving and fluent than both baselines. According to the provided guide to the annotators for semantic similarity, the score of 67.8 shows that the two sentences are roughly equivalent, but some details may differ. Moreover, a fluency of 66.4 demonstrates that although the generated adversarial examples by TransFool are more fluent than the baselines, there is still room to improve the performance in this regard.

We follow the direct assessment strategy to measure the effectiveness of the adversarial attacks on translation quality. According to \citep{ma2018results}, since a sufficient level of agreement of translation quality is difficult to achieve with human evaluation, direct assessment simplifies the task to a simpler monolingual assessment instead of a bilingual task. The similarity of the translations of the adversarial sentences with the reference translation is shown in Figure \ref{fig:sim_fr}. The similarity of Seq2Sick is worse than other attacks. However, its similarity in the source language is also worse. Therefore, we compute the decrease of similarity (between the original and adversarial sentences) from the source language to the target language. The results in Figure \ref{fig:sim_dec} show that all attacks affect the translation quality and the effect of TransFool  is more pronounced than that of both baselines. 

\begin{wrapfigure}{R}{0.52\textwidth}
\vspace{-20pt}
\begin{small}
\begin{minipage}{0.52\textwidth}
     \centering%

     \includegraphics[width=0.6\textwidth]{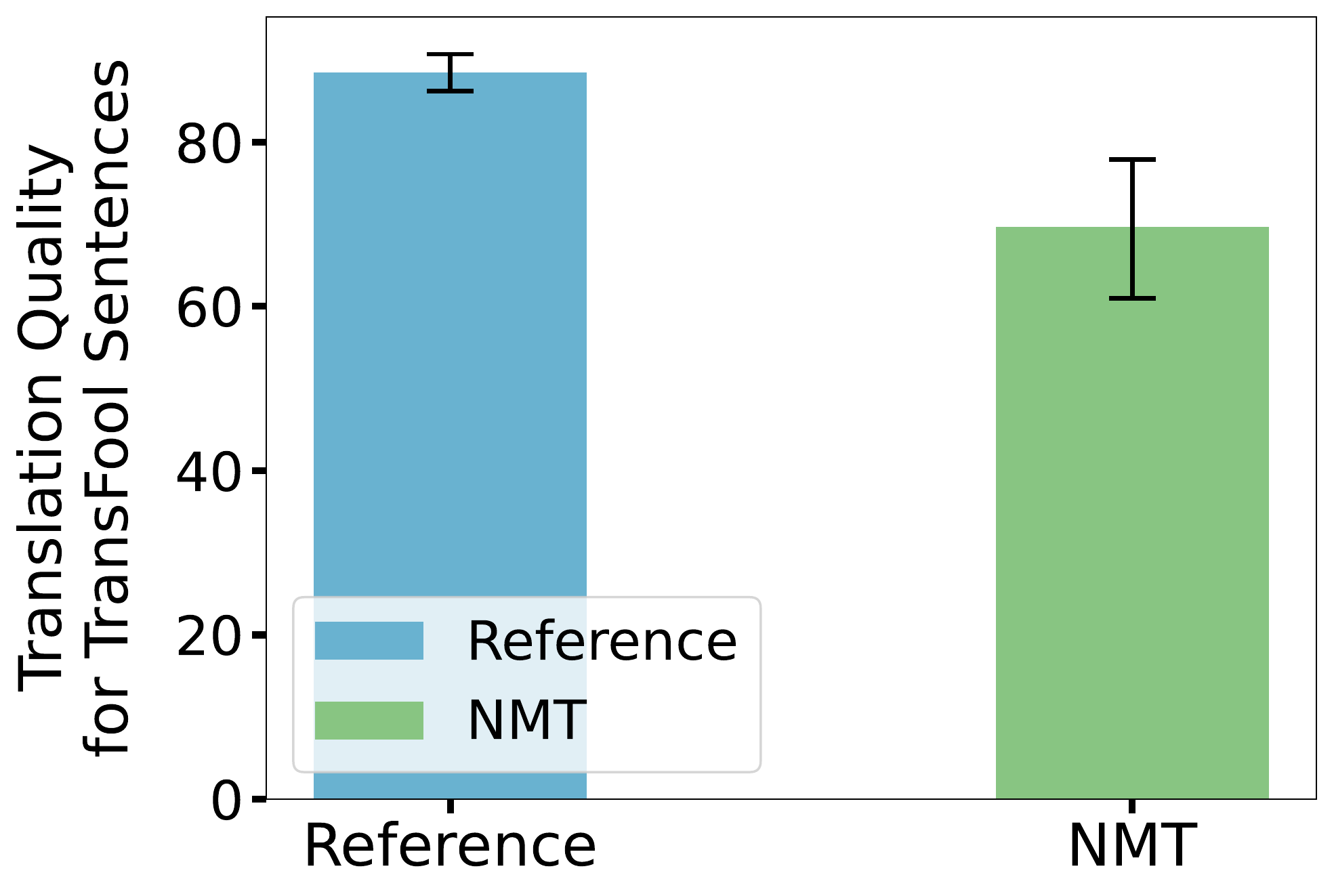}
     \vspace{-7pt}
     \caption{Human evaluation for  TransFool translations.}
     \label{fig:hu-tr}
         
\begin{table}[H]
	\centering
		\renewcommand{\arraystretch}{1}
	\setlength{\tabcolsep}{4pt}
	\caption{IAA for different surveys in human evaluation.}
	\label{tab:IAA}
	\scalebox{0.81}{
		\begin{tabular}[t]{@{} cccc @{}}
			\toprule[1pt]
		    \multirow{1}{*}{\textbf{Sentence Type}}  &
		    \textbf{Fluency}  & \textbf{Similarity in En}  & \textbf{Similarity in Fr}  \\
			\midrule[1pt]
			\multirow{1}{*}{Original}  &  0.68 & - & - \\
			\midrule[1pt]
			\multirow{1}{*}{TransFool}  &  0.85 & 0.82  & 0.79 \\ 
			\midrule[1pt]
			\multirow{1}{*}{kNN}  &  0.91 & 0.82  & 0.86 \\ 
			
			\midrule[1pt]
			\multirow{1}{*}{Seq2Sick}  &  0.89 & 0.88 & 0.83 \\ 
			
			\bottomrule[1pt]
		\end{tabular}
	}
\end{table}
\end{minipage}
\end{small}
\vspace{-20pt}
\end{wrapfigure}
Regarding the fourth survey on TransFool adversarial sentences, for the successful adversarial examples, the scores and their corresponding confidence interval for the translations generated by the  NMT model and the reference translations are presented in Figure \ref{fig:hu-tr}. We can see that the reference translation has got a  higher score than the NMT translations, which suggests that the translation quality of the target model is indeed degraded. 

Finally, for all surveys, we calculate Inter-Annotator Agreement (IAA) between the two human judgments for each sentence.  We compute IAA in terms of Pearson Correlation coefficient instead of the commonly used Cohen’s K since scores are in a continuous scale. The results are presented in Table \ref{tab:IAA}. Overall, we conclude that we achieve a good inter-annotator agreement for all sentence types and evaluation metrics. Moreover, for the fourth survey, the IAA is 0.40 and 0.65 for the reference translations and the NMT translations, respectively. The lower IAA for this survey compared to the other ones is due the the challenging task of translation quality assessment in two languages \cite{ma2018results}.

\section{More Results on the White-box Attack
} \label{white}

\subsection{Semantic Similarity Computed by Other Metrics}\label{sim_white}

To better assess the ability of adversarial attacks in maintaining semantic similarity, we can compute the similarity between the original and adversarial sentences using other metrics such as BERTScore \citep{zhang2019bertscore} and BLEURT-20 \citep{sellam2020learning}. 
It is shown in \citep{zhang2019bertscore} that BERTScore correlates well with human judgments. BLEURT-20 is also shown to correlates better with human judgment  than traditional measures \citep{freitag2021results}. 
The results are reported in Table \ref{tab:white_sim}. These results indicate that the TransFool is indeed more capable of preserving the semantics of the input sentence. In the two cases where kNN has better similarity by using the Universal Sentence Encoder (USE)  \citep{yang2020multilingual}, the performance of TransFool is better in terms of BERTScore and BLEURT-20. 

\begin{table}[H]
	\centering
		\renewcommand{\arraystretch}{1}
	\setlength{\tabcolsep}{4pt}
	\caption{Similarity performance  of white-box attacks
	.}
	\vspace{-5pt}
	\label{tab:white_sim}
	\scalebox{0.8}{
		\begin{tabular}[t]{@{} lcccccccc @{}}
			\toprule[1pt]
		    \multirow{2}{*}{\textbf{Task}}  &
		    \multirow{2}{*}{\textbf{Method}}  & \multicolumn{3}{c}{\textbf{Marian NMT}} &&   \multicolumn{3}{c}{\textbf{mBART50}} \\
			\cline{3-5}
			\cline{7-9}
			\rule{0pt}{2.5ex}    
			& & \scalebox{0.95}{USE$\uparrow$} & \scalebox{0.95}{BERTScore$\uparrow$} &
			\scalebox{0.95}{BLEURT-20 $\uparrow$} && 
			\scalebox{0.95}{USE$\uparrow$} & \scalebox{0.95}{BERTScore$\uparrow$} &
			\scalebox{0.95}{BLEURT-20 $\uparrow$} 
			\\
			\midrule[1pt]
			\multirow{3}{*}{En-Fr} & TransFool &  \textbf{0.85} & \textbf{0.95} & \textbf{0.65}
			&& \underline{0.84} & \textbf{0.96} & \textbf{0.70}
			\\ 
			& kNN &  \underline{0.82} & \underline{0.94} & \underline{0.61} 
			&& \textbf{0.85} & 0.93 & \underline{0.67}
			\\
			& Seq2Sick & 0.75 & \underline{0.94} & 0.60
			&& 0.75 & \underline{0.94} & 0.66
			\\
			\midrule[1pt]
			\multirow{3}{*}{En-De} & TransFool &  \textbf{0.84} & \textbf{0.96} & \textbf{0.67}
			&& \underline{0.83} & \textbf{0.95} & \textbf{0.69}
			\\ 
			& kNN &   \underline{0.82} & \underline{0.94} & \underline{0.61}  
			&& \textbf{0.86} & \underline{0.93} & \underline{0.67} 
			\\
			& Seq2Sick &  0.67 & 0.93 & 0.52 
			&&  0.66 & 0.92 & 0.58
			\\
			\midrule[1pt]
			\multirow{3}{*}{En-Zh} & TransFool &  \textbf{0.88} & \textbf{0.96} & \textbf{0.67} 
			&& \textbf{0.90} & \textbf{0.97} & \textbf{0.76} 
			\\ 
			& kNN &   \underline{0.86} & \underline{0.95} & \underline{0.66} 
			&& \textbf{0.90} & \underline{0.95} & \underline{0.72} 
			\\
			& Seq2Sick & 0.73 & 0.94 & 0.54
			&& 0.78 & \underline{0.95} & 0.67
			\\
			
			\bottomrule[1pt]
		\end{tabular}
	}
	\vspace{-5pt}
\end{table}

\subsection{Performance of TransFool against other Translation Tasks}\label{other-lang}
We also tested the generalizability of TransFool by attacking English-to-Russian (En-Ru) and English-to-Czech (En-Cs) translation tasks. The results are reported in Table \ref{tab:lang}. We can see that the performance of TransFool in different translation tasks is consistent with the previous ones we studied. 

\begin{table*}[h]
	\centering
		\renewcommand{\arraystretch}{0.95}
	\setlength{\tabcolsep}{4pt}
	\caption{Performance of TransFool white-box attack against other translation tasks 
	.}
	\vspace{-5pt}
	\label{tab:lang}
	\scalebox{0.75}{
		\begin{tabular}[t]{@{} lccccccccccccc @{}}
			\toprule[1pt]
		    \multirow{2}{*}{\textbf{Task}}  &
		     \multicolumn{6}{c}{\textbf{Marian NMT}} &&   \multicolumn{6}{c}{\textbf{mBART50}} \\
			\cline{2-7}
			\cline{9-14}
			\rule{0pt}{2.5ex}    
			&  \scalebox{0.95}{ASR$\uparrow$} & \scalebox{0.95}{RDBLEU$\uparrow$} & \scalebox{0.95}{RDchrF$\uparrow$} & \scalebox{0.95}{Sim.$\uparrow$} & \scalebox{0.95}{Perp.$\downarrow$} & \scalebox{0.95}{TER$\downarrow$} && \scalebox{0.95}{ASR$\uparrow$} & \scalebox{0.95}{RDBLEU$\uparrow$} & \scalebox{0.95}{RDchrF$\uparrow$} & \scalebox{0.95}{Sim.$\uparrow$} & \scalebox{0.95}{Perp.$\downarrow$} & \scalebox{0.95}{TER$\downarrow$}\\
			\midrule[1pt]
			\multirow{1}{*}{En-Ru} &   75.78 & 0.65 & 0.27 & 0.84 &  209.82 & 14.60 && 61.99 & 0.57 & 0.24 & 0.86 & 111.01 & 9.66 \\ 
			\midrule[1pt]
			\multirow{1}{*}{En-Cs} & 68.81 & 0.65 & 0.24 & 0.86 & 191.36 & 12.68 && 59.84 & 0.61 & 0.23 & 0.84 & 116.89 & 10.26\\

			\bottomrule[1pt]
		\end{tabular}
	}
	\vspace{-7pt}
\end{table*}

\subsection{Performance over Successful Attacks}\label{success_white}
The evaluation metrics of the successful adversarial examples that strongly affect the translation quality are also important, and they show the capability of the adversarial  attack. Hence, we evaluate TransFool, kNN, and Seq2Sick only over the successful adversarial examples.\footnote{As defined in Section \ref{Experiments}, the adversarial example is successful if  the BLEU score of its translation is less than half of the BLEU score of the original translation.} The results for the white-box setting are presented in Table \ref{tab:white_success}. By comparing this Table and Table \ref{tab:white}, which shows the results on the whole dataset, we can see that TransFool performance is \textit{consistent} among successful and unsuccessful attacks. Moreover, successful adversarial examples generated by TransFool 
are still semantically similar to the original sentences, and their perplexity score is low. However, the successful adversarial examples generated by Seq2Sick and kNN do not preserve the semantic similarity and are not fluent sentences; hence, they are \textit{not valid} adversarial sentences. 

\begin{table*}[h]
	\centering
		\renewcommand{\arraystretch}{1}
	\setlength{\tabcolsep}{4pt}
	\caption{Performance of white-box attack over successful adversarial examples
	.}
	\vspace{-5pt}
	\label{tab:white_success}
	\scalebox{0.75}{
		\begin{tabular}[t]{@{} lcccccccccccccc @{}}
			\toprule[1pt]
		    \multirow{2}{*}{\textbf{Task}}  &
		    \multirow{2}{*}{\textbf{Method}}  & \multicolumn{6}{c}{\textbf{Marian NMT}} &&   \multicolumn{6}{c}{\textbf{mBART50}} \\
			\cline{3-8}
			\cline{10-15}
			\rule{0pt}{2.5ex}    
			& & \scalebox{0.95}{ASR$\uparrow$} & \scalebox{0.95}{RDBLEU$\uparrow$} & \scalebox{0.95}{RDchrF$\uparrow$} & \scalebox{0.95}{Sim.$\uparrow$} & \scalebox{0.95}{Perp.$\downarrow$} & \scalebox{0.95}{TER$\downarrow$} && \scalebox{0.95}{ASR$\uparrow$} & \scalebox{0.95}{RDBLEU$\uparrow$} & \scalebox{0.95}{RDchrF$\uparrow$} & \scalebox{0.95}{Sim.$\uparrow$} & \scalebox{0.95}{Perp.$\downarrow$} & \scalebox{0.95}{TER$\downarrow$}\\
			\midrule[1pt]
			\multirow{3}{*}{En-Fr} & TransFool &  69.38 & 0.66 & 0.26 & 0.83 &  229.75 & 15.33 && 60.68 & 0.66 & 0.27 & 0.82 & 164.52 & 12.56 \\ 
			& kNN &   36.53 & 0.70 & 0.30 & 0.76 & 746.89 & 24.52 && 30.84 & 0.72 & 0.28 & 0.77 &  691.64 & 28.05 \\
			& Seq2Sick & 27.01 & 0.72 & 0.40 & 0.56 & 648.92 & 25.28 && 25.53 & 0.74 & 0.41 & 0.53 &  556.61 & 25.16 \\
			\midrule[1pt]
			\multirow{3}{*}{En-De} & TransFool &  69.49 & 0.72 & 0.25 & 0.83 & 191.51 & 14.54 && 62.87 & 0.73 & 0.26 & 0.81 & 169.76 & 12.66\\ 
			& kNN &   39.22 & 0.75 & 0.29 & 0.77 & 675.01 & 23.07 && 35.99 & 0.75 & 0.23 & 0.81 &  574.68 & 25.75 \\
			& Seq2Sick &  35.60 & 0.78 & 0.40 & 0.53 & 659.90 & 25.67 && 35.59 & 0.78 & 0.40 & 0.52 & 612.22 & 26.67\\
			\midrule[1pt]
			\multirow{3}{*}{En-Zh} & TransFool &  73.82 & 0.76 & 0.34 & 0.87 & 112.28 & 12.83 && 57.50 & 0.73 & 0.31 & 0.88 & 99.08 & 9.86\\ 
			& kNN &   31.12 & 0.72 & 0.29 & 0.80 & 355.25 & 22.55 && 27.25 & 0.76 & 0.27 & 0.85 & 295.53 &  23.58 \\
			& Seq2Sick & 28.76 & 0.72 & 0.46 & 0.58 & 437.49 & 26.84 && 24.25 & 0.79 & 0.44 & 0.60 & 292.55 &  25.59 \\
			
			\bottomrule[1pt]
		\end{tabular}
	}
	\vspace{-5pt}
\end{table*}

\subsection{Trade-off between Success Rate and Similarity/Fluency}\label{trade}

The results in our ablation study \ref{Analysis} show that there is a trade-off between the quality of adversarial example, in terms of semantic-preservation and fluency, and the attack success rate. As studied in
\citep{morris2020reevaluating}, we can filter adversarial examples with low quality based on hard constraints on semantic similarity and the number of added grammatical errors caused by adversarial perturbations. 

\begin{wrapfigure}{R}{0.56\textwidth}
\vspace{-10pt}
\begin{small}
\begin{minipage}{0.56\textwidth}
     \centering%
     \begin{subfigure}{0.45\textwidth}
         \centering
         \includegraphics[width=\textwidth]{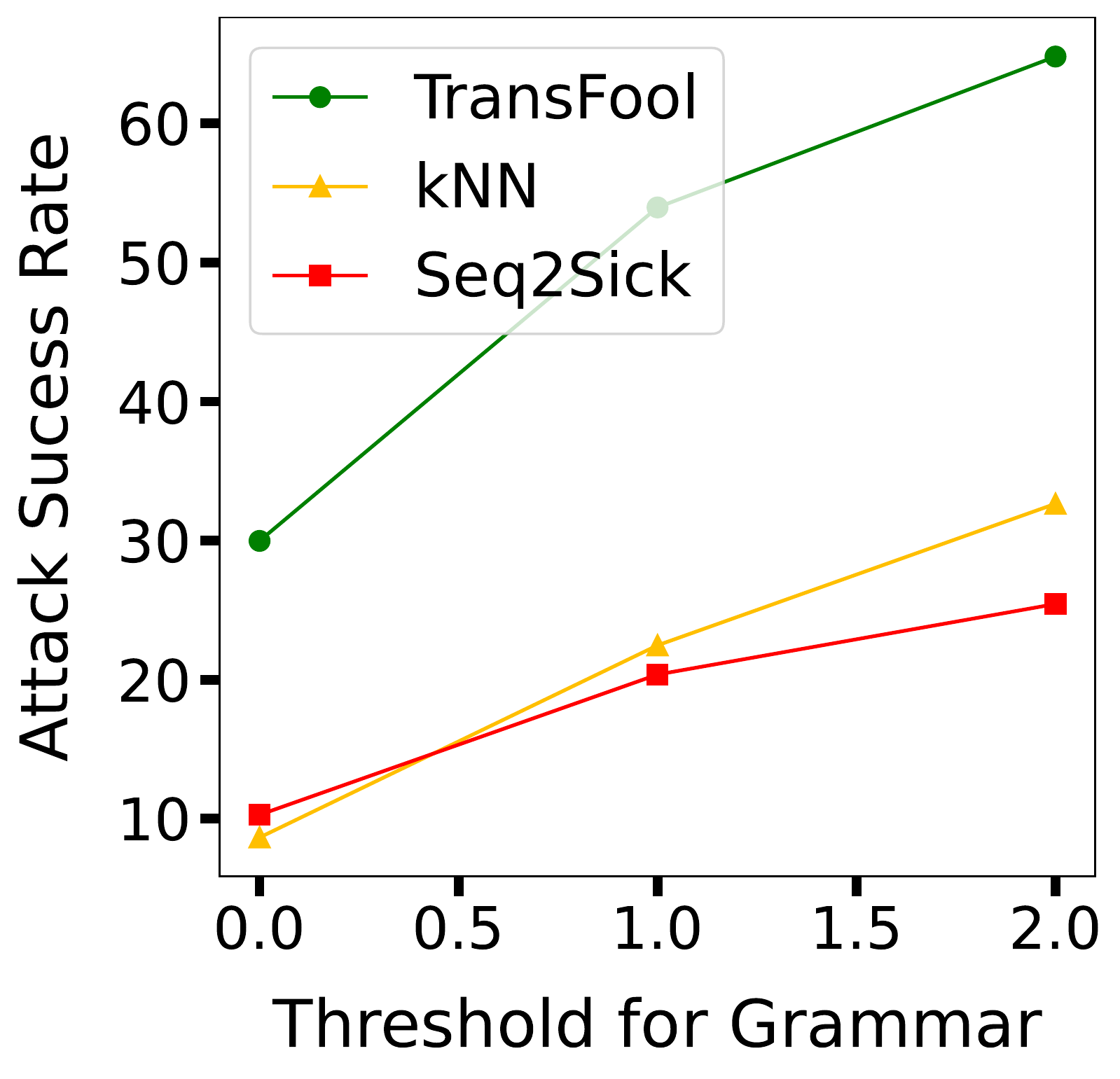}
         \vspace{-15pt}
         \caption{}
         \label{fig:tradeoff-gr}
     \end{subfigure}\qquad%
     \begin{subfigure}{0.45\textwidth}
         \centering
         \includegraphics[width=\textwidth]{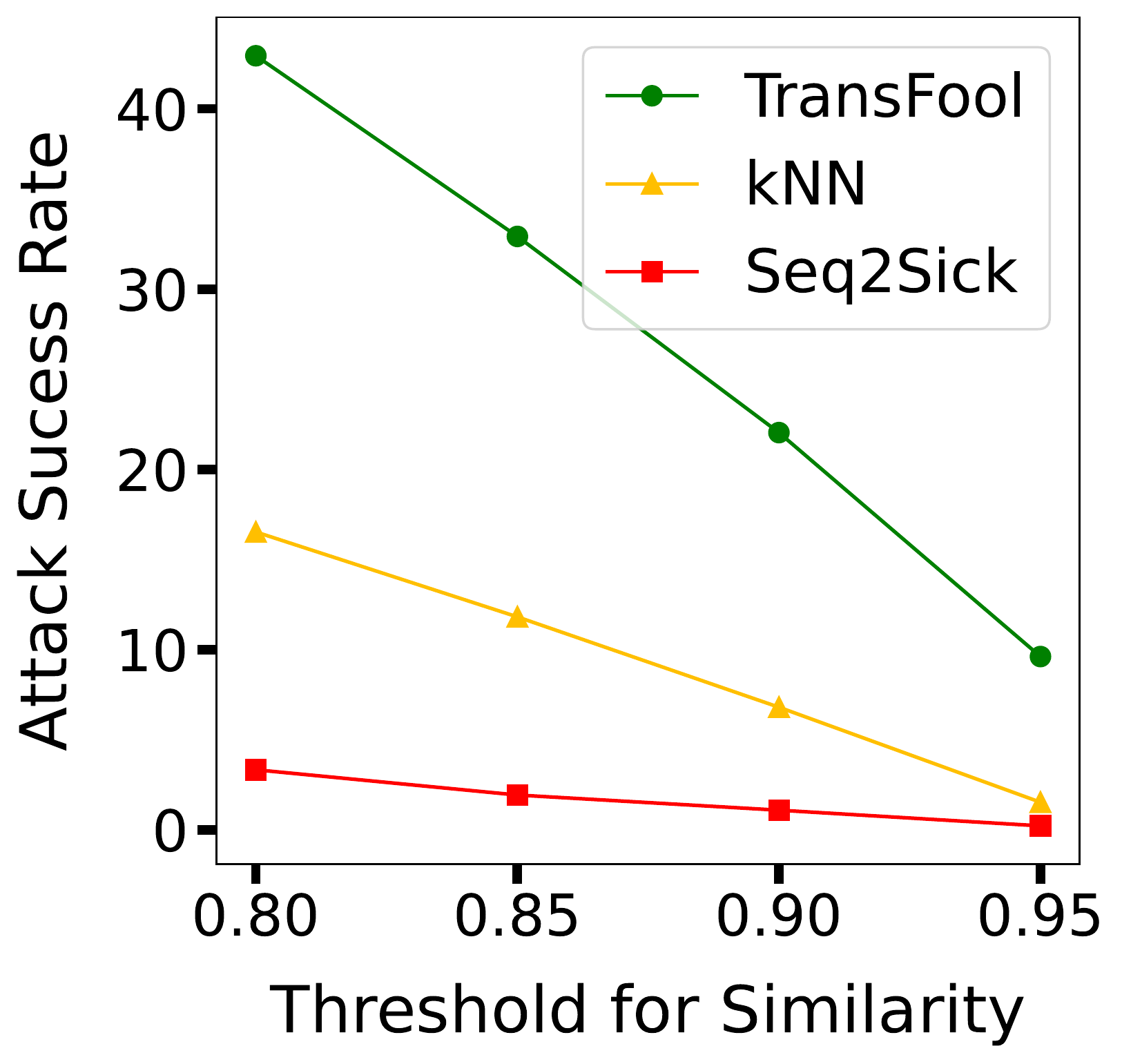}
         \vspace{-15pt}
         \caption{}
         \label{fig:tradeoff-sim}
     \end{subfigure}\,%

        \caption{Tradeoff between success rate and Similarity/fluency. The left figure shows the effect of acceptable number of added grammar errors by adversarial perturbation. The right figure shows the effect of similarity threshold.}
\end{minipage}
\end{small}
\vspace{-25pt}
\end{wrapfigure}

We can analyze the trade-off between success rate and similarity/fluency by setting different thresholds for filtering adversarial examples. If we evaluate the similarity by the sentence encoder suggested in \citep{morris2020reevaluating}, the success rate with different threshold values for similarity in the case of Marian (En-Fr) is depicted in Figure \ref{fig:tradeoff-sim}. By considering only the adversarial examples with a similarity higher than a threshold, the success rate decreases as the threshold increases, and the quality of the adversarial examples increases. 

Similarly, we can do the same analysis for fluency. As suggested in \citep{morris2020reevaluating}, we count the grammatical errors by LanguageTool \citep{naber2003rule} for the original sentences and the adversarial examples. Figure \ref{fig:tradeoff-gr} depicts the success rate for different thresholds  of the number of added grammatical errors caused by adversarial perturbations.

These analyses show that with tighter constraints, we can generate better adversarial examples while the success rate decreases. All in all, according to these results, TransFool outperforms the baselines for different thresholds of similarity and grammatical errors. 

\subsection{More Adversarial Examples} \label{samples_white}
In this Section, we present more adversarial examples generated by TransFool, kNN, and Seq2Sick. In order to show the effect of using LM embeddings on the performance of TransFool, we also include the generated adversarial examples against English to French Marian NMT model when we do not use LM embeddings. In all these tables, the tokens modified by TransFool are written in \textcolor{blue}{\textbf{blue}} in the original sentence, and the modified tokens by different adversarial attacks are written in \textcolor{red}{\textbf{red}} in their corresponding adversarial sentences. Moreover, the changes made by the adversarial attack to the translation that are not directly related to the modified tokens are written in \textcolor{orange}{orange}, while the changes that are the direct result of modified tokens are written in \textcolor{Brown}{brown}. 

As can be seen in the examples presented in Table \ref{tab:sample1}, 
TransFool makes smaller changes to the sentence. The generated adversarial example is a correct English sentence, and it is similar to the original sentence. However, kNN, Seq2Sick, and our method with the NMT embeddings 
make changes that are perceptible, and the adversarial sentences are not necessarily similar to the original sentence. The higher semantic similarity of the adversarial sentences generated by TransFool is due to the integration of LM embeddings and the LM loss in the proposed optimization problem.
We should highlight that TransFool is able to make changes to the  adversarial sentence translation that  are not directly related to the modifications of the original sentence but are the result of the NMT model failure.

\begin{table*}[h]
	\centering
		\renewcommand{\arraystretch}{.85}

	\setlength{\tabcolsep}{2pt}
	\caption{Adversarial examples against Marian NMT (En-Fr) by various methods (white-box).}\vspace{-2pt}
	\scalebox{0.73}{
		\begin{tabular}[t]{@{} l| >{\parfillskip=0pt}p{17cm} @{}}
			\toprule[1pt]
		    \textbf{Sentence}  & 
      \textbf{Text}\\
		
			\midrule[1pt]
			
		    \multirow{1}{*}{Org.} &  
      The most eager is \textcolor{blue}{\textbf{Oregon}}, which is enlisting 5,000 drivers in the country's biggest experiment. \hfill\mbox{}  \\
			 
			\cline{2-2}
			\rule{0pt}{2.5ex} 
			
			\multirow{2}{*}{Ref. Trans.} & 
   Le plus déterminé est l'Oregon, qui a mobilisé 5 000 conducteurs pour mener l'expérience la plus importante du pays. \hfill\mbox{}  \\
			
			\cline{2-2}
			\rule{0pt}{2.5ex} 
			
			\multirow{1}{*}{Org. Trans.} &  
   Le plus avide est l'Oregon, qui recrute 5 000 pilotes dans la plus grande expérience du pays. \hfill\mbox{}  \\
			 
			\cline{1-2}
			\rule{0pt}{2.5ex}

			\multirow{1}{*}{Adv. TransFool} &  
   The most eager is\textcolor{red}{\textbf{Quebec}}, which is enlisting 5,000 drivers in the country's biggest experiment. \hfill\mbox{} \\
			
			\cline{2-2}
			\rule{0pt}{2.5ex} 
			
			\multirow{2}{*}{Trans.} &  
   Le \textcolor{Brown}{Québec}, qui \textcolor{orange}{fait partie de} la plus grande expérience du pays, \textcolor{orange}{compte} 5 000 pilotes. \textcolor{orange}{(\textit{some parts are not translated at all.})} \hfill\mbox{} \\
			 
			\cline{1-2}
			\rule{0pt}{2.5ex} 
			
			\multirow{1}{*}{Adv. w/ NMT Emb.} &  
   The most eager is\textcolor{red}{\textbf{Custom}}, which is enlisting \textcolor{red}{\textbf{Disk}} drivers in the country's \textcolor{red}{\textbf{editions Licensee}}. \hfill\mbox{} \\
			
			\cline{2-2}
			\rule{0pt}{2.5ex} 
			
			\multirow{1}{*}{Trans.} &  
   Le plus avide est\textcolor{Brown}{Custom}, qui recrute \textcolor{Brown}{des} pilotes de \textcolor{Brown}{disque dans les éditions} du pays \textcolor{Brown}{ Licencié}. \hfill\mbox{} \\
			 
			\cline{1-2}
			\rule{0pt}{2.5ex} 
			
			\multirow{1}{*}{Adv. kNN} &  
   \textcolor{red}{\textbf{Theve}} eager is Oregon, \textcolor{red}{\textbf{C aren}} enlisting 5,000 drivers in \textcolor{red}{\textbf{theau}}'s biggest experiment. \hfill\mbox{} \\
			
			\cline{2-2}
			\rule{0pt}{2.5ex} 
			
			\multirow{1}{*}{Trans.} &  
   \textcolor{Brown}{Theve} avide est Oregon, \textcolor{Brown}{C sont} \textcolor{orange}{enrôlés} 5 000 pilotes dans la plus grande expérience de \textcolor{Brown}{Theau}. \hfill\mbox{} \\
			
			\cline{1-2}
			\rule{0pt}{2.5ex} 
			
			\multirow{1}{*}{Adv. Seq2Sick} &  
   The most \textcolor{red}{\textbf{buzz}} is \textcolor{red}{\textbf{FREE}}, which is \textcolor{red}{\textbf{chooseing Games comments}} in the country's great \textcolor{red}{\textbf{developer}}. \hfill\mbox{} \\
			
			\cline{2-2}
			\rule{0pt}{2.5ex} 
			
			\multirow{1}{*}{Trans.} &  
   Le plus \textcolor{Brown}{buzz est GRATUIT}, qui \textcolor{Brown}{est de choisir Jeux commentaires} dans le grand \textcolor{Brown}{développeur} du pays. \hfill\mbox{} \\

			\bottomrule[1pt]
		\end{tabular}
	}
	\label{tab:sample1}
\end{table*}

Other examples against different tasks and models are presented in  \crefrange{tab:sample3}{tab:sample7}.

\begin{table*}[ht]
	\centering
		\renewcommand{\arraystretch}{.85}

	\setlength{\tabcolsep}{2pt}
	\caption{Adversarial examples against Marian NMT (En-De) by various methods (white-box).}\vspace{-2pt}
	\scalebox{0.73}{
		\begin{tabular}[t]{@{} l| >{\parfillskip=0pt}p{17cm} @{}}
			\toprule[1pt]
		    \textbf{Sentence}  & 
      \textbf{Text}\\
		
			\midrule[1pt]
			
		    \multirow{3}{*}{Org.} &  
      The \textcolor{blue}{\textbf{devices}}, which track every mile a motorist drives and transmit that information to bureaucrats, are at the center of a \textcolor{blue}{\textbf{controversial}} attempt in Washington and \textcolor{blue}{\textbf{state}} planning offices to overhaul the \textcolor{blue}{\textbf{outdated}} system for funding America's major roads. \hfill\mbox{}  \\
			 
			\cline{2-2}
			\rule{0pt}{2.5ex} 
			
			\multirow{3}{*}{Ref. Trans.} & 
   Die Geräte, die jeden gefahrenen Kilometer aufzeichnen und die Informationen an die Behörden melden, sind Kernpunkt eines kontroversen Versuchs von Washington und den Planungsbüros der Bundesstaaten, das veraltete System zur Finanzierung US-amerikanischer Straßen zu überarbeiten. \hfill\mbox{}  \\
			
			\cline{2-2}
			\rule{0pt}{2.5ex} 
			
			\multirow{3}{*}{Org. Trans.} &  
   Die Geräte, die jede Meile ein Autofahrer fährt und diese Informationen an Bürokraten weiterleitet, stehen im Zentrum eines umstrittenen Versuchs in Washington und in den staatlichen Planungsbüros, das veraltete System zur Finanzierung der großen Straßen Amerikas zu überarbeiten.  \hfill\mbox{}  \\
			 
			\cline{1-2}
			\rule{0pt}{2.5ex}

			\multirow{3}{*}{Adv. TransFool} &  
   The \textcolor{red}{\textbf{vehicles}}, which track every mile a motorist drives and transmit that information to bureaucrats, are at the center of a \textcolor{red}{\textbf{unjustified}} attempt in Washington and \textcolor{red}{\textbf{city}} planning offices to overhaul the \textcolor{red}{\textbf{clearer}} system for funding America's major roads. \hfill\mbox{} \\
			
			\cline{2-2}
			\rule{0pt}{2.5ex} 
			
			\multirow{3}{*}{Trans.} &  
   Die \textcolor{Brown}{Fahrzeuge}, die jede Meile ein Autofahrer fährt und diese Informationen an Bürokraten weiterleitet, stehen im Zentrum eines \textcolor{Brown}{ungerechtfertigten} Versuchs in Washington und in den \textcolor{Brown}{Stadtplanungsbüros}, das \textcolor{Brown}{klarere} System zur Finanzierung der \textcolor{orange}{amerikanischen Hauptstraßen} zu überarbeiten. \hfill\mbox{} \\
			 
			\cline{1-2}
			\rule{0pt}{2.5ex}

			\multirow{3}{*}{Adv. kNN} &  
   The devices \textcolor{red}{\textbf{in}} which track every mile a motorist drives and transmit that \textcolor{red}{\textbf{M}} to bureaucrats, are \textcolor{red}{\textbf{07:0}} the center of a controversial attempt in Washington and state planning offices to overhaul the outdated \textcolor{red}{\textbf{Estate}} for funding America's major roads.\hfill\mbox{} \\
			
			\cline{2-2}
			\rule{0pt}{2.5ex} 
			
			\multirow{3}{*}{Trans.} &  
   Die \textcolor{orange}{Vorrichtungen}, \textcolor{Brown}{in denen} jede Meile ein Autofahrer fährt und diese \textcolor{Brown}{M} an Bürokraten \textcolor{orange}{überträgt}, sind \textcolor{Brown}{07:0} das Zentrum eines umstrittenen Versuchs in Washington und staatlichen Planungsbüros, das veraltete \textcolor{Brown}{Estate für} die Finanzierung der \textcolor{orange}{amerikanischen Hauptstraßen} zu überarbeiten. \hfill\mbox{} \\

			\cline{1-2}
			\rule{0pt}{2.5ex}

			\multirow{3}{*}{Adv. Seq2Sick} &  
   The devices, which \textcolor{red}{\textbf{road}} every\textcolor{red}{\textbf{ably}} a motorist drives and transmit that information to \textcolor{red}{\textbf{walnut socialisms}}, are at the center of a \textcolor{red}{\textbf{Senate}} attempt in Washington and state planning offices to\textcolor{red}{\textbf{establishment}} the outdated system for funding America's major \textcolor{red}{\textbf{paths}}. \hfill\mbox{} \\
			
			\cline{2-2}
			\rule{0pt}{2.5ex} 
			
			\multirow{3}{*}{Trans.} &  
   Die Geräte, die \textcolor{orange}{allgegenwärtig} ein Autofahrer \textcolor{orange}{antreibt} und diese Informationen an \textcolor{Brown}{Walnusssozialismen überträgt}, stehen im Zentrum eines \textcolor{Brown}{Senatsversuchs} in Washington und in den staatlichen Planungsbüros, das veraltete System zur Finanzierung der \textcolor{orange}{wichtigsten Wege} Amerikas \textcolor{Brown}{einzurichten}. \hfill\mbox{} \\

			\bottomrule[1pt]
		\end{tabular}
	}
	\label{tab:sample3}
\end{table*}

\begin{table*}[h]
	\centering
		\renewcommand{\arraystretch}{.85}

	\setlength{\tabcolsep}{2pt}
	\caption{Adversarial examples against Marian NMT (En-Zh) by various methods (white-box).}\vspace{-2pt}
	\scalebox{0.73}{
		\begin{tabular}[t]{@{} l| >{\parfillskip=0pt}p{17cm} @{}}
			\toprule[1pt]
		    \textbf{Sentence}  & 
      \textbf{Text}\\
		
			\midrule[1pt]
			
		    \multirow{1}{*}{Org.} &  
      And \textcolor{blue}{\textbf{what}} your husband said... {{if}} Columbus had done it, we'd all be Indians. \hfill\mbox{}  \\
			 
			\cline{2-2}
			\rule{0pt}{2.5ex} 
			
			\multirow{1}{*}{Ref. Trans.} & 
   \begin{CJK*}{UTF8}{gbsn}{ 你丈夫说的... 要是哥伦布没发现美洲,我们现在就都是印第安人了} \end{CJK*}    \hfill\mbox{}  \\
			
			\cline{2-2}
			\rule{0pt}{2.5ex} 
			
			\multirow{1}{*}{Org. Trans.} &  
   \begin{CJK*}{UTF8}{gbsn}{你丈夫说的话... 如果哥伦布做到了 我们都会是印第安人}\end{CJK*}    \hfill\mbox{}  \\
			 
			\cline{1-2}
			\rule{0pt}{2.5ex}

			\multirow{1}{*}{Adv. TransFool} &  
   And \textcolor{red}{\textbf{with}} your husband said... if Columbus had done it, we'd all be Indians. \hfill\mbox{} \\
			
			\cline{2-2}
			\rule{0pt}{2.5ex} 
			
			\multirow{1}{*}{Trans.} &  
   \begin{CJK*}{UTF8}{gbsn}{你丈夫说 如果哥伦布做到了 我们都会是印第安人}\end{CJK*} \textcolor{orange}{\textit{("..." is not in the translation.)}}   \hfill\mbox{}  \\ 
			 
			\cline{1-2}
			\rule{0pt}{2.5ex}

			\multirow{1}{*}{Adv. kNN} &  
   And what your husband said... if Columbus had\textcolor{red}{\textbf{60,}} we\textcolor{red}{\textbf{' Nineteen}} all \textcolor{red}{\textbf{it}} Indians. \hfill\mbox{} \\
			
			\cline{2-2}
			\rule{0pt}{2.5ex} 
			
			\multirow{1}{*}{Trans.} &  
   \begin{CJK*}{UTF8}{gbsn}{你丈夫说的话... 如果哥伦布\textcolor{Brown}{有60"} 我们\textcolor{Brown}{19}\textcolor{orange}{个印度}人}\end{CJK*}    \hfill\mbox{}  \\

			\cline{1-2}
			\rule{0pt}{2.5ex}

			\multirow{1}{*}{Adv. Seq2Sick} &  
   And \textcolor{red}{\textbf{completing}} your \textcolor{red}{\textbf{penalties}} said... if \textcolor{red}{\textbf{timely}} had done it, we'd all be \textcolor{red}{\textbf{briefed}}. \hfill\mbox{} \\
			
			\cline{2-2}
			\rule{0pt}{2.5ex} 
			
			\multirow{1}{*}{Trans.} &  
   \begin{CJK*}{UTF8}{gbsn}{\textcolor{Brown}{完成你的处罚 说}... 如果\textcolor{Brown}{及时完成,}我们都会\textcolor{Brown}{得到简报}}\end{CJK*}    \hfill\mbox{}  \\

			\bottomrule[1pt]
		\end{tabular}
	}
	\label{tab:sample4}
\end{table*}


\begin{table*}[h]
	\centering
		\renewcommand{\arraystretch}{.85}

	\setlength{\tabcolsep}{2pt}
	\caption{Adversarial examples against mBART50 (En-Fr) crafted by various methods (white-box).}\vspace{-2pt}
	\scalebox{0.73}{
		\begin{tabular}[t]{@{} l| >{\parfillskip=0pt}p{17cm} @{}}
			\toprule[1pt]
		    \textbf{Sentence}  & 
      \textbf{Text}\\
		
			\midrule[1pt]
			
		    \multirow{2}{*}{Org.} &  
      Wearing a wingsuit, he flew past over the famous Monserrate Sanctuary at 160km/h. The sanctuary is located at an altitude of over 3000 meters and numerous spectators had gathered there to watch his exploit. \hfill\mbox{}  \\
			 
			\cline{2-2}
			\rule{0pt}{2.5ex} 
			
			\multirow{2}{*}{Ref. Trans.} & 
   Equipé d'un wingsuit, il est passé à 160 km/h au-dessus du célèbre sanctuaire Monserrate, situé à plus de  3 000 mètres d'altitude, où de nombreux badauds s'étaient rassemblés pour observer son exploit. \hfill\mbox{}  \\
			
			\cline{2-2}
			\rule{0pt}{2.5ex} 
			
			\multirow{2}{*}{Org. Trans.} &  
   Il a survolé à 160 km/h le célèbre sanctuaire de Monserrate, situé à une altitude de plus de 3000   mètres, où de nombreux spectateurs se sont réunis pour assister à son exploit. \hfill\mbox{}  \\
			 
			\cline{1-2}
			\rule{0pt}{2.5ex}

			\multirow{2}{*}{Adv. TransFool} &  
   Wearing a wingsuit, he flew past over the famous \textcolor{red}{\textbf{Interesserrage}} Sanctuary at 160km/h. The sanctuary is   located at an altitude of over 3000 meters and numerous spectators had gathered there to watch his exploit. \hfill\mbox{} \\
			
			\cline{2-2}
			\rule{0pt}{2.5ex} 
			
			\multirow{2}{*}{Trans.} & 
   Le sanctuaire est situé à une altitude de plus de 3000 mètres \textcolor{orange}{et} de nombreux spectateurs se sont réunis  pour assister à son exploit. \textcolor{orange}{\textit{(first part of the sentence is not translated at all.)}} \hfill\mbox{} \\
			 
			\cline{1-2}
			\rule{0pt}{2.5ex}

			\multirow{2}{*}{Adv. kNN} &  
   Wearing a wingsuit\textcolor{red}{\textbf{.}} he flew past over the famous Monserrate Sanctuary at 160km/h. The sanctuary is     located at \textcolor{red}{\textbf{anzu opinionstitude}} of over \textcolor{red}{\textbf{8000}} meters and numerous spectators had gathered there \textcolor{red}{\textbf{the}} watch his exploit. \hfill\mbox{} \\

			\cline{2-2}
			\rule{0pt}{2.5ex} 
			
			\multirow{2}{*}{Trans.} &  
   Il a survolé le célèbre sanctuaire de Monserrate à 160 km/h. Le sanctuaire est situé à une \textcolor{Brown}{opiniontitude}   de plus de \textcolor{Brown}{8000} mètres \textcolor{orange}{et} de nombreux spectateurs se sont \textcolor{orange}{rassemblés là} pour \textcolor{orange}{observer} son exploit. \hfill\mbox{}  \\

			\cline{1-2}
			\rule{0pt}{2.5ex}

			\multirow{2}{*}{Adv. Seq2Sick} &  
   Wearing a wingsuit, he flew past over the famous Monserrate Sanctuary at 160km/h. The sanctuary is   located at an altitude of \textcolor{red}{\textbf{over74}} meters and numerous spectators had gathered there to watch his exploit. \hfill\mbox{} \\
			
			\cline{2-2}
			\rule{0pt}{2.5ex} 
			
			\multirow{2}{*}{Trans.} &  
   Il a survolé à 160 km/h le célèbre sanctuaire de Monserrate, situé à plus de   \textcolor{Brown}{74} mètres d'altitude, où de nombreux spectateurs se sont réunis pour assister à son exploit. \hfill\mbox{}  \\

			\bottomrule[1pt]
		\end{tabular}
	}
	\label{tab:sample5}
\end{table*}

\begin{table*}[h]
	\centering
		\renewcommand{\arraystretch}{.85}

	\setlength{\tabcolsep}{2pt}
	\caption{Adversarial examples against mBART50 (En-De) crafted by various methods (white-box).}\vspace{-2pt}
	\scalebox{0.73}{
		\begin{tabular}[t]{@{} l| >{\parfillskip=0pt}p{17cm} @{}}
			\toprule[1pt]
		    \textbf{Sentence}  & 
      \textbf{Text}\\
		
			\midrule[1pt]
			
		    \multirow{1}{*}{Org.} &  
      In Oregon, planners \textcolor{blue}{\textbf{are}} experimenting with giving drivers different choices. \hfill\mbox{}  \\
			 
			\cline{2-2}
			\rule{0pt}{2.5ex} 
			
			\multirow{1}{*}{Ref. Trans.} & 
   In Oregon experimentieren die Planer damit, Autofahrern eine Reihe von Auswahlmöglichkeiten zu geben. \hfill\mbox{}  \\
			
			\cline{2-2}
			\rule{0pt}{2.5ex} 
			
			\multirow{1}{*}{Org. Trans.} &  
   In Oregon experimentieren Planer damit, Fahrern verschiedene Wahlen zu geben. \hfill\mbox{} \\
			 
			\cline{1-2}
			\rule{0pt}{2.5ex}

			\multirow{1}{*}{Adv. TransFool} &  
   In Oregon, planners \textcolor{red}{\textbf{were}} experimenting with giving drivers different choices. \hfill\mbox{} \\
			
			\cline{2-2}
			\rule{0pt}{2.5ex} 
			
			\multirow{1}{*}{Trans.} &  
   In Oregon \textcolor{Brown}{experimentierten} Planer \textcolor{orange}{mit der Bereitstellung unterschiedlicher Wahlmöglichkeiten für Fahrer.} \hfill\mbox{} \\
			 
			\cline{1-2}
			\rule{0pt}{2.5ex}

			\multirow{1}{*}{Adv. kNN} &  
   \textcolor{red}{\textbf{in}} Oregon, planners \textcolor{red}{\textbf{nemmeno}} experimenting with\textcolor{red}{\textbf{kjer}} driver\textcolor{red}{\textbf{.}} different choices\textcolor{red}{\textbf{,}}  \hfill\mbox{} \\
			
			\cline{2-2}
			\rule{0pt}{2.5ex} 
			
			\multirow{1}{*}{Trans.} &  
   \textcolor{Brown}{in} Oregon\textcolor{orange}{,} Planer \textcolor{Brown}{nemmeno} experimentieren \textcolor{Brown}{mitkjer} Fahrer\textcolor{Brown}{.} verschiedene Wahlen\textcolor{Brown}{,} \hfill\mbox{}  \\

			\cline{1-2}
			\rule{0pt}{2.5ex}

			\multirow{1}{*}{Adv. Seq2Sick} &  
   \textcolor{red}{\textbf{acontece}}, planners are \textcolor{red}{\textbf{studying}} with \textcolor{red}{\textbf{Kivakapis against decisions,}} \hfill\mbox{} \\
			
			\cline{2-2}
			\rule{0pt}{2.5ex} 
			
			\multirow{1}{*}{Trans.} &  
   In \textcolor{Brown}{acontece studieren} Planer \textcolor{orange}{mit} \textcolor{Brown}{Kivakapis gegen Entscheidungen,} \hfill\mbox{}  \\

			\bottomrule[1pt]
		\end{tabular}
	}
	\label{tab:sample6}
\end{table*}

\begin{table*}[h]
	\centering
		\renewcommand{\arraystretch}{.85}

	\setlength{\tabcolsep}{2pt}
	\caption{Adversarial examples against mBART50 (En-Zh) crafted by various methods (white-box).}\vspace{-2pt}
	\scalebox{0.73}{
		\begin{tabular}[t]{@{} l| >{\parfillskip=0pt}p{17cm} @{}}
			\toprule[1pt]
		    \textbf{Sentence}  & 
      \textbf{Text}\\
		
			\midrule[1pt]
			
		    \multirow{2}{*}{Org.} &  
      Delegations are requested to submit the names of their representatives to the Secretary of the Preparatory Committee, \textcolor{blue}{\textbf{Ms.}} Vivian Pliner-Josephs (room \textcolor{blue}{\textbf{S}}-29\textcolor{blue}{\textbf{5}}0E; fax: (21\textcolor{blue}{\textbf{2}}) 96\textcolor{blue}{\textbf{3-5}}935). \hfill\mbox{}  \\
			 
			\cline{2-2}
			\rule{0pt}{2.5ex} 
			
			\multirow{1}{*}{Ref. Trans.} & 
   \begin{CJK*}{UTF8}{gbsn}{ 请各代表团将其代表姓名送交给筹备委员会秘书VivianPliner-Josephs女士(S-2950E室;电传:(212)963-5935)。} \end{CJK*}    \hfill\mbox{}  \\
			
			\cline{2-2}
			\rule{0pt}{2.5ex} 
			
			\multirow{1}{*}{Org. Trans.} &  
   \begin{CJK*}{UTF8}{gbsn}{请各代表团向筹备委员会秘书VivianPliner-Josephs(S-2950E室;传真:(212)963-5935)提出代表的姓名。}\end{CJK*}    \hfill\mbox{}  \\
			 
			\cline{1-2}
			\rule{0pt}{2.5ex}

			\multirow{2}{*}{Adv. TransFool} &  
   Delegations are requested to submit the names of their representatives to the Secretary of the Preparatory Committee, \textcolor{red}{\textbf{Mr.}} Vivian Pliner-Josephs (room \textcolor{red}{\textbf{C}}-29\textcolor{red}{\textbf{3}}0E; fax: (21\textcolor{red}{\textbf{1}}) 96 \textcolor{red}{\textbf{25-3}}0935). \hfill\mbox{}  \\
			
			\cline{2-2}
			\rule{0pt}{2.5ex} 
			
			\multirow{2}{*}{Trans.} &  
   \begin{CJK*}{UTF8}{gbsn}{请各代表团\textcolor{orange}{将其代表的姓名提交}筹备委员会秘书\textcolor{orange}{维维安·普林纳-约瑟夫斯先生(房间}\textcolor{Brown}{C}-29\textcolor{Brown}{3}0E;传真:(21\textcolor{Brown}{1})96\textcolor{Brown}{25-3}0935)。}\end{CJK*}    \hfill\mbox{}  \\ 
			 
			\cline{1-2}
			\rule{0pt}{2.5ex}

			\multirow{2}{*}{Adv. kNN} &  
   Delegations are requested to submit the names of their representatives \textcolor{red}{\textbf{that}} the Secretary of the Preparatory Committee, Ms. Vivian\textcolor{red}{\textbf{Pliner-Joseph,}} (room S-2950 \textcolor{red}{\textbf{•,}} fax: (212) 963-5935). \hfill\mbox{} \\
			
			\cline{2-2}
			\rule{0pt}{2.5ex} 
			
			\multirow{1}{*}{Trans.} &  
   \begin{CJK*}{UTF8}{gbsn}{请各代表团向筹备委员会秘书VivianPliner-Joseph(S-2950室;传真:(212)963-5935)\textcolor{orange}{递交}代表的姓名。}\end{CJK*}    \hfill\mbox{}  \\ 

			\cline{1-2}
			\rule{0pt}{2.5ex}

			\multirow{2}{*}{Adv. Seq2Sick} &  
   Delegations are requested to submit the names of their representatives to the Secretary of the Preparatory Committee, Ms.\textcolor{red}{\textbf{jadan}} Pliner-Josephs (room S-2950E; \textcolor{red}{\textbf{599}}: 212 96 \textcolor{red}{\textbf{2010,}}935. \hfill\mbox{} \\
			
			\cline{2-2}
			\rule{0pt}{2.5ex} 
			
			\multirow{1}{*}{Trans.} &  
   \begin{CJK*}{UTF8}{gbsn}{请各代表团\textcolor{orange}{将其代表的姓名提交}筹备委员会秘书\textcolor{orange}{贾丹·普林纳-约塞夫斯女士}(S-2950E室;\textcolor{Brown}{599:}212\textcolor{Brown}{962010,}935)。}\end{CJK*}    \hfill\mbox{}  \\ 

			\bottomrule[1pt]
		\end{tabular}
	}
	\label{tab:sample7}
\end{table*}

\clearpage

\section{More Results on the Black-box Attack} \label{black}

\subsection{Attacking Google Translate} \label{google}

\begin{wrapfigure}{R}{0.5\textwidth}
\vspace{-25pt}
\begin{small}
\begin{minipage}{0.5\textwidth}
\begin{table}[H]
	\centering
		\renewcommand{\arraystretch}{1.3}
	\setlength{\tabcolsep}{3pt}
	\caption{Performance of  black-box attack against Google Translate (En-Fr)
	.}
	\vspace{-5pt}
	\label{tab:google}
	\scalebox{0.85}{
		\begin{tabular}[t]{@{} lcccccccc @{}}
			\toprule[1pt]
		    \multirow{1}{*}{\textbf{Method}}  & 
		    \scalebox{0.95}{ASR$\uparrow$} & \scalebox{0.95}{RDBLEU$\uparrow$} & \scalebox{0.95}{RDchrF$\uparrow$} & \scalebox{0.95}{Sim.$\uparrow$} & \scalebox{0.95}{Perp.$\downarrow$} & \scalebox{0.95}{WER$\downarrow$} 
		       \\
			\midrule[1pt]
			TransFool &  \textbf{67.83} & \textbf{0.55} & \textbf{0.23} & \textbf{0.85} & \underline{184.35} & \underline{20.85}  \\ 
			 kNN & \underline{37.22} & \underline{0.35} & \underline{0.17} & \underline{0.82} & 389.45 & 30.24 \\
			 Seq2Sick & 23.49 & 0.20 & 0.15 & 0.75 &  \textbf{174.88} & \textbf{20.34} \\
			
			

			\bottomrule[1pt]
		\end{tabular}
	}
	\vspace {-10pt}
\end{table}

\begin{table}[H]
	\centering
		\renewcommand{\arraystretch}{1.3}
	\setlength{\tabcolsep}{3pt}
	\caption{Performance of TransFool black-box attack against Google Translate (En-De), when the target language is different.
	.}
	\vspace{-5pt}
	\label{tab:google2}
	\scalebox{0.79}{
		\begin{tabular}[t]{@{} lcccccccc @{}}
			\toprule[1pt]
		    \multirow{1}{*}{\textbf{Task}}  & 
		    \scalebox{0.95}{ASR$\uparrow$} & \scalebox{0.95}{RDBLEU$\uparrow$} & \scalebox{0.95}{RDchrF$\uparrow$} & \scalebox{0.95}{Sim.$\uparrow$} & \scalebox{0.95}{Perp.$\downarrow$} & \scalebox{0.95}{WER$\downarrow$} 
		       \\
			\midrule[1pt]
			En-Fr $\rightarrow$ En-De &  67.42 & 0.65 & 0.26 & 0.85 & 198.56 & 20.78  \\ 

			\bottomrule[1pt]
		\end{tabular}
	}
\end{table}
\end{minipage}
\end{small}
\vspace{-10pt}
\end{wrapfigure}

To evaluate the effect of different attacks in practice, we attack Google Translate\footnote{We should note that as we do not have a tokenizer, we compute Word Error Rate (WER) instead of Token Error Rate (TER).} by TransFool, kNN, and Seq2Sick. Since querying Google Translate is limited per day, we were not able to attack with WSLS, which requires high number of queries.  Table \ref{tab:google} presents the performance of the English to French translation task. The results demonstrate that adversarial sentences crafted by TransFool can degrade the translation quality more while preserving the semantics better. The perplexity score and word error rate of TransFool compete with those metrics of Seq2Sick, but Seq2Sick adversarial sentences are not as similar and are less effective. 

We also performed the cross-lingual black-box attack. We consider Marian NMT (En-Fr) as the reference model  and attack En-De Google Translate. The results for TransFool are reported in Table \ref{tab:google2}.

\subsection{Semantic Similarity Computed by Other Metrics} \label{sim_black}

\begin{wrapfigure}{R}{0.5\textwidth}
\vspace{-25pt}
\begin{small}
\begin{minipage}{0.5\textwidth}
\begin{table}[H]
	\centering
		\renewcommand{\arraystretch}{1}
	\setlength{\tabcolsep}{4pt}
	\caption{Similarity performance  of black-box attacks
	.}
	\vspace{-5pt}
	\label{tab:black_sim}
	\scalebox{0.85}{
		\begin{tabular}[t]{@{} lcccc @{}}
			\toprule[1pt]
		    \multirow{1}{*}{\textbf{Task}}  &
		    \multirow{1}{*}{\textbf{Method}} & 
            \scalebox{0.95}{USE$\uparrow$} & \scalebox{0.95}{BERTScore$\uparrow$} &
            \scalebox{0.95}{BLEURT-20 $\uparrow$} 
			\\
			\midrule[1pt]
			\multirow{3}{*}{En-Fr} & TransFool &  \textbf{0.85} & \textbf{0.95}  & \textbf{0.66} 
			\\ 
			& WSLS & 0.84 & {0.93} & 0.58 
			\\
			\midrule[1pt]
			\multirow{3}{*}{En-De} & TransFool &  {0.84} & \textbf{0.96} & \textbf{0.67} 
			\\ 
			& WSLS & \textbf{0.86} & {0.94} & 0.61 
			\\
			\midrule[1pt]
			\multirow{3}{*}{En-Zh} & TransFool &  \textbf{0.88} & \textbf{0.96} & \textbf{0.68} 
			\\ 
			& WSLS & 0.83 & {0.93} & 0.56 
			\\
			
			\bottomrule[1pt]
		\end{tabular}
	}
\end{table}
\end{minipage}
\end{small}
\vspace{-5pt}
\end{wrapfigure}
Similar to the white-box attack, we compute the similarity between the adversarial and original sentences by BERTScore and BLEURT-20, since they correlate well with human judgments. The similarity performance of TransFool and WSLS\footnote{The results of kNN and Seq2Sick are not reported as they are transfer attacks, and their performance is  reported in Table \ref{tab:white_sim}.} in the black-box settings are demonstrated in Table \ref{tab:black_sim}. According to Table \ref{tab:black_sim}, TransFool is better at maintaining semantic similarity. It may be because we used LM embeddings instead of the NMT ones in the similarity constraint.

\subsection{Some Adversarial Examples}\label{samples_black}

We also present some adversarial examples generated by TransFool and WSLS, in the black-box setting, in  Table \ref{tab:sample10}. 
In this table, the tokens modified by TransFool are written in \textcolor{blue}{\textbf{blue}} in the original sentence, and the modified tokens by different adversarial attacks are written in \textcolor{red}{\textbf{red}} in their corresponding adversarial sentences. Moreover, the changes made by the adversarial attack to the translation that are not directly related to the modified tokens are written in \textcolor{orange}{orange}, while the changes that are the direct result of modified tokens are written in \textcolor{Brown}{brown}. 
These examples show that modifications made by TransFool are less detectable, i.e., the generated adversarial examples are more natural and  similar to the original sentence. Moreover, TransFool makes changes to the translation that are not the direct result of the modified tokens of the adversarial sentence.

\begin{table*}[h]
	\centering
		\renewcommand{\arraystretch}{.85}

	\setlength{\tabcolsep}{2pt}
	\caption{Adversarial examples against mBART50 (En-Zh) crafted by various methods (black-box).}\vspace{-2pt}
	\scalebox{0.73}{
		\begin{tabular}[t]{@{} l| >{\parfillskip=0pt}p{17cm} @{}}
			\toprule[1pt]
		    \textbf{Sentence}  & 
      \textbf{Text}\\
		
			\midrule[1pt]
			
		    \multirow{2}{*}{Org.} &  
      \textcolor{blue}{\textbf{(}}c) To provide care and support by strengthening programming for orphans and vulnerable children \textcolor{blue}{\textbf{infected}}/\textcolor{blue}{\textbf{affected}} by AIDS and by expanding life skills training for young people. \hfill\mbox{}  \\
			 
			\cline{2-2}
			\rule{0pt}{2.5ex} 
			
			\multirow{2}{*}{Ref. Trans.} & 
   \begin{CJK*}{UTF8}{gbsn}{ (c)以加强协助艾滋病孤儿和被艾滋病感染/影响脆弱儿童的方案,以及扩大助益年轻人的生活技能培训方式,提供照顾和支助。} \end{CJK*}    \hfill\mbox{}  \\
			
			\cline{2-2}
			\rule{0pt}{2.5ex} 
			
			\multirow{1}{*}{Org. Trans.} &  
   \begin{CJK*}{UTF8}{gbsn}{(c)通过加强对艾滋病感染/受害的孤儿和脆弱儿童的方案和扩大对年轻人的生活技能培训,提供照顾和支助。}\end{CJK*}    \hfill\mbox{}  \\
			 
			\cline{1-2}
			\rule{0pt}{2.5ex}

			\multirow{2}{*}{Adv. TransFool} &  
   \textcolor{red}{\textbf{[}}c) To provide care and support by strengthening programming for orphans and vulnerable children \textcolor{red}{\textbf{Disabled}}/ \textcolor{red}{\textbf{afflicted}} by AIDS and by expanding life skill training for young people. \hfill\mbox{}  \\

			\cline{2-2}
			\rule{0pt}{2.5ex} 
			
			\multirow{1}{*}{Trans.} &  
   \begin{CJK*}{UTF8}{gbsn}{\textcolor{Brown}{[}c)通过加强\textcolor{orange}{为孤儿和受}艾滋病\textcolor{orange}{影响的弱势儿童提供照顾和支助,并}扩大对年轻人的生活技能培训。}\end{CJK*}    \hfill\mbox{}  \\ 
			 
			\cline{1-2}
			\rule{0pt}{2.5ex}

			\multirow{2}{*}{Adv. WSLS} &  
   (c) To provide \textcolor{red}{\textbf{nursing}} and \textcolor{red}{\textbf{unstinted\_}}support by strengthening \textcolor{red}{\textbf{i\_Lifetv}} for orphans and \textcolor{red}{\textbf{susceptable}} children infected/affected by \textcolor{red}{\textbf{CPR\_mannequins}} and by \textcolor{red}{\textbf{broadening}} life skills training for young people. \hfill\mbox{} \\

			\cline{2-2}
			\rule{0pt}{2.5ex} 
			
			\multirow{2}{*}{Trans.} &  
   \begin{CJK*}{UTF8}{gbsn}{(c)通过加强\textcolor{orange}{孤儿和受}\textcolor{Brown}{CPR\_迷彩}感染/\textcolor{Brown}{影响的易}\textcolor{orange}{受感染}儿童的\textcolor{Brown}{i\_Lifetv,并为}年轻人\textcolor{orange}{提供更广泛}的生活技能培训,提供护理和无毒的支助。}\end{CJK*}    \hfill\mbox{}  \\ 

			\cline{1-2}
			\rule{0pt}{2.5ex}

			\multirow{2}{*}{Adv. kNN} &  
   ( \textcolor{red}{\textbf{so}}) \textcolor{red}{\textbf{address}} provide care and support by strengthening \textcolor{red}{\textbf{prioritization}} for orphans and vulnerable children infected/affected by AIDS and by expanding life skills \textcolor{red}{\textbf{issue}} for young people. \hfill\mbox{} \\
			
			\cline{2-2}
			\rule{0pt}{2.5ex} 
			
			\multirow{2}{*}{Trans.} &  
   \begin{CJK*}{UTF8}{gbsn}{\textcolor{Brown}{因此,}通过加强对艾滋病感染/受害的孤儿和脆弱儿童的\textcolor{Brown}{优先事项}和扩大对年轻人的生活技能\textcolor{Brown}{的问题},\textcolor{orange}{解决}提供照顾和支助。}\end{CJK*}    \hfill\mbox{}  \\ 

			\cline{1-2}
			\rule{0pt}{2.5ex}

			\multirow{2}{*}{Adv. Seq2Sick} &  
   (c) To provide care and support by strengthening \textcolor{red}{\textbf{digital}} for \textcolor{red}{\textbf{dress}} and \textcolor{red}{\textbf{harmful}} children \textcolor{red}{\textbf{Journal}}/ \textcolor{red}{\textbf{Letter}} by \textcolor{red}{\textbf{Region}} and by \textcolor{red}{\textbf{disappear Violence}} skills training for young people. \hfill\mbox{} \\
			
			\cline{2-2}
			\rule{0pt}{2.5ex} 
			
			\multirow{1}{*}{Trans.} &  
   \begin{CJK*}{UTF8}{gbsn}{(c)通过加强\textcolor{Brown}{服装和有害}儿童的\textcolor{orange}{数字,}\textcolor{Brown}{按区域}\textcolor{orange}{分发}\textcolor{Brown}{新闻/信,并为}年轻人\textcolor{orange}{提供}\textcolor{Brown}{暴力}技能培训,提供照顾和支\textcolor{orange}{持}。}\end{CJK*}    \hfill\mbox{}  \\ 

			\bottomrule[1pt]
		\end{tabular}
	}
	\label{tab:sample10}
\end{table*}


\section{Effect of Back-Translation Model Choice on WSLS Performance} \label{back_trans}

WSLS uses a back-translation model for crafting an adversarial example. In \citep{zhang2021crafting}, the authors  investigate the En-De task and use  the winner model of the WMT19 De-En sub-track \citep{ng2019facebook} for the back-translation model.

\begin{wrapfigure}{R}{0.57\textwidth}
\vspace{-25pt}
\begin{small}
\begin{minipage}{0.57\textwidth}
\begin{table}[H]
	\centering
		\renewcommand{\arraystretch}{1}
	\setlength{\tabcolsep}{3.5pt}
	\caption{Performance of  WSLS (En-De) with two back-translation models
	.}
	\label{tab:black-translate}
	\scalebox{0.78}{
		\begin{tabular}[t]{@{} ccccccc @{}}
			\toprule[1pt]
		    \multirow{1}{*}{\textbf{Back-Translation}}  &
		    \textbf{ASR}  & \textbf{RDBLEU}  & \textbf{RDchrF}  & \textbf{Sim.}  & \textbf{Perp.}  & \textbf{\#Queries}   \\
			\midrule[1pt]
			\multirow{1}{*}{Marian NMT}  &  44.33 & 0.50 & 0.19 & 0.86 & 219.32 & 1262\\
			\midrule[1pt]
			\multirow{1}{*}{\citep{ng2019facebook}
			}  &  51.68 & 0.58 & 0.21 & 0.81 & 241.96 & 1307\\ 
			
			\bottomrule[1pt]
		\end{tabular}
	}
\end{table}
\end{minipage}
\end{small}
\vspace{-5pt}
\end{wrapfigure}
However, they do not evaluate their method for En-Fr and En-Zh tasks. To evaluate the performance of WSLS in Table \ref{tab:black}, We have used pre-trained Marian NMT models for all three back-translation models. In order to show the effect of our choice of back-translation model, we compare the performance of WSLS for the En-De task when we use Marian NMT or \citep{ng2019facebook} as the back-translation model in Table \ref{tab:black-translate}. As this Table shows, WSLS with Marian NMT as the back-translation model results in even more semantic similarity and lower perplexity score. On the other hand, WSLS with \citep{ng2019facebook} as the back-translation model has a slightly more success rate. These results show that our choice of back-translation model does not highly affect the performance of WSLS.

\section{License Information and Details} \label{license}
In this Section, we provide some details about the datasets, codes, and models used in this paper. We should note that we used the models and datasets that are available in HuggingFace transformers \citep{wolf-etal-2020-transformers} 
and datasets \citep{lhoest-etal-2021-datasets} 
libraries.\footnote{These two libraries are available at this GitHub repository: \url{https://github.com/huggingface}.} They are licensed under Apache License 2.0. Moreover, we used PyTorch for all experiments \citep{paszke2019pytorch}, which is released under the BSD license\footnote{\url{https://github.com/pytorch/pytorch/blob/master/LICENSE}}.

\subsection{Datasets}
\paragraph{WMT14} In the Ninth Workshop on Statistical Machine Translation, WMT14 was introduced for four tasks. We used the En-De and En-Fr news translation tasks. There is no license available for this dataset.

\paragraph{OPUS-100} OPUS-100 is a multilingual translation corpus for 100 languages, which is randomly sampled from the OPUS collection \citep{tiedemann2012parallel}. 
There is no license available for this dataset.

\subsection{Models}
\paragraph{Marian NMT} Marian is a Neural Machine Translation framework, which is mainly  developed by the Microsoft Translator team, and it is released under MIT License\footnote{\url{https://github.com/marian-nmt/marian/blob/master/LICENSE.md}}. This model uses a beam size of 4.

\paragraph{mBART50} mBART50 is a multilingual machine translation model of 50 languages, which has been introduced by Facebook. This model is published in the Fairseq library, which is released under MIT License\footnote{\url{https://github.com/facebookresearch/fairseq/blob/main/LICENSE}}. This model uses a beam size of 5.

\subsection{Codes}
\paragraph{kNN} In order to compare our method with kNN \citep{michel2019evaluation}, we used the code provided by the authors, which is released under the BSD 3-Clause "New" or "Revised" License.\footnote{The source code is available at \url{https://github.com/pmichel31415/translate/tree/paul/pytorch_translate/research/adversarial/experiments} and the license is avialable at \url{https://github.com/pmichel31415/translate/blob/paul/LICENSE}} 

\paragraph{Seq2Sick} To compare our method with Seq2Sick \citep{cheng2020seq2sick}, we used the code published by the authors.\footnote{The source code is available at \url{https://github.com/cmhcbb/Seq2Sick}.} There is no license available for their code.

\paragraph{WSLS} We implemented and evaluated WSLS \citep{zhang2021crafting} using the source code published by the authors.\footnote{\url{https://github.com/JHL-HUST/AdvNMT-WSLS}}

\appendix

\end{document}